\documentclass{article} 
\usepackage{iclr2026_conference,times}

\usepackage{amsmath,amsfonts,bm}

\def\eqref#1{equation~\ref{#1}}

\def\1{\bm{1}}

\DeclareMathAlphabet{\mathsfit}{\encodingdefault}{\sfdefault}{m}{sl}
\SetMathAlphabet{\mathsfit}{bold}{\encodingdefault}{\sfdefault}{bx}{n}

\usepackage{hyperref}
\usepackage{url}

\usepackage{titlesec}
\usepackage{floatrow}

\usepackage{xspace}

\usepackage{wrapfig}

\usepackage{amsmath}
\usepackage{amssymb}
\usepackage{mathtools}
\usepackage{amsthm}
\usepackage{pifont}
\usepackage{multirow}
\usepackage{booktabs}

\usepackage{algpseudocode}
\usepackage{algorithm}

\usepackage{multicol}

\usepackage{listings}

\usepackage{algpseudocode}
\usepackage{algorithm}
\usepackage[font=footnotesize]{caption}

\usepackage{bbm}
\usepackage{xspace}
\usepackage[dvipsnames]{xcolor}

\def\ours{\textsc{PaPSP}\@\xspace}
\def\oursma{\textsc{MA-PaPSP}\@\xspace}

\newcommand{\cmark}{\color{ForestGreen} \ding{51}}%
\newcommand{\xmark}{\color{red} \ding{55}}%

\def\upit#1{\textcolor{ForestGreen}{\textbf{#1}}}

\def\ie{\emph{i.e.}}
\def\eg{\emph{e.g.}}

\title{Leveraging Data to Say No: Memory Augmented Plug-and-Play Selective Prediction}

\author{Aditya Sarkar$^{1,2}$ \quad Yi Li$^{3}$ \quad Jiacheng Cheng$^{2,4,\dagger}$ \quad Shlok Mishra$^{1,5}$ \quad Nuno Vasconcelos$^{2}$\\
$^{1}$University of Maryland, College Park \quad $^{2}$University of California, San Diego \\  $^{3}$Qualcomm AI \quad $^{4}$Yale University \quad $^{5}$Meta AI
}

\iclrfinalcopy 
\begin{document}

\maketitle
\begingroup
\renewcommand\thefootnote{\fnsymbol{footnote}}
\footnotetext[2]{Corresponding author: Jiacheng Cheng.}  
\endgroup

\begin{abstract}

Selective prediction aims to endow predictors with a reject option, to avoid low confidence predictions. However, existing literature has primarily focused on closed-set tasks, such as visual question answering with predefined options or fixed-category classification. This paper considers selective  prediction for visual language foundation models, addressing a taxonomy of tasks ranging from closed to open set and from finite to unbounded vocabularies, as in image captioning. We seek training-free approaches of low-complexity, applicable to any foundation model and consider methods based on external vision-language model embeddings, like CLIP. This is denoted as {\it Plug-and-Play Selective Prediction\/} (\ours). We identify two key challenges: (1) \emph{instability of the visual-language representations}, leading to high variance in image-text embeddings, and (2) \emph{poor calibration of similarity scores}. To address these issues, we propose a {\it memory augmented\/} \ours (\oursma) model, which augments \ours with a retrieval dataset of image-text pairs. This is leveraged to reduce embedding variance by averaging retrieved nearest-neighbor pairs and is complemented by the use of contrastive normalization to improve score calibration. Through extensive experiments on multiple datasets, we show that \oursma outperforms \ours and other selective prediction baselines for selective captioning, image-text matching, and fine-grained classification. Code is publicly available at \url{https://github.com/kingston-aditya/MA-PaPSP}.
\end{abstract}    
\section{Introduction}
\label{sec:intro}

\begin{figure}[t]\RawFloats
    \centering
    \scriptsize
    \begin{tabular}{c}
         \includegraphics[width=0.75\textwidth]{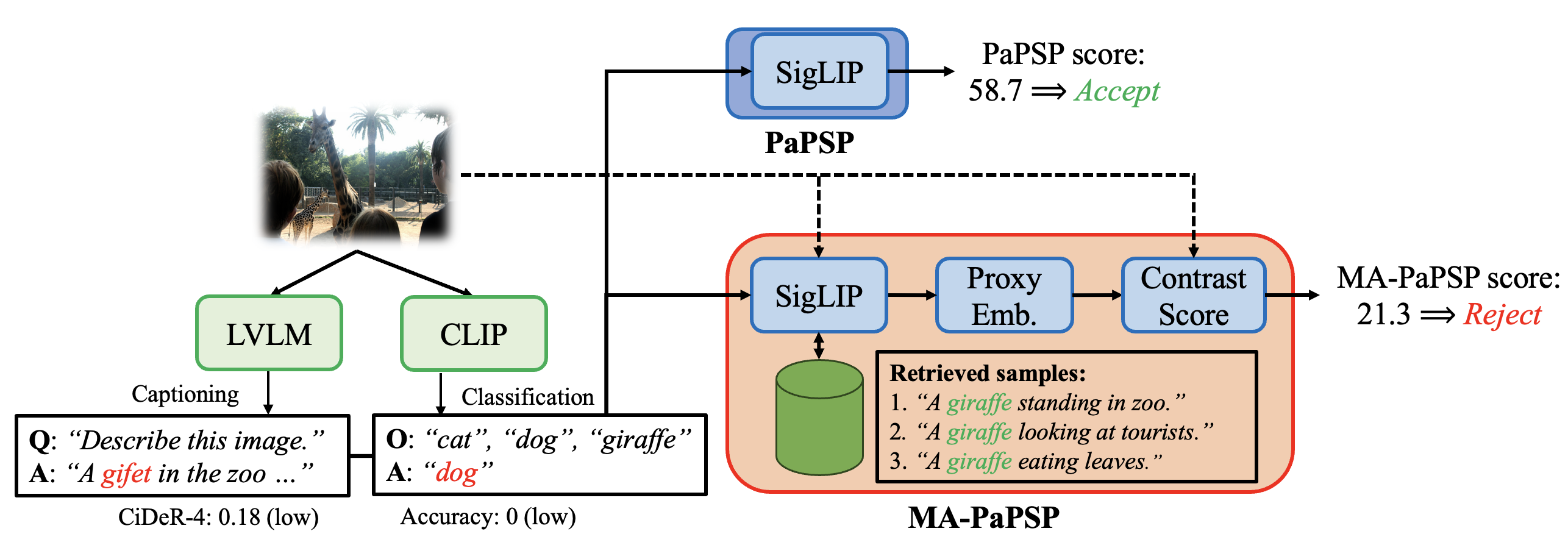}
    \end{tabular}
    \captionof{figure}{\ours uses an external representation model and the CLIP score to enable selective prediction for VLM tasks like captioning without training. \oursma augments this model with an external dataset, which is leveraged to estimate proxy embeddings of greater stability and better calibrated contrastive scores.  The figure shows an example where \ours fails but \oursma succeeds at rejecting an incorrect caption for the image shown. Also shown is the Cider-4 score between predicted and ground truth captions.}
    \label{fig:teaser}
\end{figure}

The success of vision-language models (VLMs) enables a wide range of promising applications.  However, these models are prone to erroneous predictions for reasons that include incorrect alignment of the two modalities, image or language ambiguities, or samples from the tails of their training distribution. This hampers their usefulness for real-world applications that require performance guarantees. As shown in Figure~\ref{fig:teaser}, \emph{selective prediction} (SP) \citep{chow1957optimum,chow1970optimum, el2010foundations, whitehead2022reliable, dancette2023improving, geifman2017selective, wang2018towards, srinivasan2024selective} methods address this problem by \emph{refusing} to make low confidence predictions, with the goal of minimizing the prediction \emph{risk} on the samples that they \emph{accept}. Optimal selective predictors optimize the trade-off between \emph{risk} and \emph{coverage}, which is the acceptance ratio throughout a dataset. While there is a literature in SP, the problem has mostly been considered for \emph{closed-set} prediction, usually with finite and small output vocabularies. Examples include classification with finite labels~\citep{wang2018towards, wu2020solving} or visual question answering from a finite set of choices~\citep{wang2023clipn, dancette2023improving}, \eg binary questions, questions involving a finite set of attributes, etc. However, vision-language tasks like the {\it selective captioning} problem of Figure~\ref{fig:teaser} require \emph{open-set} prediction, where the label set is infinite. This is, in fact, the regime that most foundation models operate in. There is a need for SP methods for open set tasks.

One possibility is to endow the foundation model with some prediction refusal score and train or fine-tune it to  optimize the ensuing risk-coverage trade-off. However, most foundation models are large and require large-scale training to avoid overfitting \citep{stevens2024bioclip, gu2025bioclip}. In many cases, the model may even be a proprietary black box that is impossible to retrain. In this work, we investigate an alternative solution, based on an {\it external SP model\/} that can be attached to any VLM, as illustrated in Figure~\ref{fig:teaser}, to compute a confidence score for the predictions of the latter.  SP can then be implemented by rejecting predictions of score below a threshold. This is denoted as {\it Plug-and-Play Selective Prediction} (\ours) and can be implemented by using a Large Visual Language Model (LVLM), \eg Qwen~\citep{bai2023qwen}, InternVL~\citep{zhu2025internvl3}, as \ours model, in what is called the LLM-as-judge strategy~\citep{dong2024can, ye2025learning}. However, LVLMs are large and have computationally expensive inference, sometimes much heavier than the VLM which makes the original predictions. Furthermore, they are not calibrated for \ours and requires some type of training or adaptation, which can itself be expensive and induce overfitting. Ideally, the \ours model should be lightweight and flexible enough to support any task without training.

In this work, we investigate the use of a visual language representation models (VLRMs), like CLIP~\citep{radford2021learning} or SigLIP~\citep{zhai2023sigmoid}, to implement \ours{}. This is denoted as the {\it Selective Prediction-VLM\/} (SP-VLM). A simple yet effective implementation, illustrated in the top branch of Figure~\ref{fig:teaser}, is to project the image and text produced by the foundation model into the SP-VLM embedding and reject image-text pairs of low CLIP-score~\citep{hessel2021clipscore}. We find, however that SP-VLM embeddings also suffer from calibration shortcomings, which we hypothesize to have two main causes: 1) SP-VLM representation \emph{instability}, which increases the variance of image and text embeddings, compromising the reliability of similarity scores, and 2) {\it poor calibration} of similarity score across the SP-VLM embedding. To overcome these, we propose the {\it Memory Augmented\/} \ours  (\oursma) procedure of the lower branch of Figure~\ref{fig:teaser}. The external SP-VLM is augmented with a retrieval dataset of image-text pairs. Methods that leverage this dataset are then proposed to 1) reduce the variance of the SP-VLM representation, by using the average representation of the nearest neighbors of the query image as a \emph{proxy embedding} for the latter, and 2) enhance score calibration, by using \emph{contrastive} scores that normalize the similarity score with respect to alternative text predictions. \oursma can be seen as a new instantiation of {\it memory augmented\/} approaches recently shown successful for various applications~\citep{nakata2022revisiting,zeng2024meacap,geirhos2024towards,silva2024learning}. Like these, it is a \emph{training-free} method that instead leverages an external dataset to enhance SP-VLM functionality. 
 
\oursma has several interesting properties. First, it is a relatively lightweight solution, usually much lighter than an LVLM. Second, it requires no training of either foundation model or the SP-VLM. Third, it can implement SP over a taxonomy of foundation model tasks, including classification~\citep{zhou2022conditional, zhou2022learning}, image-text matching~\citep{ma2023crepe, hsieh2024sugarcrepe}, and captioning, that range from closed to open set and from finite to unbounded prediction vocabularies. Fourth, it can be used with any foundation model. We show its effectiveness for models ranging from small VLRMs to large LVLMs.  Finally, \oursma can be tailored to any application domain by simply collecting data from that domain. This usually leads to improved performance over \ours, as illustrated in Figure~\ref{fig:teaser}. We show this for domains ranging from COCO captioning to medical imaging and that good performance can be achieved even with generic datasets, like CC12M \citep{changpinyo2021conceptual}, not directly tied to the application, for both open-set and closed-set SP tasks.

The  contributions of this work are three-fold: 1) we formulate the \ours problem over a taxonomy of SP tasks, ranging from closed to open set and finite to unbounded vocabularies. To our knowledge, this is the first attempt to solve \ours for all these tasks; 2)  we propose the lightweight, training free \oursma method, which leverages an external dataset to improve SP-VLM; and 3) we conduct an extensive evaluation demonstrating the efficacy of \oursma for selective captioning, image-text matching and classification tasks across a range of foundation models and application domains. 


\section{Related Works}
\label{sec:related}
\paragraph{Selective Prediction.} The problem of learning to reject has been studied in machine learning for over fifty years, primarily for unimodal classification~\citep{chow1957optimum,chow1970optimum,geifman2017selective, wang2018towards,black2022selective, srinivasan2024selective}. It aims to design models that abstain from making predictions of low confidence. Various metrics and frameworks have been proposed~\citep{geifman2019selectivenet,wang2018towards, de2000reject}. While these studies focus on selective classification and multi-way visual question-answering, where the label set is finite, \oursma targets the more challenging setting of open-set tasks (like selective captioning or image-text matching) where labels are sentences and their cardinality can be unbounded. 

\paragraph{Image-Text Alignment Score.} Selective prediction performance can be improved by model retraining~\citep{dancette2023improving}. However, for foundation models, this requires large scale training and is not practical for many applications. One possibility is to rely on external models, e.g. the LLM-as-judge strategy~\citep{ye2025learning, dong2024can}. However, these are memory and compute intensive. Recent studies have proposed using visual question-answering to assess image-text alignment, such as SeeTRUE~\citep{yarom2023you} or VQASquare~\cite{lin2024evaluating}. However, even these approaches rely on complex models, with long inference times. We show that they underperform \oursma, which has lower complexity.

\paragraph{Retrieval-Augmented Scoring.} Memory-augmented pipelines, first introduced for language generation~\citep{lewis2020retrieval}, have been extended to vision–language representation learning~\citep{iscen2023retrieval, iscen2023improving, xie2023ra}. While they do retrieval augmentation just like \oursma, most models' pretraining is very similar to CLIPs, and as a result, they have the same limitations like CLIP \ie unstable representations and uncalibrated scores. 
More related works are discussed in \ref{sec:morerelatedworks}.



\section{Plug and Play Selective Prediction}
\label{sec:prelims}

A VLM $f: \mathcal{X} \times \mathcal{Y} \rightarrow \mathcal{Y}$ maps an image ${\bf x} \in {\cal X}$ and a text ${\bf t} \in {\cal Y}$ into a text sequence ${\bf y} \in {\cal Y}$. Modern VLMs support a wide range of tasks. In this work, we consider a taxonomy of tasks of increasing complexity, which we refer to as the {\it VLM task taxonomy} in the remainder of the paper.

{\bf Level 1 - Coarse-grained Discrimination:} Tasks such as classification, where $f$ predicts a finite label set $\mathcal{Y}=\{\mathbf{y}_k\}_{k=1}^C$. 

{\bf Level 2 - Fine-grained Discrimination:} Classification-style tasks where the labels in $\mathcal{Y}$ are very similar. For example, in {\it image-text matching\/} (ITM), the model chooses between two similar image captions, \eg  ``A girl walking a cat.'' versus ``A girl walking a dog.''

{\bf Level 3 - Language Production:} Tasks like captioning, where  ${\cal Y}$ is unbounded, \eg the set of sentences in English.

\subsection{Selective Prediction}

In many applications, there is value in identifying and abstaining from making prediction errors. \emph{Selective prediction} \citep{chow1957optimum, el2010foundations, whitehead2022reliable} implements this by augmenting the model output with an option to abstain $\varnothing$. A selective model $h: \mathcal{X} \rightarrow  \mathcal{Y} \cup \{\varnothing \}$ usually composes  the model $f:\mathcal{X} \rightarrow{\cal Y}$ with a selection function $g:\mathcal{X} \rightarrow \{0,1\}$, 
\begin{equation}
    h({\bf x}) = (f,g)({\bf x}) = \begin{cases} 
      f({\bf x}) & \text{if } g({\bf x}) = 1 \\
      \varnothing & \text{if } g({\bf x}) = 0. 
   \end{cases}, \quad\text{where}\quad g({\bf x}) = \mathbbm{1} [ s({\bf x}, f({\bf x})) \geq \tau],
   \label{eq:h}
\end{equation}
where $\mathbbm{1}[\cdot]$ is the indicator function. 
If the selection function $g$ decides that a prediction should be made, the model prediction $f$ is output, otherwise $h$ abstains. 
The selection function $g$ is implemented by computing a score $s:{\cal X} \times {\cal Y}  \rightarrow \mathbb{R}$
for the confidence of prediction $f({\bf x})$ for image $\bf x$, which is compared to
a confidence threshold $\tau \in \mathbb{R}$.
Ideally, the score $s$ should yield a high (low) value when $f({\bf x})$ is correct (incorrect). 

Selective prediction aims to optimize the trade-off between coverage and risk~\citep{el2010foundations}. Given a set  $\mathcal{D}  = \{({\bf x}_i, {\bf a}_i)\in\mathcal{X}\times\mathcal{Y}\}_{i=1}^{|\mathcal{D}|}$ of image ${\bf x}_i$ and text ${\bf a}_i$  (captions or class label) pairs, coverage is the proportion of examples for which a prediction is made
\begin{equation}
    \mathcal{R}(f,g) = \frac{\frac{1}{|\mathcal{D}|} \sum_{i} \mathcal{L}(f({\bf  x}_i), {\bf a}_i) g({\bf x}_i)}{\mathcal{C}(g)}, \quad\text{where}\quad \mathcal{C}(g) = \frac{1}{|\mathcal{D}|} \sum_{i} g({\bf x}_i),
\end{equation}
while risk is the average loss on the covered subset,
where $\mathcal{L}:\mathcal{Y}\times\mathcal{Y}\rightarrow\mathbbm{R}$ is a loss function. In this work, we use the 0-1 loss function for classification problems and
the loss
\begin{equation}
    \mathcal{L}(f({\bf  x}_i), {\bf a}_i) = \begin{cases} 
      1 & \text{if } r(f({\bf  x}_i), {\bf a}_i) \geq \beta \\
      0 & \text{otherwise }
   \end{cases}
   \label{eq:loss}
\end{equation}
for captioning problems, where $r:\mathcal{Y}\times\mathcal{Y}\rightarrow\mathbb{R}$ is a measure of caption similarity, such as CiderN \citep{vedantam2015cider},
or METEOR \citep{banerjee2005meteor} and $\beta\in\mathbb{R}$ a similarity threshold. 
We compute the Area Under the Risk-Coverage curve (AURC)~\citep{geifman2018biasreduced} for a summary of performance across different coverage levels. The lower the area, the better the model. 


\subsection{\texorpdfstring{Plug and Play Selective Prediction (\ours)}{}}

\paragraph{Motivation.}  The tasks in the VLM task taxonomy can be performed by a multitude of models, including {\it visual language representation models\/} (VLRMs) like CLIP~\citep{radford2021learning} or {\it large visual language models} (LVLMs) like LLaVA~\citep{liu2023visual}. These are frequently large and/or black-box, in which cases selective prediction training is undesirable or impossible. To address this, we investigate the design of {\it plug-and-play selective prediction\/} (\ours) modules that can be attached to an existing VLM to implement the score function $s$ of (\ref{eq:h}). Ideally, the \ours{} module should support an {\it open vocabulary}, be relatively light-weight, and require no training. However, selective prediction has mostly been studied for closed-set classification problems, where $s$ can be derived from the posterior class probability estimates predicted by the classifier $f$. In fact, if the classifier produces calibrated probability estimates~\citep{guo2017calibration,cheng2022calibrating}, the largest class probability is the optimal score for the implementation of (\ref{eq:h})~\citep{chow1970optimum}.

For visual language tasks, the situation is far more complex. Because the normalization of probabilities by softmax type of operations  is not feasible when $\mathcal{Y}$ is unbounded, as in captioning, it is frequently impossible to obtain calibrated probability estimates for the model predictions. Unbounded prediction also introduces ambiguities, e.g. different captions can be semantically equivalent or otherwise match a given image, which do not appear in classification. To address this, we propose to implement \ours with an {\it external} CLIP-style VLRM, based on a pair of encoders of shared output space ${\cal F}^e$. An image $\mathbf{x}\in\mathcal{X}$ is mapped by an image encoder ${\phi}^e_{\text{img}}$ into a feature vector $\mathbf{v} = {\phi}^e_{\text{img}}(\mathbf{x}) \in {\cal F}^e$. Similarly, a caption $\mathbf{t}\in\mathcal{Y}$ is mapped by a text encoder ${\phi}^e_{\text{txt}}$ into a feature vector $\mathbf{w} = {\phi}^e_{\text{txt}}(\mathbf{t}) \in {\cal F}^e$. The encoder pair is pre-trained with a loss function that encourages the alignment of the feature vectors $\bf v$ and $\bf w$ in ${\cal F}^e$. Note the use of the  $e$ superscript to emphasize that the \ours model is {\it external}. The VLM making the original prediction $f$ can also be a CLIP-style VLM, \eg if the task is classification. To avoid confusion, we refer to the latter as the {\it predictive VLM} (P-VLM) and the VLM used to implement \ours as the {\it selective prediction VLM} (SP-VLM).

\begin{figure*}[t]\RawFloats
    \centering
    \begin{tabular}{c}
         \includegraphics[width=.98\textwidth]{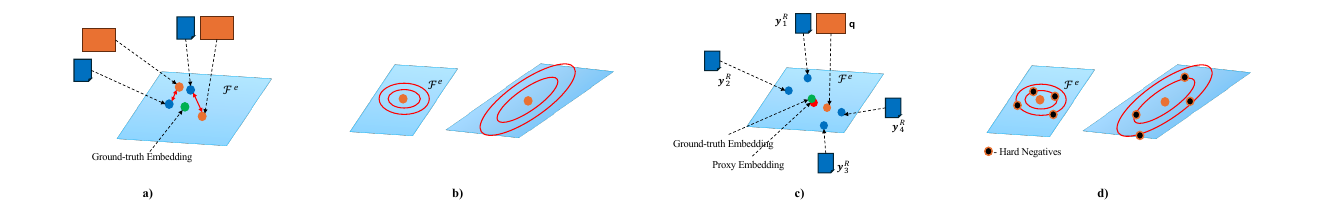}
    \end{tabular}
    \captionof{figure}{\textbf{Left:} VLM problems. a) \emph{instability of representations:} the representations of images (orange) and texts (blue) of the same concept can vary significantly, leading to unreliable similarity scores. b) {\it poor calibration:} distances between concepts of identical similarity (red ellipses) vary across the VLM embedding ${\cal F}^e$. 
    {\bf Right:} \ours solutions: c) \emph{proxy embeddings} of a query $\bf q$ (orange) average multiple nearest neighbor representations from a retrieval dataset (blue) to produce a more stable representation (red), closer to the concept ground-truth (green). d) \emph{contrastive scores} normalize similarity scores between image and predicted caption by those between the image and a set of hard-negatives, to ensure consistency across the space.}
    \label{fig:ex}
\end{figure*}

\begin{figure}[t]\RawFloats 
    \centering
    \begin{minipage}{0.48\linewidth}
    \centering
    \includegraphics[width=\textwidth,height=0.9in]{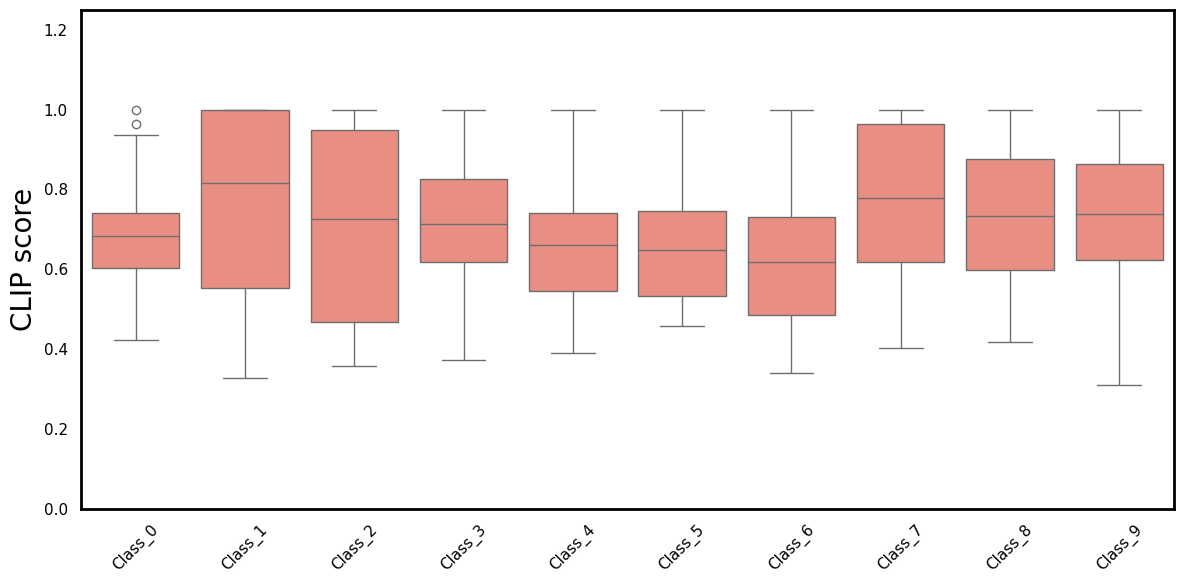}
    \caption{CLIP scores between class labels and images of the class. 
    }
    \label{fig:hypot_test}
\end{minipage}
    \begin{minipage}{0.5\linewidth}
    \centering
    \scriptsize{
    \begin{tabular}{c|ccc}
    \toprule
        Variant & Query & Proxy & Type \\
        \midrule 
        i2tr & $\phi_{\text{img}}(\bf x)$ & $\tilde{ {\bf \phi }}^e_\text{txt}({\bf x})$ & cross-modal\\
        i2ir & $\phi_{\text{img}}({\bf x}) $ & $\tilde{ {\bf \phi }}^e_\text{img}({\bf x})$ & uni-modal\\
        t2tr & $\phi_{\text{txt}}(f({\bf x}))$ & $\tilde{ {\bf \phi }}^e_\text{txt}({\bf f(x)})$ & uni-modal \\
        t2ir & $\phi_{\text{txt}}(f({\bf x}))$ & $\tilde{ {\bf \phi }}^e_\text{img}({\bf f(x)})$ & cross-modal\\
    \bottomrule
    \end{tabular}
    }
    \captionof{table}{Variants of retrieval-based proxy embeddings supported by \oursma.}
    \label{tab:variants}
    \end{minipage}
\end{figure}

We hypothesize that, if the text $\phi^{e}_{\text{txt}}$ and image $\phi^{e}_{\text{img}}$ encoders of the SP-VLM differ from those of the P-VLM, an erroneous prediction $f({\bf x})$ produced by latter is {\it unlikely\/} to induce a feature vector $\phi^{e}_{\text{txt}}(f({\bf x}))$ close to the projection $\phi^{e}_{\text{img}}({\bf x})$ of the image in the feature space ${\cal F}^e$ of the SP-VLM. Hence, a natural implementation of (\ref{eq:h}) is to simply compute the similarity score in  ${\cal F}^e$,
\begin{equation}
    s({\bf x}, f({\bf x})) = \text{cos}(\phi^{e}_\text{img}({\bf x}), \phi^{e}_\text{txt}(f({\bf x}))).
    \label{eq:baseSc}
\end{equation} 
This is known as the CLIP-score~\citep{hessel2021clipscore}. We denote its use in (\ref{eq:h}) as the \ours model and show that it has relatively weak selective prediction ability.
We hypothesize that this is due to two problems. The first is that the SP-VLM feature space is somewhat {\it unstable\/}, producing an imperfect alignment of images and text. This can be due to either poor training or just the ambiguity of image-text alignment, since different sentences can have similar meaning and sentences with almost the same words can have very different meaning. In result, as illustrated in Figure~\ref{fig:ex} a), there can be non-trivial variability across images (orange) and texts (blue) of the same groundtruth concept and the similarity scores of two equivalent image-text pairs can be quite different. The second is that while scores computed in ${\cal F}^e$ are effective for relative judgments of similarity, such softmax classification, they are too {\it poorly calibrated\/} for the absolute judgments required by the selection function $g$. As illustrated in Figure~\ref{fig:ex} b), the distances between an image or a text and semantically equivalent examples can vary significantly depending on the location of ${\cal F}^e$ where they are projected. 

We tested these hypotheses on UCF-101 \citep{soomro2012ucf101}. Class labels and images per class were projected into the feature space ${\cal F}^e$ of CLIP$_{\text{B/16}}$. The score of (\ref{eq:baseSc}) was then computed between images and their labels. Figure \ref{fig:hypot_test} shows that certain classes have high score variance, confirming the instability of the semantic representation. Some classes also exhibit much higher variance than others, confirming that the representation is poorly calibrated for certain classes or regions of ${\cal F}^e$. 
While these problems can be ameliorated by fine-tuning the SP-VLM on the target data, this can overfit and degrade model generalization. Fine-tuning is challenging for open set problems, like captioning, where the boundaries of ``application data" are not even well defined.

\section{\texorpdfstring{Memory Augmented \ours (\oursma)}{}}

\begin{figure*}\RawFloats
    \begin{minipage}{0.67\linewidth}
        \centering
        \includegraphics[width=\textwidth]{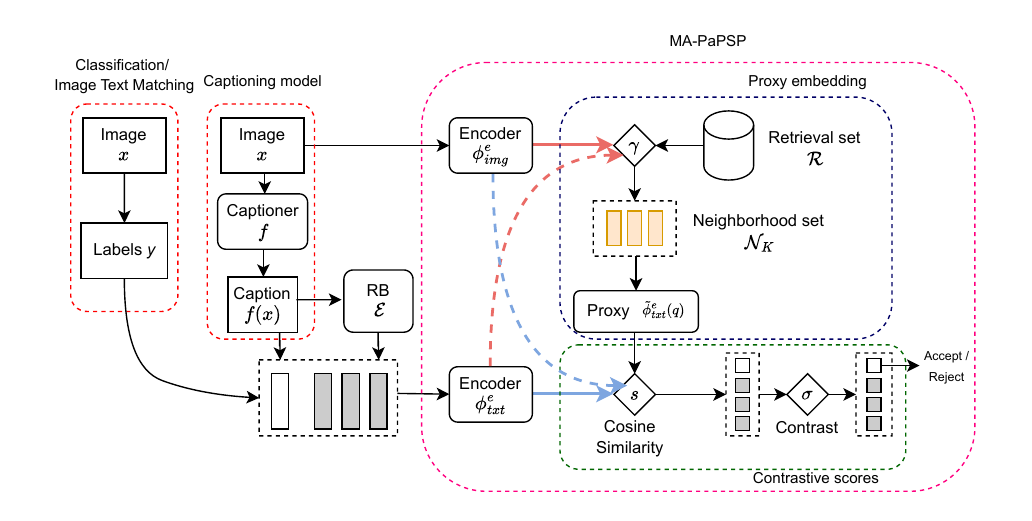}
    \caption{\textbf{\textbf{\texttt{\textbf{\texttt{\oursma}}}} architecture} (i2tr variant of Table~\ref{tab:variants}). 1) a VLM encoder extracts a query image embedding $\phi^{e}_\text{img}({\bf x})$. 2) This is used to retrieve a set ${\cal N}_k$ of image-caption pairs from retrieval set $\cal R$. 3) The captions in this set are used to compute a  proxy embedding $\tilde{\phi}^{e}_\text{txt}(f({\bf x}))$, which is a weighted average of the retrieved text embeddings according to (\ref{eq:wtilde_txt}). This serves as an estimate of the ground-truth caption of $\phi^{e}_\text{img}({\bf x})$. 4) A text embedding $\phi^{e}_\text{txt}({\bf x})$ is computed for the caption predicted by (in image captioning) or chosen by (in image-text matching) the VLM. 5) The \oursma score is computed by computing the cosine similarity between predicted, $\phi^{e}_\text{img}({\bf x})$, and estimated groundtruth, $\tilde{\phi}^{e}_\text{txt}(f({\bf x}))$, captions and using (\ref{eq:stc}) to compute a contrastive score. For image captioning, this leverages a set of hard-negative captions produced by an RB approach.}
    \label{fig:model1} 
    \end{minipage}
    \,
       \begin{minipage}{.31\linewidth}
        \centering
    \centering
    \setlength{\tabcolsep}{2pt}
    \begin{tabular}{c}
    \includegraphics[width=\linewidth]{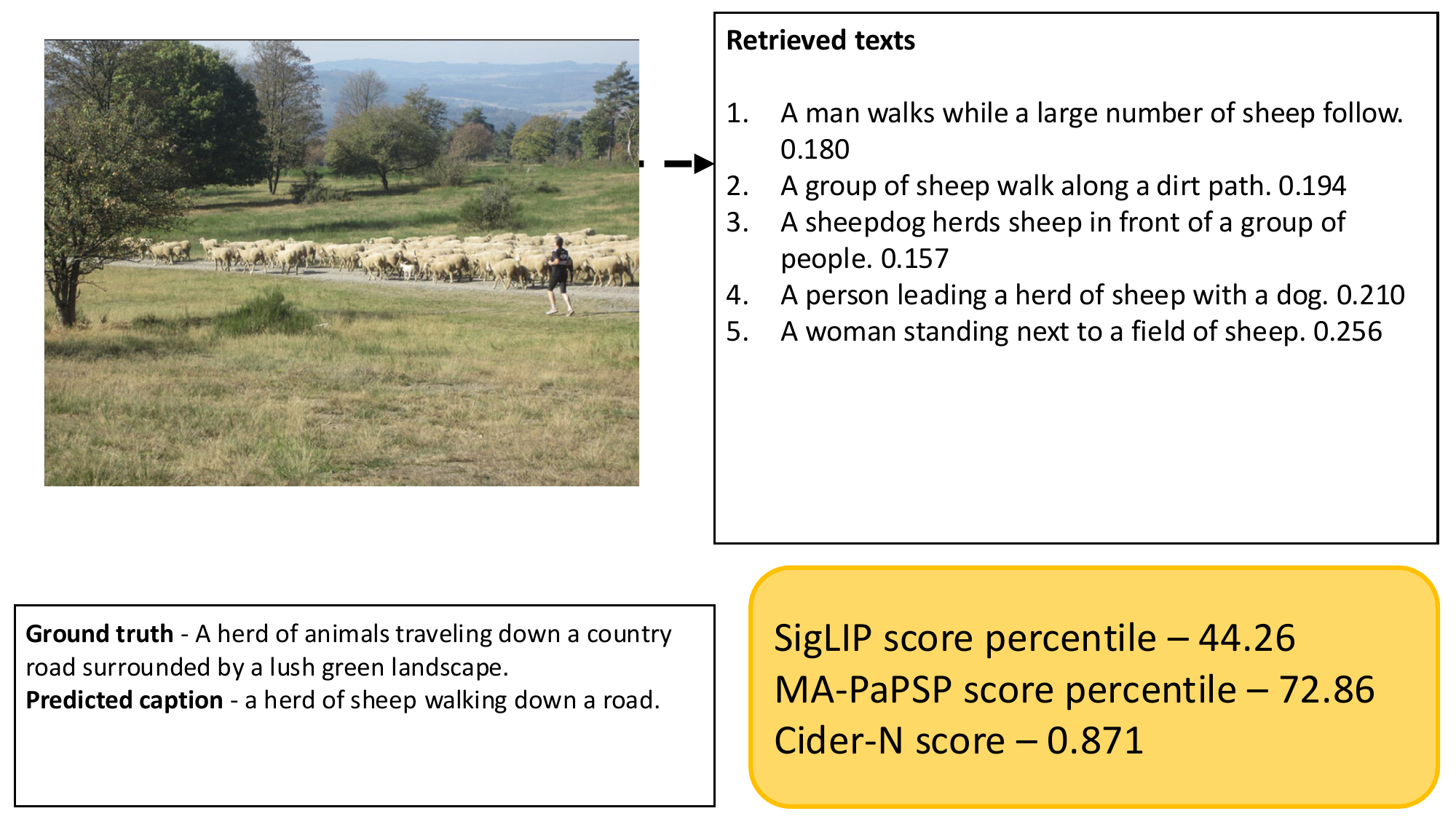} \\
    \includegraphics[width=\linewidth]{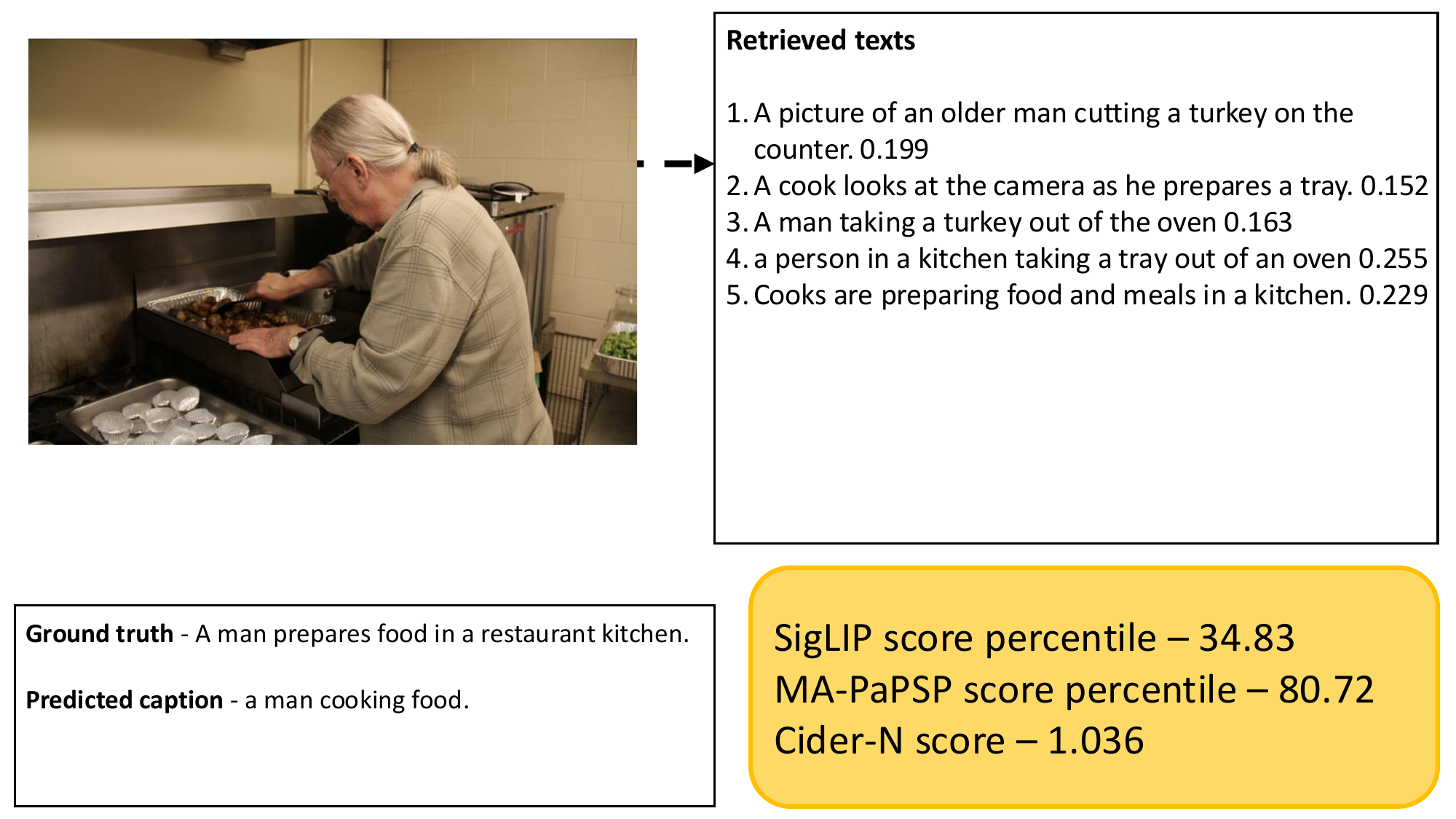}
    \end{tabular}
    \caption{Selective captioning by \oursma. Each example shows an image, predicted and ground-truth captions, retrieved captions with normalized weights, CLIP and \oursma scores, and the ground-truth score (CIDEr-N). 
    Top: CLIP accepts, \oursma rejects; Bottom: CLIP rejects, \oursma accepts.}
    \label{fig:Miscc}
        \end{minipage}
\end{figure*}

To improve the performance of \ours without training, we propose to complement the pre-trained SP-VLM with retrieval augmentation, as illustrated in Figure~\ref{fig:model1}. This is denoted as {\it Memory Augmented} \ours (\oursma) and relies on a reference dataset $\mathcal{R}=\{({\bf x}^{R}_i, {\bf y}^{R}_i)\}_{i=1}^{|\mathcal{R}|}$ of image-text pairs. The corresponding image-text projections  $\{{\bf z}^R_i =({\bf \phi}^e_{\text{img}}({\bf x}^{R}_{i}), {\bf \phi }^e_{\text{txt}}({\bf y}^{R}_{i})\}_{i=1}^{|\mathcal{R}|}$ in ${\cal F}^e$ are leveraged to address the instability and calibration problems of the pre-trained ${\cal F}^e$ using the two blocks of the figure, which are discussed next. 

\paragraph{Proxy Embeddings.} To overcome the instability of the pre-trained ${\cal F}^e$, \oursma estimates the groundtruth embedding of a query $\bf q$ by a {\it proxy embedding} $\tilde{ {\bf \phi }}^e_\text{mod}({\bf q})$. The query $\bf q$ can be an image $\bf x$ or a text $\bf y$ and \textbf{\texttt{mod}} the image or text modality. The proxy embedding is computed in two steps. First, the $K$ nearest neighbors of $\bf q$ are retrieved from ${\cal R}$ to form a neighborhood set 
\begin{equation}
    {\cal N}_K({\bf q}) = \{ ({\bf x}^R_i, {\bf y}^R_i) \in {\cal R} | i \in {\cal I}_K({\bf q})\}
    \label{eq:Nbg}
\end{equation}
where ${\cal I}_K({\bf q})$ is the set of indexes of $K$ nearest neighbors of $\bf q$ in ${\cal R}$. Retrieval is based
on a similarity measure $\gamma({\bf q}, {\bf z}^R_i)$, which can be either a measure of uni-modal similarity between images or between text vectors or a measure of cross-modal similarity between image and text vectors.  The proxy embedding is then computed by averaging neighbors according to their proximity to $\bf q$ 
\begin{equation}
    \tilde{ {\bf \phi }}^e_\text{mod}({\bf q})=\sum_{i\in {\cal I}_K({\bf q})} \frac{ \gamma({\bf q}, {\bf z}^R_i) }{\sum_{j\in {\cal I}_K({\bf q})} \gamma({\bf q}, {\bf z}^R_j)} {\bf \phi }^e_{\text{mod}}({\bf y}^R_{i}). 
    \label{eq:wtilde_txt}
\end{equation}
Note that these operations can produce a text proxy $\tilde{ {\bf \phi }}^e_{txt}({\bf q})$ or an image proxy $\tilde{ {\bf \phi }}^e_{img}({\bf q})$ for $\bf q$, independently of whether $\bf q$ is an image or a text. Table~\ref{tab:variants} summarizes the possible variants. Proxies of the modality of the query are denoted uni-modal, otherwise they are cross-modal.  

Figure~\ref{fig:ex} (c) illustrates the computation of the proxy embedding $\tilde{ {\bf \phi }}^e_{txt}({\bf q})$, where $\bf q$ is an image and $\gamma$ a cross-modal similarity function that retrieves the neighborhood set  ${\cal N}_{4}({\bf q})$ of 4 neighboring texts, which are used in the averaging operation of (\ref{eq:wtilde_txt}). This reduces the instability of the projections of the text embeddings in ${\cal F}^e$ and produces a proxy embedding closer to the ground-truth concept than any of the retrieved texts, thereby reducing the instability problem of Figure~\ref{fig:ex} (a). Figure~\ref{fig:Miscc} shows some real examples, where $\bf q$ is an image and $5$ captions are retrieved from $\cal R$ per example. 


\paragraph{Contrastive Scores.} Given the greater stability of the proxy representations, the simple replacement of the \ours similarity score of (\ref{eq:baseSc}) by a score based on either the image-based proxy $\tilde{ {\bf \phi }}^e_\text{img}({\bf x})$
\begin{equation}
    s^\text{ip}({\bf x}, f({\bf x})) = \text{cos}(\tilde{ {\bf \phi }}^e_\text{img}({\bf x}), \phi^{e}_\text{txt}(f({\bf x}))))
    \quad\text{or}\quad
    s^\text{tp}({\bf x}, f({\bf x})) = \text{cos}(\tilde{ {\bf \phi }}^e_\text{txt}({\bf x}), \phi^{e}_\text{txt}(f({\bf x}))))
    \label{eq:ipsc}
\end{equation} 
or the text-based proxy $\tilde{ {\bf \phi }}^e_\text{txt}({\bf x})$
should improve the selective prediction performance of the SP-VLM. This, however, does not address the poor calibration of the pretrained ${\cal F}^e$, illustrated in Figure~\ref{fig:ex} (b). The problem is that due to local variations of the geometry of this space, the cosine similarities between a feature vector (orange dot) and its neighbors of fixed semantic similarity (red ellipses) vary depending on the ${\cal F}^e$ location. To overcome this, we propose a to implement \oursma with a text-based contrastive score
\begin{equation}
    s^{\text{tc}}({\bf x}, f({\bf x}), {\cal E}(f({\bf x}))) = \frac{\exp(s^\text{tp}({\bf x}, f({\bf x)})/\tau)}{\sum_{{\bf y}_k \in {\cal E}(f({\bf x}))}\exp(s^\text{tp}({\bf x}, {\bf y}_k)/\tau)}
    \label{eq:stc}
\end{equation}
 where ${\cal E}(f({\bf x}))$ is a set of hard-negative captions produced from the P-VLM prediction $f({\bf x})$.  
In our implementation, these are obtained by replacing the nouns in $f({\bf x})$ with a different set of nouns, which significantly alter caption meaning. 
For this, we either used a small language model (SLM) or a rule based approach (RB) to ensure constructing a meaningful sentence that is very different from the main caption. 
This operation produces a set of close neighbors of $f({\bf x})$ with varying degrees of semantic similarity. The normalization of (\ref{eq:stc}) by the scores of these neighbors results in two scores in $[0,1]$, which have much more uniform magnitudes across the SP-VLM embedding ${\cal F}^e$ than the non-contrastive scores of (\ref{eq:ipsc}), as illustrated in Figure~\ref{fig:ex} (d). This is similar to the stronger homogeneity of softmax outputs than input logits for any classifier. Rule-based approach is discussed in supplementary section \ref{sec:moreablations}, and the SLM prompt used in our experiments for generating negatives is discussed in \ref{sec:qualres}. Note that hard-negatives are not needed for the classification and ITM tasks, since these already define a set of alternative labels. 


Figure~\ref{fig:model1} summarizes the proposed \oursma architecture. Given image $\mathbf{x}$, a P-VLM produces caption $f(\mathbf{x})$. The \oursma selective prediction model starts by mapping these into a pair of feature vectors, $\phi^{e}_\text{img}({\bf x})$ and text $\phi^{e}_\text{txt}(f({\bf x}))$, using the encoder pair of the SP-VLM. Either the image or text-based feature vector can then be used as a query $\mathbf{q}$. This is used to retrieve a set of $K$ image text-image pairs from a retrieval dataset $\cal R$, which are used to generate a proxy embedding for $\bf q$ with (\ref{eq:wtilde_txt}). This can be an embedding $\tilde{ {\bf \phi }}^e_\text{txt}({\bf q})$ based on the retrieved texts, or an embedding $\tilde{ {\bf \phi }}^e_\text{img}({\bf q})$ based on the retrieved images. The proxy embeddings are then used to compare a similarity score between image $\mathbf{x}$ and caption $f(\mathbf{x})$. This can be implemented with the \emph{base scores} of (\ref{eq:ipsc}) or the  \emph{contrastive score} of~(\ref{eq:stc}), which rely on a set $\cal E$  of hard negative variants of the caption $f(\mathbf{x})$ produced by RB, for captioning, or a set of class labels
$\{\phi_{\text{txt}}({\bf w}_k)\}_{k=1}^{|{\cal Y}|}$, for classification. 


\section{Experiments}
\label{sec:experts}
\vspace{-5pt}
In this section, we report on selective prediction experiments for three tasks representative of the different levels of the VLM task taxonomy: classification, image matching, and captioning.

\noindent{\bf Retrieval Set.} We considered three types of retrieval sets:1) \textbf{in-domain} retrieval set of training samples of the evaluation dataset, \textit{e.g.} training split of MS-COCO (110K); 2) \textbf{out-of-domain} retrieval set combining Conceptual Caption CC12M and CC3M \citep{changpinyo2021conceptual} and SBU-1M \citep{ordonez2011im2text}, which have no overlap with any of the evaluation datasets, captioned using BLIP-2; 3) a \textbf{mixed} dataset concatenating the two above. In all cases, there is \emph{no overlap} between evaluation and retrieval sets. \textbf{Out-of-domain} was used for retrieval unless otherwise specified. 


\noindent{\bf Predictive-VLM.} For classification and ITM, we compared P-VLMs of different different sizes (base and large). We present results for SigLIP \citep{zhai2023sigmoid} in the paper and for other models in supplementary section \ref{sec:moreablations}. For captioning, we mostly used BLIP w/ ViT-B \citep{li2022blip}, but also present results for InternVL~\citep{zhu2025internvl3} and Qwen-2.5-VL~\citep{bai2023qwen}. RB approach used in our experiments is described in \ref{sec:moreablations}.

\noindent{\bf Selective Prediction-VLM.} We compare various SP-VLMs of different sizes and present results for SigLIP \citep{zhai2023sigmoid} in the paper and other models in supplementary section \ref{sec:moreablations}. \ours (X)  denotes the implementation of \ours with SP-VLM X. X is omitted for the default SigLIP$_{\text{SO-400M}}$. 


\noindent{\bf Metrics.} Performance is evaluated with the risk and coverage measures of section~\ref{sec:prelims}. For captioning, the threshold $\beta$ of (\ref{eq:loss}) was set to 0.6 and 0.24 for Cider-N (N = 4 for all the experiments) and METEOR respectively. These thresholds were obtained empirically. Fig. \ref{fig:Miscc} shows examples of captions that are above/below these thresholds for different metrics.

\noindent{\bf Baselines.} \ours is compared to two Image-Text Alignment (ITA) scoring methods: VQASquare \citep{lin2024evaluating} and SeeTRUE \citep{yarom2023you}. More comparisons can be found in supplementary section \ref{sec:baselines}.

\noindent{\bf Results.} We performed many experiments to evaluate the effectiveness of \oursma. Due to space limitations, we present most of the ablations (size of the neighborhood set ${\cal N}_K$ and retrieval set $\cal R$, proxy embedding variants of Table~\ref{tab:variants}, inference latency, design choices, \ours with finetuned CLIP) on the supplemental sections \ref{sec:baselines} and \ref{sec:moreablations}. In this section, we ablate the components of the \oursma architecture and present comparisons to competing techniques. \oursma is implemented with the image-to-text retrieval (i2tr) variant of Table~\ref{tab:variants} and the contrastive score of (\ref{eq:stc}).

\begin{figure*}\RawFloats
\setlength{\tabcolsep}{2pt}
\centering
\scriptsize
\begin{tabular}{l|cc|cc|ccc|ccccc}
\toprule
\multirow{2}{*}{} & \multicolumn{4}{c|}{Captioning} & \multicolumn{3}{c|}{Classification} & \multicolumn{5}{|c}{ITM} \\
\multirow{2}{*}{Methods} & \multicolumn{2}{c|}{MS-COCO} & \multicolumn{2}{c|}{Flickr-30K} & \multirow{2}{*}{Flowers} & \multirow{2}{*}{Pets} & \multirow{2}{*}{UCF101} & \multirow{2}{*}{SugarCrepe} & \multirow{2}{*}{WinoGround} & \multirow{2}{*}{What'sUp} & \multirow{2}{*}{VL-Checklist} & \multirow{2}{*}{Foil}    \\
 & Cider-N & Meteor & Cider-N & Meteor & & & & & & & &  \\
 \midrule
 VQAScore & {\bf 0.146} &  0.321 & {\bf 0.241} & {\bf 0.312} & {\bf 0.211} & {\bf 0.207} & 0.217 & {\bf 0.146} & {\bf 0.240} & {\bf 0.236} & {\bf 0.226} & {\bf 0.217} \\
 SeeTRUE & 0.158 & {\bf 0.316} & 0.251 & 0.411 & 0.214 & 0.213 & {\bf 0.171} & 0.153 & 0.242  & 0.236 & 0.234 & 0.223 \\
 \midrule
\ours-S & 0.142 & 0.317 & 0.237 & 0.311 &  0.093 & 0.211 & 0.154 & 0.162 & 0.258 & 0.262 & 0.258 & 0.245 \\
 \oursma-S & {\bf 0.121} & {\bf 0.312} & {\bf 0.235} & {\bf 0.310} & {\bf 0.077} & {\bf 0.171} & {\bf 0.116} & {\bf 0.079} & {\bf 0.228} & {\bf 0.214} & {\bf 0.210} & {\bf 0.213} \\
 Gain (S) & \upit{14.78} & \upit{1.57} & \upit{0.84} & \upit{3.25} & \upit{17.20} & \upit{18.95} & \upit{24.67} & \upit{51.23} & \upit{3.00} & \upit{18.32} & \upit{18.60} & \upit{13.06}\\
 \midrule
\ours-L  & 0.136 & 0.311 & 0.229 & 0.307 & 0.074 & 0.169 & 0.113 & 0.078 & 0.208 & 0.202 & 0.204 & 0.197 \\
 \oursma-L & \textbf{0.109} & \textbf{0.286} & \textbf{0.219} & \textbf{0.297} & \textbf{0.063} & \textbf{0.114} & \textbf{0.088} & \textbf{0.062} & \textbf{0.196} & \textbf{0.192} & \textbf{0.192} & \textbf{0.189} \\
Gain (L) & \upit{19.85} & \upit{8.03} & \upit{4.36} & \upit{3.25} & \upit{14.86} & \upit{32.52} & \upit{22.12} & \upit{20.51} & \upit{5.76} & \upit{4.90} & \upit{5.88} & \upit{4.06} \\
\bottomrule
\end{tabular}
\captionof{table}{Selective prediction AURC ($\downarrow$). S denotes the use of a small SP-VLM ($\text{SigLIP}_{\text{B/16}}$), while L denotes the larger $\text{SigLIP}_{\text{SO-400M}}$. `Gain' is the \% improvement of \oursma over \ours. `Gain (B)' shows the same for SigLIP$_{\text{B/16}}$. }.
\label{tab: aurc-cap}
\end{figure*}

\paragraph{Selective captioning, ITM, and Classification.} 
\label{sec: cap-sec}
Table \ref{tab: aurc-cap} summarizes performance for tasks in  the three levels of the VLM task taxonomy. Several conclusions can be drawn. First, the VQAScore and SeeTRUE baselines underperformed even the implementation of \oursma with the smallest SP-VLM we tried (\oursma-S). This happens despite the fact that VQAScore and SeeTRUE rely on multiple LLM inferences to produce the selective prediction score and are thus much more computationally expensive. These results demonstrate the effectiveness of \oursma. Second, \oursma significantly outperforms \ours for equal SP-VLM (\% gains shown in green). For example, with small SP-VLM, on 7 of the 10 datasets at least one metric improves by $\approx15\%$ or more. In fact, \oursma using a small SP-VLM ($\text{SigLIP}_{\text{B/16}}$ - 16M parameters) outperforms \ours with a much larger SP-VLM ($\text{SigLIP}_{\text{SO-400M}}$ - 1B parameters) on all captioning and classification tasks. Third, both \ours and \oursma benefit from larger SP-VLMs, which usually have a more stable representation. Note that the gains of \oursma over \ours are {\it larger} for the larger model (L) on most captioning and classification tasks. This suggests that representation instability and, consequently, the effectiveness of \oursma hold independently of the model size.  
Fourth, the overall gains of the best version of \oursma (L) over the best LVLM baseline (VQAScore) are very significant in all tasks. Finally, comparing the performance differences across tasks, it can be seen that \oursma has the largest gains for classification and captioning. For ITM, the gains over \ours are large on SugarCrepe but less pronounced on the remaining datatsets.

\begin{figure*}\RawFloats
\setlength{\tabcolsep}{2pt}
\scriptsize
\centering
\begin{tabular}{c|cc|cc|ccc|ccccc}
\toprule
& \multicolumn{4}{c|}{Captioning} & \multicolumn{3}{c|}{Classification} & \multicolumn{5}{c}{ITM} \\
\multirow{3}{*}{Retrieval Set} & \multicolumn{2}{c|}{MS-COCO} & \multicolumn{2}{c|}{Flickr-30K} & \multirow{2}{*}{Flowers} & \multirow{2}{*}{Pets} & \multirow{2}{*}{UCF-101} & \multirow{2}{*}{SugarCrepe} & \multirow{2}{*}{Winoground} & \multirow{2}{*}{What'sUp} & \multirow{2}{*}{VL-checklist} & \multirow{2}{*}{Foil} \\                   
& Cider-N & Meteor & Cider-N & Meteor  & & & \\ 
\midrule
Random & \multicolumn{1}{c}{0.126}   & 0.288  & \multicolumn{1}{c}{0.234}   & 0.325   & 0.062   & 0.228  & 0.232 & 0.064   & 0.321  & 0.349   & 0.307 & 0.311 \\ 
In-Domain & \multicolumn{1}{c}{0.126}   & 0.288  & \multicolumn{1}{c}{0.229}   & 0.307 & 0.062   & {\bf 0.076}  & {\bf 0.078} & 0.066   & 0.192  & 0.189   & 0.207 & 0.167 \\
Out-of-Domain & \multicolumn{1}{c}{\color{blue}0.109}   & {\color{blue}0.286}  & {\color{blue}0.219}   & {\color{blue}0.297}  & 0.063   & 0.114  & 0.088 & {\bf {\color{blue}0.062}}  & 0.196  & 0.192   & {\color{blue}0.192} &  0.189 \\ 
Mixed & \multicolumn{1}{c}{\textbf{0.107}}   & \textbf{0.282}  & \multicolumn{1}{c}{\textbf{0.219}}   & \textbf{0.296} & \textbf{0.062}   & 0.078  & 0.084  & 0.068   & {\bf 0.192}  & {\bf 0.179}  & {\bf 0.177} & {\bf 0.154} \\
\bottomrule
\end{tabular}
\captionof{table}{Impact of retrieval set type on the AURC ($\downarrow$) of \oursma for captioning, classification, and ITM. Random dataset: MS-COCO for captioning, Flowers for classification, and SugarCrepe for ITM. {\color{blue} Blue} denotes improved performance  of out-of domain vs in-domain and {\bf boldface} the best results.}
\label{tab:aurc-design-pec}
\end{figure*}

\paragraph{Impact of the Retrieval Dataset.} Table \ref{tab:aurc-design-pec} shows how the type of retrieval dataset impacts \oursma performance. Beyond the in-domain, out-of-domain and mixed datasets, we also present results for a random dataset, which is one of the task datasets.  Comparing the random with in-domain results shows that the training dataset of a domain (e.g. Flowers classification) has good performance for that domain (e.g.  Flowers classification) but weaker performance for the remaining domains (e.g. Pets classification). Comparing the in-domain and out-of-domain results, it can be seen that the out-of-domain dataset overcomes this problem. For captioning, out-of-domain retrieval outperforms in-domain for all datasets. For ITM, which combines language and fine-grained discrimination, the two dataset categories have similar results. This shows that relying on a larger {\it generic\/} (out-of-domain) dataset (15M samples) can outperform or achieve similar performance to a smaller in-domain dataset (\eg 110K MS-COCO samples), for most language tasks. It is only for  classification, a very fine-grained task that requires almost no language modeling, that the in-domain dataset is usually superior. However, the out-of-domain dataset is vastly superior to a small domain specific dataset mismatched to the target domain (see random for Pets and UCF-101). Finally, the mixed dataset achieves the best of both worlds, obtaining the best results for almost all tasks and datasets. Overall, it can be inferred that  \oursma performance depends on both \emph{size} and \emph{domain coverage} of the retrieval set. A large generic dataset performs quite well, but can be outperformed by smaller retrieval sets covering the target domain, especially for classification. If the retrieval set is small and does not cover the target domain, performance degrades. We note that the choice of retrieval set is more of a problem for open-set problems like captioning and ITM. For classification, we found that a \emph{small number} of samples (\emph{15}) suffices. 

\begin{figure*}[t]\RawFloats
\begin{minipage}{0.5\linewidth}
\setlength{\tabcolsep}{1.5pt}
\scriptsize
\centering
\begin{tabular}{c|l|cc|cc}
\toprule
\multirow{2}{*}{P-VLM} & \multicolumn{1}{c|}{\multirow{2}{*}{Model}} & \multicolumn{2}{c|}{MS-COCO}          & \multicolumn{2}{c}{Flickr-30K}       \\ 
                           & \multicolumn{1}{c|}{}                       & \multicolumn{1}{c}{Cider-N} & Meteor & \multicolumn{1}{c}{Cider-N} & Meteor \\ \midrule
\multirow{2}{*}{BLIP-1 (0.1B)}    & \ours                                    & \multicolumn{1}{c|}{0.138}   & 0.320   & \multicolumn{1}{c|}{0.233}   & 0.313  \\  
                           & \oursma{}                               & \multicolumn{1}{c|}{\textbf{0.114}}     &
                           
                           \textbf{0.286}  & \multicolumn{1}{c|}{\textbf{0.211}}   & \textbf{0.289}  \\ \midrule
\multirow{2}{*}{BLIP-2 (2.7B)} & \ours                                      & \multicolumn{1}{c|}{0.136}   & 0.311  & \multicolumn{1}{c|}{0.229}   & 0.307  \\ 
                                                   & \oursma                               & \multicolumn{1}{c|}{\textbf{0.109}}   & \textbf{0.286}  & \multicolumn{1}{c|}{\textbf{0.219}}   & \textbf{0.297}  \\
                           
                           \midrule 
\multirow{2}{*}{InternVL-3.5 (4B)}   & \ours                                      & \multicolumn{1}{c|}{0.106}   & 0.206  & \multicolumn{1}{c|}{0.119}   & 0.207  \\ 
                           & \oursma                             & \multicolumn{1}{c|}{\textbf{0.068}}   & \textbf{0.152}  & \multicolumn{1}{c|}{\textbf{0.089}}   & \textbf{0.165}  \\ 
                           \midrule
\multirow{2}{*}{Qwen-2.5-VL (7B)} & \ours                                     & \multicolumn{1}{c|}{0.102}   & 0.204  & \multicolumn{1}{c|}{0.106}   & 0.168  \\ 
                           & \oursma                              & \multicolumn{1}{c|}{\textbf{0.066}}   & \textbf{0.162}  & \multicolumn{1}{c|}{\textbf{0.092}}   & \textbf{0.151}  \\ \bottomrule
\end{tabular}
\captionof{table}{Selective captioning AURC ($\downarrow$) for different P-VLMs. In all cases, the SP-VLM is SigLIP$_{\text{SO400M/14}}$.}
\label{tab: aurc-cross-caps1}
\end{minipage}
\begin{minipage}{0.5\linewidth}
\scriptsize
\setlength{\tabcolsep}{2pt}
\centering
\begin{tabular}{cr|ll|ll}
\toprule
\multicolumn{1}{c|}{\multirow{2}{*}{RS}} & \multicolumn{1}{c}{\multirow{2}{*}{SP-VLM}} & \multicolumn{2}{|c}{CheXpert}                              & \multicolumn{2}{|c}{MIMIC-CXR}                             \\
\multicolumn{1}{c|}{} & \multicolumn{1}{c}{}                        & \multicolumn{1}{|c}{Cider-N} & \multicolumn{1}{c}{Meteor} & \multicolumn{1}{|c}{Cider-N} & \multicolumn{1}{c}{Meteor} \\
\midrule
\multicolumn{6}{c}{\ours} \\
\midrule
\multicolumn{1}{c|}{} & SigLIP$_{\text{SO-400M}}$                                    & \multicolumn{1}{l}{0.264}   & 0.313                       & \multicolumn{1}{l}{0.272}   & 0.368                       \\ 
\multicolumn{1}{c|}{} & BioMedCLIP                                    & \multicolumn{1}{l}{\bf 0.164}   & {\bf 0.238}                       & \multicolumn{1}{l}{\bf 0.175}   & {\bf 0.229}                       \\ 
\midrule
\multicolumn{6}{c}{\oursma} \\
\midrule
\multicolumn{1}{c|}{\multirow{1}{*}{Out-of-domain}} & SigLIP$_{\text{SO-400M}}$                                       & \multicolumn{1}{l}{0.252}   & 0.326                       & \multicolumn{1}{l}{0.275}   & 0.276                       \\ 
\midrule
\multicolumn{1}{c|}{CheXpert} & SigLIP$_{\text{SO-400M}}$                         & \multicolumn{1}{l}{0.137}   & 0.276                       & \multicolumn{1}{l}{0.157}   & 0.326                      \\ 

\multicolumn{1}{c|}{MIMIC-CXR} & SigLIP$_{\text{SO-400M}}$                          & \multicolumn{1}{l}{\bf 0.134}   & {\bf 0.215}                     & \multicolumn{1}{l}{0.143}   & 0.292                      \\ 

\multicolumn{1}{c|}{Mixed} & SigLIP$_{\text{SO-400M}}$                          & \multicolumn{1}{l}{ 0.136}   & {\bf 0.217}                      & \multicolumn{1}{l}{\bf 0.132}   & {\bf 0.272}                       \\ 
\midrule
\multicolumn{1}{c|}{CheXpert} & BioMedCLIP                            & \multicolumn{1}{l}{\textbf{0.126}}   & \textbf{0.198}                       & \multicolumn{1}{l}{0.142}   & 0.216                      \\ 

\multicolumn{1}{c|}{MIMIC-CXR} & BioMedCLIP                            & \multicolumn{1}{l}{0.138}   & 0.208                     & \multicolumn{1}{l}{\textbf{0.124}}   & \textbf{0.202}                       \\ 

\multicolumn{1}{c|}{Mixed} & BioMedCLIP                            & \multicolumn{1}{l}{0.136}   & 0.204                      & \multicolumn{1}{l}{0.138}   & 0.268                       \\ 
\bottomrule
\end{tabular}
\captionof{table}{AURC ($\downarrow$) for biomedical models and datasets.}
\label{tab: aurc-caps-meds1}
\end{minipage}
\end{figure*}

\paragraph{Generalization across P-VLMs.} Table \ref{tab: aurc-cross-caps1} compares the selective captioning performance of \ours and \oursma for various P-VLMs, including BLIP and two LVLM models. Two conclusions can be inferred. First, for both methods, performance improves with P-VLM model size. This is because the captions generated by larger models LLMs are closer to ground truth captions. Second, while \oursma outperforms \ours irrespective of P-VLM size, the gains increase for the larger P-VLMs. This suggests that \oursma should be effective for even larger models.

\paragraph{Generalization to Specialized Domains.} To test the flexibility of \oursma, we considered the worst-case scenario of applying it to very specialized domains for which neither the P-VLM or the SP-VLM  are specifically trained. These experiments addressed selective captioning on the medical datasets CheXpert-5x200 \citep{irvin2019chexpert} and MIMIC-CXR-5x200 \citep{johnson2019mimic}, using Qwen-2.5-VL (7B) as P-VLM. We compared generic versions of \ours and \oursma  ($\text{SigLIP}_{\text{SO-400M}}$ and the out-of-domain dataset) to implementations of \ours and \oursma with a specialized SP-VLM, BioMedCLIP \citep{zhang2023biomedclip},
and three retrieval sets: CheXpert, MIMIC-CXR, and a mixed dataset combining the above plus ROCO~\citep{ruckert2024rocov2}, PadChest~\citep{de2025padchest}, COVID~\citep{shuja2020covid} and RSNA~\citep{colak2021rsna}. Ground truth captions were created with template: \textit{``An image of \{class label\}"}.  The results of table \ref{tab: aurc-caps-meds1} support the following conclusions. First, for comparable SP-VLM, \ours was comparable to \oursma with the out-of-domain dataset but much weaker than \oursma with any of the specialized datasets.  In  fact, starting from generic  \ours{} ($\text{SigLIP}_{\text{SO-400M}}$), it is more beneficial to maintain the SP-VLM and use \oursma with specialized datasets than to implement \ours with the specialized SP-VLM (BioMedCLIP). This confirms the power and flexibility of \oursma.  Second, with the specialized SP-VLM (BioMedCLIP) all specialized retrieval sets achieved results similar to the best obtained on domains like MS-COCO. This illustrates how \oursma{} can leverage models and data that are available already to cover a vast spectrum of domains without training. 


\paragraph{Proxy estimation and Contrastive components.} Table \ref{fig:compos2} presents an ablation study of the two main components of \oursma{}. Starting from the \ours model, both proxy estimation and contrastive scores improve AURC performance and the combination of the two provides an additional gain. This is evidence in support of the two hypothesis of Figure~\ref{fig:ex}, namely the high variance of the SP-VLM representation and its poor calibration. The gains are larger for classification, followed by ITM, and captioning, suggesting greater benefits for finer-grained tasks.


\begin{figure}[t]\RawFloats
\scriptsize
\centering
\setlength{\tabcolsep}{2pt}
\begin{tabular}{cc|cc|ccc|ccccc}
\toprule
& & \multicolumn{2}{|c|}{Captioning} &\multicolumn{3}{c|}{Classification} &\multicolumn{5}{c}{ITM} \\
Proxy & Contrastive & MS-COCO & Flickr-30K & Flowers & UCF-101 & Pets &  SugarCrepe & WinoGround & What'sUp & VL-Check & Foil\\
\midrule
\xmark & \xmark  & 0.160 & 0.243 & 0.172 & 0.143 & 0.178& 0.204 & 0.237 & 0.231 & 0.232 & 0.242\\
\cmark & \xmark & 0.143 & 0.239 & 0.152 & 0.147 & 0.163 & 0.174 & 0.223 & 0.231 & 0.222 & 0.216 \\
\xmark & \cmark & 0.156 & 0.241 &  0.121 & 0.142 & 0.136 & 0.082 & 0.199  & 0.202 & 0.206 & 0.198\\
\cmark & \cmark & \textbf{0.109} & \textbf{0.219}  & \textbf{0.063} & \textbf{0.114} & \textbf{0.088} & \textbf{0.062} & \textbf{0.196} & \textbf{0.192} & \textbf{0.192} & \textbf{0.189}\\
\bottomrule
\end{tabular}
\captionof{table}{AURC ($\downarrow$) ablation of the impact of proxy embedding and contrastive scores on \oursma{}. In all cases, the SP-VLM is $\text{SigLIP}_{\text{SO-400M}}$.}
\label{fig:compos2}
\end{figure}


\section{Limitations and Conclusion}
\label{sec:conclude}
This work extends the problem of selective prediction for foundation models to open-set problems such as captioning. We proposed \oursma, a memory-augmented approach that leverages an external visual language representation model and a dataset of text-image pairs to improve embedding stability and calibration of similarity scores. This enables a lightweight solution, that requires no training and can be applied to any foundation model. Experiments on a taxonomy of visual-language tasks demonstrated the effectiveness, flexibility, and robustness of  \oursma for a range of foundation models. However, they have also shown that performance depends on the overlap between the retrieval set and the target application domain. While large generic datasets can achieve good performance, specialized datasets and even representation models may enable non-trivial gains for specialized domain, as we have demonstrated for medical imaging. This may require some trial and error by practitioners, to achieve the very best performance on a specific domain. 

\section*{Acknowledgment}
This work was partially funded by NSF awards IIS-2303153 and NAIRR-240300, the NVIDIA Academic grant, and a gift from Qualcomm. We also acknowledge the NRP Nautilus cluster, used for some of the experiments discussed above.

\clearpage

\bibliography{iclr2026_conference}

@String(IJCV = {Int. J. Comput. Vis.})

@String(CVPR= {IEEE Conf. Comput. Vis. Pattern Recog.})

@String(ICCV= {Int. Conf. Comput. Vis.})

@String(ECCV= {Eur. Conf. Comput. Vis.})

@String(ICLR = {Int. Conf. Learn. Represent.})

@String(AAAI = {AAAI})

@String(CVPRW= {IEEE Conf. Comput. Vis. Pattern Recog. Worksh.})

@String(IJCV  = {IJCV})

@String(CVPR  = {CVPR})

@String(ICCV  = {ICCV})

@String(ECCV  = {ECCV})

@String(ICLR  = {ICLR})

@String(CVPRW= {CVPRW})

@article{chow1970optimum,
  title={On optimum recognition error and reject tradeoff},
  author={Chow, C.},
  journal={IEEE Transactions on information theory (TIT)},
  volume={16},
  number={1},
  pages={41--46},
  year={1970},
  publisher={IEEE}
}

@article{el2010foundations,
  author  = {Ran El-Yaniv and Yair Wiener},
  title   = {On the Foundations of Noise-free Selective Classification},
  journal = {Journal of Machine Learning Research (JMLR)},
  year    = {2010},
  volume  = {11},
  number  = {53},
  pages   = {1605--1641},
}

@article{geifman2017selective,
  title={Selective classification for deep neural networks},
  author={Geifman, Yonatan and El-Yaniv, Ran},
  journal={NeurIPS},
  year={2017}
}

@inproceedings{wang2018towards,
  title={Towards realistic predictors},
  author={Wang, Pei and Vasconcelos, Nuno},
  booktitle={ECCV},
  year={2018}
}

@inproceedings{whitehead2022reliable,
  title={Reliable visual question answering: Abstain rather than answer incorrectly},
  author={Whitehead, Spencer and Petryk, Suzanne and Shakib, Vedaad and Gonzalez, Joseph and Darrell, Trevor and Rohrbach, Anna and Rohrbach, Marcus},
  booktitle={ECCV},
  year={2022},
}

@inproceedings{dancette2023improving,
  title={Improving selective visual question answering by learning from your peers},
  author={Dancette, Corentin and Whitehead, Spencer and Maheshwary, Rishabh and Vedantam, Ramakrishna and Scherer, Stefan and Chen, Xinlei and Cord, Matthieu and Rohrbach, Marcus},
  booktitle={CVPR},
  year={2023}
}

@inproceedings{geifman2019selectivenet,
  title={Selectivenet: A deep neural network with an integrated reject option},
  author={Geifman, Yonatan and El-Yaniv, Ran},
  booktitle={ICML},
  year={2019},
}

@inproceedings{black2022selective,
title={Selective Ensembles for Consistent Predictions},
author={Emily Black and Klas Leino and Matt Fredrikson},
booktitle={ICLR},
year={2022},
}

@inproceedings{radford2021learning,
  title={Learning transferable visual models from natural language supervision},
  author =       {Radford, Alec and Kim, Jong Wook and Hallacy, Chris and Ramesh, Aditya and Goh, Gabriel and Agarwal, Sandhini and Sastry, Girish and Askell, Amanda and Mishkin, Pamela and Clark, Jack and Krueger, Gretchen and Sutskever, Ilya},  
    booktitle={ICML},
  year={2021},
}

@article{hessel2021clipscore,
  title={Clipscore: A reference-free evaluation metric for image captioning},
  author={Hessel, Jack and Holtzman, Ari and Forbes, Maxwell and Bras, Ronan Le and Choi, Yejin},
  journal={arXiv preprint arXiv:2104.08718},
  year={2021}
}

@inproceedings{li2022blip,
  title={Blip: Bootstrapping language-image pre-training for unified vision-language understanding and generation},
  author={Li, Junnan and Li, Dongxu and Xiong, Caiming and Hoi, Steven},
  booktitle={ICML},
  year={2022},
}

@inproceedings{zeng2024meacap,
  title={MeaCap: Memory-Augmented Zero-shot Image Captioning},
  author={Zeng, Zequn and Xie, Yan and Zhang, Hao and Chen, Chiyu and Chen, Bo and Wang, Zhengjue},
  booktitle={CVPR},
  year={2024}
}

@inproceedings{ma2023crepe,
  title={Crepe: Can vision-language foundation models reason compositionally?},
  author={Ma, Zixian and Hong, Jerry and Gul, Mustafa Omer and Gandhi, Mona and Gao, Irena and Krishna, Ranjay},
  booktitle={CVPR},
  year={2023}
}

@inproceedings{hsieh2024sugarcrepe,
  title={Sugarcrepe: Fixing hackable benchmarks for vision-language compositionality},
  author={Hsieh, Cheng-Yu and Zhang, Jieyu and Ma, Zixian and Kembhavi, Aniruddha and Krishna, Ranjay},
  booktitle={NeurIPS},
  volume={36},
  year={2024}
}

@inproceedings{thrush2022winoground,
  title={Winoground: Probing vision and language models for visio-linguistic compositionality},
  author={Thrush, Tristan and Jiang, Ryan and Bartolo, Max and Singh, Amanpreet and Williams, Adina and Kiela, Douwe and Ross, Candace},
  booktitle={CVPR},
  year={2022}
}

@inproceedings{stevens2024bioclip,
  title={Bioclip: A vision foundation model for the tree of life},
    author    = {Stevens, Samuel and Wu, Jiaman and Thompson, Matthew J and Campolongo, Elizabeth G and Song, Chan Hee and Carlyn, David Edward and Dong, Li and Dahdul, Wasila M and Stewart, Charles and Berger-Wolf, Tanya and Chao, Wei-Lun and Su, Yu},
  booktitle={CVPR},
  year={2024}
}

@inproceedings{zhou2022conditional,
  title={Conditional prompt learning for vision-language models},
  author={Zhou, Kaiyang and Yang, Jingkang and Loy, Chen Change and Liu, Ziwei},
  booktitle={CVPR},
  year={2022}
}

@article{zhou2022learning,
  title={Learning to prompt for vision-language models},
  author={Zhou, Kaiyang and Yang, Jingkang and Loy, Chen Change and Liu, Ziwei},
  journal={International Journal of Computer Vision (IJCV)},
  volume={130},
  number={9},
  pages={2337--2348},
  year={2022},
  publisher={Springer}
}

@inproceedings{wu2020solving,
  title={Solving long-tailed recognition with deep realistic taxonomic classifier},
  author={Wu, Tz-Ying and Morgado, Pedro and Wang, Pei and Ho, Chih-Hui and Vasconcelos, Nuno},
  booktitle={ECCV},
  year={2020},
}

@inproceedings{wang2023clipn,
  title={Clipn for zero-shot ood detection: Teaching clip to say no},
  author={Wang, Hualiang and Li, Yi and Yao, Huifeng and Li, Xiaomeng},
  booktitle={ICCV},
  year={2023}
}

@article{wang2024comprehensive,
  title={A comprehensive survey of continual learning: theory, method and application},
  author={Wang, Liyuan and Zhang, Xingxing and Su, Hang and Zhu, Jun},
  journal={IEEE Transactions on Pattern Analysis and Machine Intelligence (TPAMI)},
  year={2024},
  publisher={IEEE}
}

@misc{ilharco_gabriel_2021_5143773,
  author       = {Ilharco, Gabriel and
                  Wortsman, Mitchell and
                  Wightman, Ross and
                  Gordon, Cade and
                  Carlini, Nicholas and
                  Taori, Rohan and
                  Dave, Achal and
                  Shankar, Vaishaal and
                  Namkoong, Hongseok and
                  Miller, John and
                  Hajishirzi, Hannaneh and
                  Farhadi, Ali and
                  Schmidt, Ludwig},
  title        = {OpenCLIP},
  month        = jul,
  year         = 2021,
  publisher    = {Zenodo},
  version      = {0.1},
  doi          = {10.5281/zenodo.5143773},
  url          = {https://doi.org/10.5281/zenodo.5143773}
}

@article{chow1957optimum,
  title={An optimum character recognition system using decision functions},
  author={Chow, Chi-Keung},
  journal={IRE Transactions on Electronic Computers},
  volume={4},
  pages={247--254},
  year={1957},
  publisher={IEEE}
}

@inproceedings{
geifman2018biasreduced,
title={Bias-Reduced Uncertainty Estimation for Deep Neural Classifiers},
author={Yonatan Geifman and Guy Uziel and Ran El-Yaniv},
booktitle={ICLR},
year={2019},
}

@inproceedings{vedantam2015cider,
  title={Cider: Consensus-based image description evaluation},
  author={Vedantam, Ramakrishna and Lawrence Zitnick, C and Parikh, Devi},
  booktitle={CVPR},
  year={2015}
}

@inproceedings{banerjee2005meteor,
  title={METEOR: An automatic metric for MT evaluation with improved correlation with human judgments},
  author={Banerjee, Satanjeev and Lavie, Alon},
  booktitle={Proceedings of the acl workshop on intrinsic and extrinsic evaluation measures for machine translation and/or summarization},
  pages={65--72},
  year={2005}
}

@inproceedings{guo2017calibration,
  title={On calibration of modern neural networks},
  author={Guo, Chuan and Pleiss, Geoff and Sun, Yu and Weinberger, Kilian Q},
  booktitle={ICML},
  year={2017},}

@article{de2000reject,
  title={To reject or not to reject: that is the question-an answer in case of neural classifiers},
  author={De Stefano, Claudio and Sansone, Carlo and Vento, Mario},
  journal={IEEE Transactions on Systems, Man, and Cybernetics, Part C (Applications and Reviews)},
  volume={30},
  number={1},
  pages={84--94},
  year={2000},
  publisher={IEEE}
}

@inproceedings{lin2014microsoft,
  title={Microsoft coco: Common objects in context},
  author={Lin, Tsung-Yi and Maire, Michael and Belongie, Serge and Hays, James and Perona, Pietro and Ramanan, Deva and Doll{\'a}r, Piotr and Zitnick, C Lawrence},
  booktitle={ECCV},
  year={2014},
}

@inproceedings{PyTorch,
title = {PyTorch: An Imperative Style, High-Performance Deep Learning Library},
author = {Paszke, Adam and Gross, Sam and Massa, Francisco and Lerer, Adam and Bradbury, James and Chanan, Gregory and Killeen, Trevor and Lin, Zeming and Gimelshein, Natalia and Antiga, Luca and Desmaison, Alban and Kopf, Andreas and Yang, Edward and DeVito, Zachary and Raison, Martin and Tejani, Alykhan and Chilamkurthy, Sasank and Steiner, Benoit and Fang, Lu and Bai, Junjie and Chintala, Soumith},
booktitle = {NeurIPS},
year = {2019}
}

@article{lewis2020retrieval,
  title={Retrieval-augmented generation for knowledge-intensive nlp tasks},
  author={Lewis, Patrick and Perez, Ethan and Piktus, Aleksandra and Petroni, Fabio and Karpukhin, Vladimir and Goyal, Naman and K{\"u}ttler, Heinrich and Lewis, Mike and Yih, Wen-tau and Rockt{\"a}schel, Tim and others},
  journal={Advances in neural information processing systems},
  volume={33},
  pages={9459--9474},
  year={2020}
}

@inproceedings{ding2021support,
  title={Support-set based cross-supervision for video grounding},
  author={Ding, Xinpeng and Wang, Nannan and Zhang, Shiwei and Cheng, De and Li, Xiaomeng and Huang, Ziyuan and Tang, Mingqian and Gao, Xinbo},
  booktitle={Proceedings of the IEEE/CVF International Conference on Computer Vision},
  pages={11573--11582},
  year={2021}
}

@article{yarom2023you,
  title={What you see is what you read? improving text-image alignment evaluation},
  author={Yarom, Michal and Bitton, Yonatan and Changpinyo, Soravit and Aharoni, Roee and Herzig, Jonathan and Lang, Oran and Ofek, Eran and Szpektor, Idan},
  journal={Advances in Neural Information Processing Systems},
  volume={36},
  pages={1601--1619},
  year={2023}
}

@inproceedings{lin2024evaluating,
  title={Evaluating text-to-visual generation with image-to-text generation},
  author={Lin, Zhiqiu and Pathak, Deepak and Li, Baiqi and Li, Jiayao and Xia, Xide and Neubig, Graham and Zhang, Pengchuan and Ramanan, Deva},
  booktitle={European Conference on Computer Vision},
  pages={366--384},
  year={2024},
  organization={Springer}
}

@inproceedings{plummer2015flickr30k,
  title={Flickr30k entities: Collecting region-to-phrase correspondences for richer image-to-sentence models},
  author={Plummer, Bryan A and Wang, Liwei and Cervantes, Chris M and Caicedo, Juan C and Hockenmaier, Julia and Lazebnik, Svetlana},
  booktitle={Proceedings of the IEEE international conference on computer vision},
  pages={2641--2649},
  year={2015}
}

@article{nilsback2008102,
  title={102 category flower dataset},
  author={Nilsback, Maria-Elena and Zisserman, Andrew},
  journal={Department of Engineering Science, University of Oxford. Retrieved March},
  volume={4},
  pages={2024},
  year={2008}
}

@article{soomro2012ucf101,
  title={UCF101: A dataset of 101 human actions classes from videos in the wild},
  author={Soomro, Khurram and Zamir, Amir Roshan and Shah, Mubarak},
  journal={arXiv preprint arXiv:1212.0402},
  year={2012}
}

@inproceedings{changpinyo2021conceptual,
  title={Conceptual 12m: Pushing web-scale image-text pre-training to recognize long-tail visual concepts},
  author={Changpinyo, Soravit and Sharma, Piyush and Ding, Nan and Soricut, Radu},
  booktitle={CVPR},
  year={2021}
}

@inproceedings{zhai2023sigmoid,
  title={Sigmoid loss for language image pre-training},
  author={Zhai, Xiaohua and Mustafa, Basil and Kolesnikov, Alexander and Beyer, Lucas},
  booktitle={ICCV},
  year={2023}
}

@article{fang2024eva,
  title={Eva-02: A visual representation for neon genesis},
  author={Fang, Yuxin and Sun, Quan and Wang, Xinggang and Huang, Tiejun and Wang, Xinlong and Cao, Yue},
  journal={Image and Vision Computing},
  volume={149},
  pages={105171},
  year={2024},
  publisher={Elsevier}
}

@article{geirhos2024towards,
  title={Towards flexible perception with visual memory},
  author={Geirhos, Robert and Jaini, Priyank and Stone, Austin and Medapati, Sourabh and Yi, Xi and Toderici, George and Ogale, Abhijit and Shlens, Jonathon},
  journal={arXiv preprint arXiv:2408.08172},
  year={2024}
}

@article{silva2024learning,
  title={Learning from memory: Non-parametric memory augmented self-supervised learning of visual features},
  author={Silva, Thalles and Pedrini, Helio and Rivera, Ad{\'\i}n Ram{\'\i}rez},
  journal={ICML},
  year={2024}
}

@inproceedings{nakata2022revisiting,
  title={Revisiting a knn-based image classification system with high-capacity storage},
  author={Nakata, Kengo and Ng, Youyang and Miyashita, Daisuke and Maki, Asuka and Lin, Yu-Chieh and Deguchi, Jun},
  booktitle={ECCV},
  year={2022},
}

@inproceedings{irvin2019chexpert,
  title={Chexpert: A large chest radiograph dataset with uncertainty labels and expert comparison},
  author={Irvin, Jeremy and Rajpurkar, Pranav and Ko, Michael and Yu, Yifan and Ciurea-Ilcus, Silviana and Chute, Chris and Marklund, Henrik and Haghgoo, Behzad and Ball, Robyn and Shpanskaya, Katie and others},
  booktitle={Proceedings of the AAAI conference on artificial intelligence},
  volume={33},
  pages={590--597},
  year={2019}
}

@article{johnson2019mimic,
  title={MIMIC-CXR, a de-identified publicly available database of chest radiographs with free-text reports},
  author={Johnson, Alistair EW and Pollard, Tom J and Berkowitz, Seth J and Greenbaum, Nathaniel R and Lungren, Matthew P and Deng, Chih-ying and Mark, Roger G and Horng, Steven},
  journal={Scientific data},
  volume={6},
  number={1},
  pages={317},
  year={2019},
  publisher={Nature Publishing Group UK London}
}

@article{zhang2023biomedclip,
  title={Biomedclip: a multimodal biomedical foundation model pretrained from fifteen million scientific image-text pairs},
  author={Zhang, Sheng and Xu, Yanbo and Usuyama, Naoto and Xu, Hanwen and Bagga, Jaspreet and Tinn, Robert and Preston, Sam and Rao, Rajesh and Wei, Mu and Valluri, Naveen and others},
  journal={arXiv preprint arXiv:2303.00915},
  year={2023}
}

@article{ordonez2011im2text,
  title={Im2text: Describing images using 1 million captioned photographs},
  author={Ordonez, Vicente and Kulkarni, Girish and Berg, Tamara},
  journal={NurIPS},
  year={2011}
}

@article{kamath2023s,
  title={What's" up" with vision-language models? investigating their struggle with spatial reasoning},
  author={Kamath, Amita and Hessel, Jack and Chang, Kai-Wei},
  journal={arXiv preprint arXiv:2310.19785},
  year={2023}
}

@article{zhao2022vl,
  title={Vl-checklist: Evaluating pre-trained vision-language models with objects, attributes and relations},
  author={Zhao, Tiancheng and Zhang, Tianqi and Zhu, Mingwei and Shen, Haozhan and Lee, Kyusong and Lu, Xiaopeng and Yin, Jianwei},
  journal={arXiv preprint arXiv:2207.00221},
  year={2022}
}

@inproceedings{shekhar2017foil_acl,
   title={"FOIL it! Find One mismatch between Image and Language caption"},
   author={Shekhar, Ravi and Pezzelle, Sandro and Klimovich, Yauhen and Herbelot, Aurelie and Nabi, Moin and Sangineto, Enver and Bernardi, Raffaella},
   booktitle = {Proceedings of the 55th Annual Meeting of the Association for Computational Linguistics (ACL) (Volume 1: Long Papers)},
   pages = {255--265},
   year={2017}
}

@article{iscen2023retrieval,
  title={Retrieval-enhanced contrastive vision-text models},
  author={Iscen, Ahmet and Caron, Mathilde and Fathi, Alireza and Schmid, Cordelia},
  journal={arXiv preprint arXiv:2306.07196},
  year={2023}
}

@article{gu2025bioclip,
  title={Bioclip 2: Emergent properties from scaling hierarchical contrastive learning},
  author={Gu, Jianyang and Stevens, Samuel and Campolongo, Elizabeth G and Thompson, Matthew J and Zhang, Net and Wu, Jiaman and Kopanev, Andrei and Mai, Zheda and White, Alexander E and Balhoff, James and others},
  journal={arXiv preprint arXiv:2505.23883},
  year={2025}
}

@article{bai2023qwen,
  title={Qwen technical report},
  author={Bai, Jinze and Bai, Shuai and Chu, Yunfei and Cui, Zeyu and Dang, Kai and Deng, Xiaodong and Fan, Yang and Ge, Wenbin and Han, Yu and Huang, Fei and others},
  journal={arXiv preprint arXiv:2309.16609},
  year={2023}
}

@article{liu2023visual,
  title={Visual instruction tuning},
  author={Liu, Haotian and Li, Chunyuan and Wu, Qingyang and Lee, Yong Jae},
  journal={NeurIPS},
  year={2023}
}

@article{ruckert2024rocov2,
  title={Rocov2: Radiology objects in context version 2, an updated multimodal image dataset},
  author={R{\"u}ckert, Johannes and Bloch, Louise and Br{\"u}ngel, Raphael and Idrissi-Yaghir, Ahmad and Sch{\"a}fer, Henning and Schmidt, Cynthia S and Koitka, Sven and Pelka, Obioma and Abacha, Asma Ben and G. Seco de Herrera, Alba and others},
  journal={Scientific Data},
  volume={11},
  number={1},
  pages={688},
  year={2024},
  publisher={Nature Publishing Group UK London}
}

@article{de2025padchest,
  title={Padchest-gr: A bilingual chest X-ray dataset for grounded radiology report generation},
  author={de Castro, Daniel Coelho and Bustos, Aurelia and Bannur, Shruthi and Hyland, Stephanie L and Bouzid, Kenza and Wetscherek, Maria Teodora and S{\'a}nchez-Valverde, Maria Dolores and Jaques-P{\'e}rez, Lara and P{\'e}rez-Rodr{\'\i}guez, Lourdes and Takeda, Kenji and others},
  journal={NEJM AI},
  volume={2},
  number={7},
  pages={AIdbp2401120},
  year={2025},
  publisher={Massachusetts Medical Society}
}

@article{shuja2020covid,
  title={COVID-19 datasets: a survey and future challenges},
  author={Shuja, Junaid and Alanazi, Eisa and Alasmary, Waleed and Alashaikh, Abdulaziz},
  journal={MedRxiv},
  pages={2020--05},
  year={2020},
  publisher={Cold Spring Harbor Laboratory Press}
}

@article{colak2021rsna,
  title={The RSNA pulmonary embolism CT dataset},
  author={Colak, Errol and Kitamura, Felipe C and Hobbs, Stephen B and Wu, Carol C and Lungren, Matthew P and Prevedello, Luciano M and Kalpathy-Cramer, Jayashree and Ball, Robyn L and Shih, George and Stein, Anouk and others},
  journal={Radiology: Artificial Intelligence},
  volume={3},
  number={2},
  pages={e200254},
  year={2021},
  publisher={Radiological Society of North America}
}

@article{dong2024can,
  title={Can llm be a personalized judge?},
  author={Dong, Yijiang River and Hu, Tiancheng and Collier, Nigel},
  journal={arXiv preprint arXiv:2406.11657},
  year={2024}
}

@inproceedings{ye2025learning,
  title={Learning LLM-as-a-judge for preference alignment},
  author={Ye, Ziyi and Li, Xiangsheng and Li, Qiuchi and Ai, Qingyao and Zhou, Yujia and Shen, Wei and Yan, Dong and Liu, Yiqun},
  booktitle={ICLR},
  year={2025}
}

@inproceedings{iscen2023improving,
  title={Improving image recognition by retrieving from web-scale image-text data},
  author={Iscen, Ahmet and Fathi, Alireza and Schmid, Cordelia},
  booktitle={CVPR},
  year={2023}
}

@inproceedings{xie2023ra,
  title={Ra-clip: Retrieval augmented contrastive language-image pre-training},
  author={Xie, Chen-Wei and Sun, Siyang and Xiong, Xiong and Zheng, Yun and Zhao, Deli and Zhou, Jingren},
  booktitle={CVPR},
  year={2023}
}

@article{srinivasan2024selective,
  title={Selective" selective prediction": Reducing unnecessary abstention in vision-language reasoning},
  author={Srinivasan, Tejas and Hessel, Jack and Gupta, Tanmay and Lin, Bill Yuchen and Choi, Yejin and Thomason, Jesse and Chandu, Khyathi Raghavi},
  journal={arXiv preprint arXiv:2402.15610},
  year={2024}
}

@article{liu2023react,
  author      = {Liu, Haotian and Son, Kilho and Yang, Jianwei and Liu, Ce and Gao, Jianfeng and Lee, Yong Jae and Li, Chunyuan},
  title       = {Learning Customized Visual Models with Retrieval-Augmented Knowledge},
  journal   = {arXiv:2301.07094},
  year        = {2023},
}

@article{ghosh2023geneval,
  title={Geneval: An object-focused framework for evaluating text-to-image alignment},
  author={Ghosh, Dhruba and Hajishirzi, Hannaneh and Schmidt, Ludwig},
  journal={NeurIPS},
  year={2023}
}

@inproceedings{wang2019aspect,
  title={Aspect-ratio-preserving multi-patch image aesthetics score prediction},
  author={Wang, Lijie and Wang, Xueting and Yamasaki, Toshihiko and Aizawa, Kiyoharu},
  booktitle={CVPRW},
  year={2019}
}

@article{traub2024overcoming,
  title={Overcoming common flaws in the evaluation of selective classification systems},
  author={Traub, Jeremias and Bungert, Till J and L{\"u}th, Carsten T and Baumgartner, Michael and Maier-Hein, Klaus H and Maier-Hein, Lena and J{\"a}ger, Paul F},
  journal={NeurIPS},
  year={2024}
}

@article{douze2024faiss,
  title={The faiss library},
  author={Douze, Matthijs and Guzhva, Alexandr and Deng, Chengqi and Johnson, Jeff and Szilvasy, Gergely and Mazar{\'e}, Pierre-Emmanuel and Lomeli, Maria and Hosseini, Lucas and J{\'e}gou, Herv{\'e}},
  journal={arXiv preprint arXiv:2401.08281},
  year={2024}
}

@inproceedings{guo2020accelerating,
  title={Accelerating large-scale inference with anisotropic vector quantization},
  author={Guo, Ruiqi and Sun, Philip and Lindgren, Erik and Geng, Quan and Simcha, David and Chern, Felix and Kumar, Sanjiv},
  booktitle={ICML},
  year={2020},
}

@article{miller1995wordnet,
  title={WordNet: a lexical database for English},
  author={Miller, George A},
  journal={Communications of the ACM},
  volume={38},
  number={11},
  pages={39--41},
  year={1995},
  publisher={ACM New York, NY, USA}
}

@article{zhu2025internvl3,
  title={Internvl3: Exploring advanced training and test-time recipes for open-source multimodal models},
  author={Zhu, Jinguo and Wang, Weiyun and Chen, Zhe and Liu, Zhaoyang and Ye, Shenglong and Gu, Lixin and Tian, Hao and Duan, Yuchen and Su, Weijie and Shao, Jie and others},
  journal={arXiv preprint arXiv:2504.10479},
  year={2025}
}

@inproceedings{cheng2022calibrating,
  title={Calibrating deep neural networks by pairwise constraints},
  author={Cheng, Jiacheng and Vasconcelos, Nuno},
  booktitle={CVPR},
  year={2022}
}
\bibliographystyle{iclr2026_conference}

\clearpage
\appendix

\setcounter{page}{1}

\counterwithin{figure}{section}
\counterwithin{table}{section}
\numberwithin{equation}{section}
\counterwithin{algorithm}{section}


The appendix is structured as follows - 

\begin{multicols}{2}
\begin{enumerate}
    \item Additional Implementation Details   
    \begin{itemize}
        \item Evaluation sets
        \item Implementation
        \item Algorithm
    \end{itemize}

    \item Comparison with baselines
    \begin{itemize}
        \item \ours with SP-VLMs
        \item Selective Predictors 
        \item Image-text alignment modules
        \item LVLM-as-a-judge
        \item SP-VLM Finetuning
        \item Per-sample Inference Latency
        \item Different Metrics
        \item Out-of-distribution detection
    \end{itemize}
    
    \item Ablations
    \begin{itemize}
        \item Neighborhood size
        \item Size of Retrieval Set
        \item Retrieval variants
        \item Effect of relevance threshold
        \item Contrastive component (CC)
    \end{itemize}
    
    \item More related works
    \begin{itemize}
        \item Foundation Models for Scoring
    \end{itemize}
    
    \item Qualitative results 
\end{enumerate}
\end{multicols}

\section{Additional Implementation Details}
\label{sec:moreimplement}
In this section, we discuss additional implementation details of \ours and \oursma.

\noindent{\bf Evaluation Sets.} Performance is evaluated on the test splits of the following datasets.
\begin{itemize}
    \item \noindent{\it Selective classification:} Oxford Pets, Oxford Flowers~\citep{nilsback2008102} and UCF-101~\citep{soomro2012ucf101}.
    
    \item \noindent{\it Selective image-text matching.} SugarCrepe \citep{hsieh2024sugarcrepe}, WinoGround \citep{thrush2022winoground}, What'sUp \citep{kamath2023s}, VL-Checklist \citep{zhao2022vl} and Foil \citep{shekhar2017foil_acl}.

    \item \noindent{\it Selective captioning:} MS-COCO \citep{lin2014microsoft} and Flickr-30K \citep{plummer2015flickr30k}. 
\end{itemize}

\noindent{\bf Implementation.} All experiments implemented using PyTorch~\citep{PyTorch} and run on a single NVIDIA RTX 3090 GPU, using the pre-trained VLMs trained by OpenCLIP~\citep{ilharco_gabriel_2021_5143773}. The source codes of \oursma and \ours will be released upon publication. 

\noindent{\bf Algorithm.} 
The algorithm of \oursma is summarized in Algorithm~\ref{alg:algorithm1}. 

\begin{algorithm}[hbt!]
	\caption{\oursma} 
	\begin{algorithmic}[1]
            \State \textbf{Input:} image-caption pair \textbf{z} = \{\textbf{v}, \textbf{w}\}, retrieval set $\mathcal{R}$ = \{\textbf{x}$_i$, \textbf{y}$_i$\}$_{i=1}^N$
            \State \textbf{Output:} score $\emph{s}$
            \State Compute $\mathcal{N}(\textbf{v})$ using \eqref{eq:Nbg}.
            \State Compute proxy embedding $\Tilde{\textbf{w}}(\textbf{v})$ using \eqref{eq:wtilde_txt}
            \If {negative captions are present}
                \State Compute \emph{s} using \eqref{eq:ipsc}
            \Else
                \State Generate $\mathcal{E}$ for $f(\mathbf{x})$ using LLM
                \State Compute \emph{s} using \eqref{eq:ipsc}
            \EndIf
            \State return \emph{s}
	\end{algorithmic} 
    \label{alg:algorithm1}
\end{algorithm}

\begin{figure*}[t!]\RawFloats
\setlength{\tabcolsep}{2pt}
\centering
\scriptsize
\begin{tabular}{l|c|ccc|cccc|ccccc}
\toprule
\multirow{2}{*}{} & & \multicolumn{3}{c|}{Classification} & \multicolumn{4}{c|}{Captioning} & \multicolumn{5}{|c}{ITM} \\
\multirow{2}{*}{Model} & \multirow{2}{*}{MA} & \multirow{2}{*}{Flowers} & \multirow{2}{*}{Pets} & \multirow{2}{*}{UCF101} &\multicolumn{2}{c}{MS-COCO} & \multicolumn{2}{c|}{Flickr-30K} & \multirow{2}{*}{SugarCrepe} & \multirow{2}{*}{WinoGround} & \multirow{2}{*}{What'sUp} & \multirow{2}{*}{VL-Check} & \multirow{2}{*}{Foil}    \\
 & & & & & Cider-N & Meteor & Cider-N & Meteor & & & & &  \\
\midrule
\multicolumn{14}{c}{\bf Small Models} \\
\midrule
\multirow{2}{*}{$\text{OpenCLIP}_{\text{B/16}}$}  &  & 0.160 & 0.231 & 0.213 & 0.153 & 0.320 & 0.241 & 0.316 & 0.173 & 0.322 & 0.292 & 0.266 & 0.257 \\
 & \checkmark & \textbf{0.111} & \textbf{0.137} & \textbf{0.133} & \textbf{0.130} & \textbf{0.316} & \textbf{0.238} & \textbf{0.313} & \textbf{0.123} & \textbf{0.248} & \textbf{0.267} & \textbf{0.249} & \textbf{0.242} \\
\midrule
\multirow{2}{*}{$\text{EVA02}_{\text{B/16}}$}  & & 0.109 & 0.220 & 0.206 & 0.139 & 0.318 & 0.240 & 0.317 & 0.171 & 0.264 & 0.272 & 0.262 & 0.251 \\
 & \checkmark & \textbf{0.109} & \textbf{0.135} & \textbf{0.127} & \textbf{0.132} & \textbf{0.318} & \textbf{0.235} & \textbf{0.312} & \textbf{0.108} & \textbf{0.236} & \textbf{0.264} & \textbf{0.233} & \textbf{0.238} \\
\midrule
\multirow{2}{*}{$\text{SigLIP}_{\text{B/16}}$} & & 0.093 & 0.211 & 0.154 & 0.142 & 0.317 & 0.237 & 0.311 & 0.162 & 0.258 & 0.262 & 0.258 & 0.245 \\
 & \checkmark & {\bf 0.077} & {\bf 0.171} & {\bf 0.116} & {\bf 0.121} & {\bf 0.312} & {\bf 0.235} & {\bf 0.310}  & {\bf 0.079} & {\bf 0.228} & {\bf 0.214} & {\bf 0.210} & {\bf 0.213} \\
\midrule
\multicolumn{14}{c}{\bf Large Models} \\
\midrule
\multirow{2}{*}{$\text{OpenCLIP}_{\text{L/14}}$} & & 0.119 & 0.182 & 0.151 & 0.146 & 0.318 & 0.240 & 0.312 & 0.149 & 0.258 & 0.266 & 0.254 & 0.246 \\
 & \checkmark & \textbf{0.109} & \textbf{0.128} & \textbf{0.141} & \textbf{0.138} & \textbf{0.297} & \textbf{0.235} & \textbf{0.298} & \textbf{0.127} & \textbf{0.208} & \textbf{0.226} & \textbf{0.217} & \textbf{0.198} \\
\midrule
\multirow{2}{*}{$\text{EVA02}_{\text{L/14}}$} & & \textbf{0.102} & 0.175 & 0.126 & \textbf{0.134} & 0.313 & 0.237 & 0.313 & 0.128 & 0.242 & 0.248 & 0.232 & 0.229  \\
 & \checkmark & 0.106 & \textbf{0.117} & \textbf{0.118} & 0.121 & \textbf{0.284} & \textbf{0.228} &  \textbf{0.295} & \textbf{0.128} & \textbf{0.216} & \textbf{0.216} & \textbf{0.196} & \textbf{0.199}  \\
\midrule
\multirow{2}{*}{$\text{SigLIP}_{\text{SO-400M}}$} & & 0.074 & 0.169 & 0.113 & 0.136 & 0.311 & 0.229 & 0.307  & 0.078 & 0.208 & 0.202 & 0.204 & 0.197 \\
 & \checkmark & \textbf{0.063} & \textbf{0.114} & \textbf{0.088} & \textbf{0.109} & \textbf{0.286} & \textbf{0.219} & \textbf{ 0.297} & \textbf{0.062} & \textbf{0.196} &  \textbf{0.192} & \textbf{0.192} & \textbf{0.189} \\
\bottomrule
\end{tabular}
\captionof{table}{AURC ($\downarrow$) performance of \ours with and without memory augmentation (MA) for different visual language representation models.}
\label{tab:aurc-cls-supp-clip}
\end{figure*}

\section{Comparison with Baselines}
\label{sec:baselines}
In this section, we compare  \oursma to several baselines, including implementations of \ours with different SP-VLMs (OpenCLIP~\citep{radford2021learning}, EVA02~\citep{fang2024eva}, SigLIP~\citep{zhai2023sigmoid}), other image-text alignment modules (GenEval~\citep{ghosh2023geneval}, Aesthetics score~\citep{wang2019aspect}), retrieval-augmented CLIP (REACT~\citep{liu2023react}), and finetuned SP-VLMs (SigLIP$_{\text{B/16}}$, SigLIP$_{\text{SO-400M}}$). Beyond this, we also consider models designed for selective prediction (LYP~\citep{dancette2023improving}, ReCoVERR~\citep{srinivasan2024selective}),  and LVLM-as-a-judge (Qwen-2.5-VL). Additionally, we analyze the inference latency of all methods. All experiments were conducted using the setup described in the main paper.

\paragraph{\ours with SP-VLMs.} 
Table \ref{tab:aurc-cls-supp-clip} compares \oursma to \ours for different SP-VLMs, for all three tasks described in the main paper. The table supports the following conclusions. First, \oursma outperforms \ours consistently across tasks and datasets for both small and large SP-VLMs. Second, the gains of \oursma are larger for the larger SP-VLMs. This suggests that there is no saturation, \ie even the largest VLRMs can benefit from the increase in representation stability and calibration afforded by \oursma. 



\begin{figure*}[hbt!]\RawFloats
\setlength{\tabcolsep}{2pt}
\centering
\scriptsize
\begin{tabular}{l|ccc|cccc|ccccc}
\toprule
\multirow{2}{*}{} & \multicolumn{3}{c|}{Classification} & \multicolumn{4}{c|}{Captioning} & \multicolumn{5}{|c}{ITM} \\
\multirow{2}{*}{Methods} & \multirow{2}{*}{Flowers} & \multirow{2}{*}{Pets} & \multirow{2}{*}{UCF101} &\multicolumn{2}{c}{MS-COCO} & \multicolumn{2}{c|}{Flickr-30K} & \multirow{2}{*}{SugarCrepe} & \multirow{2}{*}{WinoGround} & \multirow{2}{*}{What'sUp} & \multirow{2}{*}{VL-Checklist} & \multirow{2}{*}{Foil}    \\
 & & & & Cider-N & Meteor & Cider-N & Meteor & & & & &  \\
\midrule
LYP & 0.216 & 0.221 & 0.226 & 0.152 & 0.342 & \underline{0.236} & 0.323 & \underline{0.158} & \underline{0.229} & \underline{0.228} & \underline{0.232} & \underline{0.225} \\
ReCoVERR & 0.225 & 0.216 & 0.223 & 0.142 & 0.336 & 0.253 & 0.327 & 0.167 & \underline{0.252} & \underline{0.249} & \underline{0.247} & \underline{0.242} \\
\midrule
\ours-S & 0.093 & 0.211 & 0.154 & 0.142 & 0.317 & 0.237 & 0.311 & 0.162 & 0.258 & 0.262 & 0.258 & 0.245 \\
\oursma-S & 0.077 & 0.171 & 0.116 & 0.121 & 0.312 & 0.235 & 0.310  & 0.079 & 0.228 & 0.214 & 0.210 & 0.213 \\
\midrule
\ours-L & 0.074 & 0.169 & 0.113 & 0.136 & 0.311 & 0.229 & 0.307  & 0.078 & 0.208 & 0.202 & 0.204 & 0.197 \\
\oursma-L & \textbf{0.063} & \textbf{0.114} & \textbf{0.088} & \textbf{0.109} & \textbf{0.286} & \textbf{0.219} & \textbf{0.297} & \textbf{0.062} & \textbf{0.196} & \textbf{0.192} & \textbf{0.192} & \textbf{0.189} \\
\bottomrule
\end{tabular}
\captionof{table}{AURC ($\downarrow$) performance of \ours and \oursma to selective predictors. 
}
\label{tab:aurc-cls-supp}
\end{figure*}

\paragraph{Selective Predictors.} 
Table \ref{tab:aurc-cls-supp} compares \oursma to methods designed specifically for selective prediction (SP): LYP ~\citep{dancette2023improving} and ReCoVERR ~\citep{srinivasan2024selective}. Phi-3.5-Instruct (3B) is used as the ReCoVERR LLM, and BLIP-2 is used as P-VLM for all methods. Shown in boldface 
are the overall best results, while the underlined results indicate where SP methods outperform \ours-S. \ours-L, and \oursma with both small and large model outperform both SP methods on 12/12 datasets.
The gains are largest for \oursma-L. For instance, \oursma with small model has an AURC gain of 21 points and 18 Cider-N points over ReCovERR on MS-COCO and Flickr-30K, respectively,
while for \oursma with the largest model the gains become of 33 and 34 points, respectively. Note that this happens despite the fact that the SP models are computationally more intensive than \oursma and \ours. 

\begin{figure*}[hbt!]\RawFloats
\setlength{\tabcolsep}{2pt}
\centering
\scriptsize
\begin{tabular}{l|ccc|cccc|ccccc}
\toprule
\multirow{2}{*}{} & \multicolumn{3}{c|}{Classification} & \multicolumn{4}{c|}{Captioning} & \multicolumn{5}{|c}{ITM} \\
\multirow{2}{*}{Methods} & \multirow{2}{*}{Flowers} & \multirow{2}{*}{Pets} & \multirow{2}{*}{UCF101} &\multicolumn{2}{c}{MS-COCO} & \multicolumn{2}{c|}{Flickr-30K} & \multirow{2}{*}{SugarCrepe} & \multirow{2}{*}{WinoGround} & \multirow{2}{*}{What'sUp} & \multirow{2}{*}{VL-Checklist} & \multirow{2}{*}{Foil}    \\
 & & & & Cider-N & Meteor & Cider-N & Meteor & & & & &  \\
\midrule
GenEval & 0.216 & 0.213 & 0.226 & 0.155 & 0.327 & 0.249 & 0.317 & \underline{0.149} & \underline{0.246} & \underline{0.240} &  \underline{0.234} & \underline{0.217} \\
Aesthetics & 0.212 & 0.211 & 0.225 & 0.155 & 0.322 & 0.242 & 0.318 & \underline{0.153} & \underline{0.241} & \underline{0.241} & \underline{0.227} & \underline{0.221} \\
VQAScore & 0.211 & \underline{0.207} & 0.217 & 0.146 & 0.321 & 0.241 & 0.312 & \underline{0.146} & \underline{0.240} & \underline{0.236} & \underline{0.226} & \underline{0.217} \\
SeeTRUE & 0.214 & 0.213 & 0.171 & 0.158 & \underline{0.316} & 0.251 & 0.411 & \underline{0.153} & \underline{0.242}  & \underline{0.236} & \underline{0.234} & \underline{0.223} \\
\midrule
\ours-S & 0.093 & 0.211 & 0.154 & 0.142 & 0.317 & 0.237 & 0.311 & 0.162 & 0.258 & 0.262 & 0.258 & 0.245 \\
\oursma-S & 0.077 & 0.171 & 0.116 & 0.121 & 0.312 & 0.235 & 0.310  & 0.079 & 0.228 & 0.214 & 0.210 & 0.213 \\
\midrule
\ours-L & 0.074 & 0.169 & 0.113 & 0.136 & 0.311 & 0.229 & 0.307  & 0.078 & 0.208 & 0.202 & 0.204 & 0.197 \\
\oursma-L & \textbf{0.063} & \textbf{0.114} & \textbf{0.088} & \textbf{0.109} & \textbf{0.286} & \textbf{0.219} & \textbf{0.297} & \textbf{0.062} & \textbf{0.196} & \textbf{0.192} & \textbf{0.192} & \textbf{0.189} \\
\bottomrule
\end{tabular}
\captionof{table}{AURC ($\downarrow$) for all three tasks. The table shows the comparison between \ours, \oursma and Image-text alignment modules.}
\label{tab:aurc-ita-supp}
\end{figure*}

\paragraph{Image-Text Alignment Modules.} 

Table \ref{tab:aurc-ita-supp} compares the performance of \ours and \oursma against state-of-the-art image-text alignment (ITA) models, including GenEval Score~\citep{ghosh2023geneval},  Aesthetics Score~\citep{wang2019aspect}, VQAScore~\citep{lin2024evaluating} and SeeTRUE~\citep{yarom2023you}. Boldface highlights the best-performing models, while the underlined results indicate where ITA methods outperform \ours-S. \ours-L and \oursma with both small and large model outperform both ITA methods on 12/12 datasets. The gains of the larger versions of \oursma over ITA baselines are also more pronounced than for smaller models. Specifically, \oursma with the largest model outperforms the GenEval score by 44 points on MS-COCO and 30 points on Flickr-30K, under the Cider-N setting. This gain holds despite the fact that the ITA methods are significantly more computationally expensive than \ours and \oursma, as they rely on either visual-question answering with large vision-language models (LVLMs) or object detection combined with large language models (LLMs) to identify the presence of objects mentioned in the prompt. This again highlights the effectiveness of \oursma.
\begin{figure*}[ht]\RawFloats
\centering
\scriptsize
\setlength{\tabcolsep}{1.5pt}
\begin{tabular}{c|cc|cc|ccc|ccccc}
\toprule
& \multicolumn{4}{c|}{Captioning} & \multicolumn{3}{c|}{Classification} & \multicolumn{5}{c}{ITM} \\
\multirow{3}{*}{LVLM-as-Judge} & \multicolumn{2}{c|}{MS-COCO} & \multicolumn{2}{c|}{Flickr-30K} & \multirow{2}{*}{Flowers} & \multirow{2}{*}{Pets} & \multirow{2}{*}{UCF-101} & \multirow{2}{*}{SugarCrepe} & \multirow{2}{*}{Winoground} & \multirow{2}{*}{What'sUp} & \multirow{2}{*}{VL-checklist} & \multirow{2}{*}{Foil} \\                   
& Cider-N & Meteor & Cider-N & Meteor  & & & \\  
\midrule
\multicolumn{13}{c}{LVLM-\ours} \\
\midrule
\multicolumn{1}{c|}{Qwen-VL-7B } & 0.120 & 0.301 & 0.233 & 0.312 & 0.075 & 0.126 & 0.099 & 0.073 & 0.209 & 0.205 & 0.205 & 0.202                 \\ 
\multicolumn{1}{c|}{Qwen-VL-72B } & 0.115 & 0.294 & 0.227 & 0.305 & 0.070 & 0.120 & 0.094 & 0.068 & 0.203 & 0.199 & 0.199 & 0.196               \\ 
\midrule
\multicolumn{13}{c}{\oursma} \\
\midrule
\multicolumn{1}{c|}{$\text{SigLIP}_{\text{SO-400M}}$} & \textbf{0.109} & \textbf{0.286} &  \textbf{0.219} & \textbf{0.297} & \textbf{0.063} & \textbf{0.114} & \textbf{0.088} & \textbf{0.062} & 0.196 &  0.192 & \textbf{0.192} & \textbf{0.189}\\
\bottomrule
\end{tabular}
\captionof{table}{Performance of \oursma as compared to LVLM-as-a-judge baselines.
}
\label{tab:ablation-lvlm}
\end{figure*}

\paragraph{LVLM-as-a-judge.} We implemented a version of LVLM-as-judge, referred to as LVLM-\ours, which is comparable to \ours and \oursma. In this setup, an LVLM is tasked with generating an alignment score between the query image and its corresponding caption. This score is subsequently thresholded to facilitate selective prediction. Table \ref{tab:ablation-lvlm} presents the results for both the smaller (7B parameters) and larger (72B parameters) LVLMs from the Qwen-2.5-VL family. \oursma outperforms LVLM-\ours across all datasets and tasks, for both the 7B and 72B parameter models. These findings confirm the hypotheiss that LVLMs are not well-calibrated to produce reliable confidence scores for selective prediction.


\begin{table}[ht]
    \centering
    \scriptsize
    \setlength{\tabcolsep}{2pt}
    \begin{tabular}{l|cc|cc|ccc|ccccc}
    \toprule
    & \multicolumn{4}{c|}{Captioning} & \multicolumn{3}{c|}{Classification} & \multicolumn{5}{c}{ITM} \\
\multirow{3}{*}{Model} & \multicolumn{2}{c|}{MS-COCO} & \multicolumn{2}{c|}{Flickr-30K} & \multirow{2}{*}{Flowers} & \multirow{2}{*}{Pets} & \multirow{2}{*}{UCF-101} & \multirow{2}{*}{SugarCrepe} & \multirow{2}{*}{Winoground} & \multirow{2}{*}{What'sUp} & \multirow{2}{*}{VL-checklist} & \multirow{2}{*}{Foil} \\                   
& Cider-N & Meteor & Cider-N & Meteor  & & & \\  
    \midrule
    \ours & 0.136 & 0.311 & 0.229 & 0.307 & 0.074 & 0.169 & 0.113 & 0.078 & 0.208 & 0.202 & 0.204 & 0.197 \\
    \ours-FT & \underline{0.115}   & \underline{0.278}   & \underline{0.224} & \underline{0.301} & 0.067 & \underline{0.074} & \textbf{0.071} & 0.074 & 0.192 & \underline{0.188} & \underline{0.202} &  \textbf{0.159} \\
    \midrule
    \oursma & 0.126   & 0.288  & 0.229 & 0.307 & 0.062   & 0.076  & 0.078 & 0.066   & 0.192  & 0.189   & 0.207 & 0.167 \\
    \oursma-FT & \textbf{0.104}   & \textbf{0.234}  & \textbf{0.217} & \textbf{0.288} & \textbf{0.060}   & \textbf{0.068} & 0.072 & \textbf{0.064}   & \textbf{0.188}  & \textbf{0.172}   & \textbf{0.201} & 0.162 \\
    \bottomrule
    \end{tabular}
    \caption{Performance of \oursma as compared to \ours and \ours with finetuned SigLIP$_{\text{SO-400M}}$. \oursma is implemented with SigLIP$_{\text{SO-400M}}$.}
    \label{tab:ft-clip-baseline}
\end{table}

\paragraph{SP-VLM Finetuning.} We evaluated the benefits of fine-tuning the SP-VLM on the in-domain dataset for both \ours and \oursma. In all experiments, the SP-VLM used was SigLIP$_{\text{SO-400M}}$, and the retrieval set for \oursma consisted of the in-domain dataset. The results are summarized in Table \ref{tab:ft-clip-baseline}, were boldface denotes the best results and underline indicates when \ours-FT outperforms \oursma without finetuning. A few conclusions are possible. First, while \ours-FT outperforms \oursma on 9/12 datasets, the differences are small: around 10 points on MS-COCO, around 5 points on Flickr, between 2 and 7 points for the classification tasks, and between 1 and 8 points for ITM. Note that \ours-FT, requires training the SP-VLM for each retrieval set, which is cumbersome and computationally intensive. For example, finetuning the SigLIP$_{\text{SO-400M}}$ VLRM on the MS-COCO train split using a single RTX-A4000 requires 20 hours of training. The fact that \oursma is competitive {\it without any training} is a sign of its effectiveness. Second, if the training data available for finetuning is small, the finetuned VLRM  can overfit leading to weak AURC. \oursma is more robust to small data sizes. For instance, in the ITM tasks, which tend to have small datasets (e.g. 1K samples for SugarCrepe), \oursma outperforms or is almost identical to \ours-FT on 3/5 datasets. Third, the best overall results are obtained with \ours-FT, which has the top performance in 10/12 datasets, and a loss of at most 3 points in the remaining 2/12. This shows that, even if a fine-tuned model is available (\ours-FT), \oursma is still beneficial. In fact the gains of \oursma-FT over \ours-FT can be larger than those of \oursma over \ours. For example, on Flickr-30K, \oursma has no gain over \ours but \oursma-FT has gains between 7 and 13 points over \ours-FT. In summary, the decision to finetune the SP-VLM must take into account whether the gains of \ours-FT over \oursma justify the additional training complexity. However, once the finetuned VLRM is available there is almost always a benefit in augmenting it with \oursma.

\begin{figure}[ht]\RawFloats
\centering
\footnotesize
\begin{tabular}{l|c}
\toprule
Methods & Time (ms) $\downarrow$ \\
\midrule
\ours-S/L & 14.07/14.11\\
\midrule
\multicolumn{2}{c}{Close-set Task (Classification, ITM)} \\
\midrule
\oursma-L (P) & 14.87\\
\midrule
\multicolumn{2}{c}{Open-set Task (Captioning)} \\
\midrule
\oursma-L (w RB) (P+C)  & 17.54\\
\oursma-L (w SLM) (P+C)  & 57.93\\
\oursma-L + PP (w SLM) (P+C) & 25.85\\
\bottomrule
\end{tabular}
\captionof{table}{Average inference times. P and C denote the two components of \oursma with RB or SLM, Proxy Embedding and Contrastive Score, respectively. PP denotes Pipeline parallelism over 2 GPUs.}
\label{tab:infer}
\end{figure}

\paragraph{Per-sample Inference Latency.} Table~\ref{tab:infer} presents a comparison of the inference time per sample across various methods. The top third of the table shows the inference times for \ours using both the small and large SP-VLMs, revealing that model size has minimal impact on latency. The middle third of the table displays the \oursma inference times for tasks, such as classification and ITM, that only require the computation of the proxy embedding (P). For these tasks, \oursma is comparable to \ours. The bottom third of the table shows the times for tasks like captioning, which involve both the proxy embedding and the contrastive score (P+C). This section highlights that the most time-consuming operation is the computation of contrastive scores, primarily due to the use of the LLM for generating hard negatives.
 

For captioning, \oursma is approximately three times slower than \ours. However, we note that several optimizations could significantly improve its speed. For instance, by employing pipeline parallelism across two RTX-3090 GPUs to parallelize the LLM and \oursma computations (denoted as \oursma-L+PP in the table), the overall inference time for \oursma can be reduced by more than half. Additionally, the nearest neighbor operation involved in proxy computation could be substantially accelerated using techniques such as Faiss~\citep{douze2024faiss}, ScaNN~\citep{guo2020accelerating}, or indexing structures from classical retrieval methods~\citep{lewis2020retrieval, iscen2023retrieval, xie2023ra}. The contrastive score computations could also be sped up significantly by replacing the LLM with rule-based sentence manipulations, as seen in prior work~\citep{wang2024comprehensive, shekhar2017foil_acl}. While this approach could lead to slight degradation in AURC performance, it remains an avenue for future investigation. It is important to note that the computation of the retrieval set embeddings is a one-time operation and is not included in the times reported above, consistent with standard retrieval methods. For reference, the per-sample embedding time (on RTX-3090) and storage for CLIP$_{\text{B/16}}$ are 10.361 ms and 2.274 KB, respectively.

\begin{figure*}[t!]\RawFloats
\setlength{\tabcolsep}{2pt}
\centering
\scriptsize
\begin{tabular}{l|cc|cc|ccc|ccccc}
\toprule
\multirow{2}{*}{} & \multicolumn{4}{c|}{Captioning} & \multicolumn{3}{c|}{Classification} & \multicolumn{5}{|c}{ITM} \\
\multirow{2}{*}{Methods} & \multicolumn{2}{c|}{MS-COCO} & \multicolumn{2}{c|}{Flickr-30K} & \multirow{2}{*}{Flowers} & \multirow{2}{*}{Pets} & \multirow{2}{*}{UCF101} & \multirow{2}{*}{SugarCrepe} & \multirow{2}{*}{WinoGround} & \multirow{2}{*}{What'sUp} & \multirow{2}{*}{VL-Checklist} & \multirow{2}{*}{Foil}    \\
 & Cider-N & Meteor & Cider-N & Meteor & & & & & & & &  \\
 \midrule
 \multicolumn{1}{c}{} & \multicolumn{12}{c}{AUGRC Metric ($\downarrow$)} \\
 \midrule
\ours-S & 0.207 & 0.331 & 0.224 & 0.365 & 0.106 & 0.185 & 0.124 & 0.189 & 0.293 & 0.311 & 0.271 & 0.272 \\

 \oursma-S & {\bf 0.165} & {\bf 0.318} & {\bf 0.219} & {\bf 0.319} & {\bf 0.087} & {\bf 0.162} & {\bf 0.118} & {\bf 0.168} & {\bf 0.226} & {\bf 0.278} & {\bf 0.262} & {\bf 0.253} \\
 \midrule
\ours-L  & 0.127 & 0.291 & 0.235 & 0.210 & 0.206 & 0.178 & 0.082 & 0.148 & 0.218 & 0.263 & 0.245 & 0.258 \\

 \oursma-L & \textbf{0.119} & \textbf{0.266} & \textbf{0.221} & \textbf{0.184} & \textbf{0.166} & \textbf{0.148} & \textbf{0.058} & \textbf{0.102} & \textbf{0.196} & \textbf{0.198} & \textbf{0.202} & \textbf{0.194} \\
 \midrule
 \multicolumn{1}{c}{} & \multicolumn{12}{c}{AURC Metric ($\downarrow$)} \\
 \midrule
 \ours-S & 0.093 & 0.211 & 0.154 & 0.142 & 0.317 & 0.237 & 0.311 & 0.162 & 0.258 & 0.262 & 0.258 & 0.245 \\
\oursma-S & {\bf 0.077} & {\bf 0.171} & {\bf 0.116} & {\bf 0.121} & {\bf 0.312} & {\bf 0.235} & {\bf 0.310}  & {\bf 0.079} & {\bf 0.228} & {\bf 0.214} & {\bf 0.210} & {\bf 0.213} \\
\midrule
\ours-L & 0.074 & 0.169 & 0.113 & 0.136 & 0.311 & 0.229 & 0.307  & 0.078 & 0.208 & 0.202 & 0.204 & 0.197 \\
\oursma-L & \textbf{0.063} & \textbf{0.114} & \textbf{0.088} & \textbf{0.109} & \textbf{0.286} & \textbf{0.219} & \textbf{0.297} & \textbf{0.062} & \textbf{0.196} & \textbf{0.192} & \textbf{0.192} & \textbf{0.189} \\
 
\bottomrule
\end{tabular}
\captionof{table}{Selective prediction results on AUGRC ($\downarrow$) for AURC ($\downarrow$) metrics. S denotes the use of a small SP-VLM ($\text{SigLIP}_{\text{B/16}}$), while L denotes the larger $\text{SigLIP}_{\text{SO-400M}}$. In both cases, \oursma is implemented with the \textbf{out-of-domain} retrieval dataset.
}.
\label{tab: aurc-cap-1}
\end{figure*}

\paragraph{Different Metrics.} In the main paper, all evaluations are based on AURC scores. Recent work in selective prediction~\citep{traub2024overcoming} has introduced alternative evaluation metrics. Table \ref{tab: aurc-cap-1} compares \oursma and \ours using the Area Under the Generalized Risk Coverage (AUGRC). While the absolute values differ, both metrics lead to the same conclusions.

\begin{figure}\RawFloats
\centering
\scriptsize
\setlength{\tabcolsep}{2pt}
\begin{tabular}{l|ccc}
\toprule
Methods & Flowers & Pets & UCF101  \\
\midrule
\ours($\text{SigLIP}_{\text{B/16}}$) & 0.100 & 0.112 & 0.157  \\
\oursma
($\text{SigLIP}_{\text{B/16}}$) & {\bf 0.067} & {\bf 0.108} & {\bf 0.112} \\
\midrule
\ours($\text{SigLIP}_{\text{SO-400M}}$) & 0.088 & 0.112 & 0.130  \\
 \ours ($\text{SigLIP}_{\text{SO-400M}}$) & \textbf{0.054} & \textbf{0.101} & \textbf{0.119} \\
\bottomrule
\end{tabular}
\captionof{table}{AURC ($\downarrow$) for OOD detection.}
\label{tab: aurc-ood-1}
\end{figure}

\paragraph{Out-of-distribution detection.} For classification tasks, selective prediction is usually important to distinguish out-of-distribution (OOD) samples from in-distribution (ID) samples. To evaluate \oursma in this setting, we considered  the test splits of Oxford Pets, Oxford Flowers and UCF-101. In all cases, 80\% of the classes were considered ID while the remaining 20\% were considered OOD, and a classifer implemented for the ID classes. The largest class probability was used to implement selective prediction. Table \ref{tab: aurc-ood-1} summarizes the performance of \ours and \oursma for different SP-VLMs. For both small and large SP-VLMs, \oursma outperforms \ours in all datasets. In fact,  \oursma with the small SP-VLM outperforms \ours with the large SP-VLM in all cases. 




\section{Ablations}
\label{sec:moreablations}
This section contains more ablation results for \oursma.


\begin{figure*}[t]
\setlength{\tabcolsep}{2pt}
\centering
\scriptsize
\begin{tabular}{c|cc|cc|ccc|ccccc}
\toprule
& \multicolumn{4}{c|}{Captioning} & \multicolumn{3}{c|}{Classification} & \multicolumn{5}{c}{ITM} \\
\multirow{3}{*}{$K$} & \multicolumn{2}{c|}{MS-COCO} & \multicolumn{2}{c|}{Flickr-30K} & \multirow{2}{*}{Flowers} & \multirow{2}{*}{Pets} & \multirow{2}{*}{UCF-101} & \multirow{2}{*}{SugarCrepe} & \multirow{2}{*}{Winoground} & \multirow{2}{*}{What'sUp} & \multirow{2}{*}{VL-checklist} & \multirow{2}{*}{Foil} \\                   
& Cider-N & Meteor & Cider-N & Meteor  & & & \\ 
\midrule
{1}    & \multicolumn{1}{l}{0.153}          & \multicolumn{1}{l|}{0.332}                   & \multicolumn{1}{l}{0.296}          & \multicolumn{1}{l|}{0.335}             & \multicolumn{1}{c}{0.114}          & \multicolumn{1}{c}{0.132}                   & \multicolumn{1}{c|}{0.096}  & \multicolumn{1}{c}{0.073}          & \multicolumn{1}{c}{0.204}                   & \multicolumn{1}{c}{0.216}          & \multicolumn{1}{c}{0.215}   & 0.209     \\ 

{5}    & \multicolumn{1}{l}{0.130}           & \multicolumn{1}{l|}{0.312}          & \multicolumn{1}{l}{0.237} & \multicolumn{1}{l|}{0.317}  & \multicolumn{1}{c}{\textbf{0.063}}           & \multicolumn{1}{c}{\textbf{0.114}}          & \multicolumn{1}{c|}{\textbf{0.088}} & \multicolumn{1}{c}{0.068}           & \multicolumn{1}{c}{0.197}          & \multicolumn{1}{c}{0.197} & \multicolumn{1}{c}{0.208}   &  0.206        \\ 

{15}   & \multicolumn{1}{l}{\textbf{0.109}} & \multicolumn{1}{l|}{\textbf{0.286}}           & \multicolumn{1}{l}{\textbf{0.219}}          & \multicolumn{1}{l|}{\textbf{0.297}}  & \multicolumn{1}{c}{0.065} & \multicolumn{1}{c}{\textbf{0.114}}           & \multicolumn{1}{c|}{0.089}  & \multicolumn{1}{c}{\textbf{0.062}} & \multicolumn{1}{c}{\textbf{0.196}}           & \multicolumn{1}{c}{\textbf{0.192}}          & \multicolumn{1}{c}{\textbf{0.192}}       & \textbf{0.189}     \\ 

{20}   & \multicolumn{1}{l}{0.111}          & \multicolumn{1}{l|}{0.287}        & \multicolumn{1}{l}{0.221}          & \multicolumn{1}{l|}{0.305}   & \multicolumn{1}{c}{0.064}          & \multicolumn{1}{c}{\textbf{0.114}}        & \multicolumn{1}{c|}{\textbf{0.088}}    & \multicolumn{1}{c}{0.064}          & \multicolumn{1}{c}{\textbf{0.196}}        & \multicolumn{1}{c}{\textbf{0.192}}          & \multicolumn{1}{c}{0.195}            &  0.192          \\
\bottomrule
\end{tabular}
\captionof{table}{Neighborhood size $K$ for all tasks.}
\label{tab:ablation-k}
\end{figure*}

\paragraph{Neighborhood size.} The performance of \oursma depends on the neighborhood size $K$ of (\ref{eq:Nbg}), i.e.  the size of set of texts retrieved from $\cal R$. Table \ref{tab:ablation-k} shows an ablation for this parameter, for the captioning, image-text matching and classification tasks. Initially, as $K$ increases, the AURC decreases, indicating better performance for larger retrieval sets. Note, in particular the advantages of the soft nearest neighbor embedding implemented by proxy embeddings over a hard nearest neighbor embedding, where just the top nearest neighbor is retrieved from $\cal R$. This corresponds to using $K=1$. For all tasks and datasets, the soft nearest neighbor operation has better performance. These results confirm the importance of proxy embeddings that fuse information from several image-text pairs relevant to the query. However, beyond a certain $K$, performance starts to degrade, with AURC rising. This is because larger retrieval sets start to include irrelevant or poor matches, and the retrieved proxy is less related to the query. Such behavior is fairly common for nearest neighbor approaches. For both tasks, the best AURC score is achieved in the neighborhood of $K=15$, suggesting an optimal balance between retrieval size and performance. We use this value as the default for \oursma in all experiments.

\begin{figure}[t]
\centering
\scriptsize
\setlength{\tabcolsep}{2pt}
\begin{tabular}{c|cc|cc|ccc|ccccc}
\toprule
& \multicolumn{4}{c|}{Captioning} & \multicolumn{3}{c|}{Classification} & \multicolumn{5}{c}{ITM} \\
\multirow{3}{*}{$\mathcal{R}$ Size/\%} & \multicolumn{2}{c|}{MS-COCO} & \multicolumn{2}{c|}{Flickr-30K} & \multirow{2}{*}{Flowers} & \multirow{2}{*}{Pets} & \multirow{2}{*}{UCF-101} & \multirow{2}{*}{SugarCrepe} & \multirow{2}{*}{Winoground} & \multirow{2}{*}{What'sUp} & \multirow{2}{*}{VL-checklist} & \multirow{2}{*}{Foil} \\              
& Cider-N & Meteor & Cider-N & Meteor  & & & \\  
\midrule
\multicolumn{1}{c|}{20}   & \multicolumn{1}{l}{0.138}   & \multicolumn{1}{c}{0.314}   & \multicolumn{1}{|c}{0.231}   & \multicolumn{1}{l|}{0.321} & 0.063 & 0.114 & 0.088 & 0.074 & 0.209 & 0.265 & 0.256 & 0.248   \\ 
\multicolumn{1}{c|}{40}    & \multicolumn{1}{l}{0.122}   & \multicolumn{1}{c}{0.310}  & \multicolumn{1}{|c}{0.226}   & \multicolumn{1}{l|}{0.323} & 0.063 & 0.114 & 0.088 & 0.074 & 0.196 & 0.192 & 0.192 & 0.189\\ 
\multicolumn{1}{c|}{60}    & \multicolumn{1}{l}{0.116}   & \multicolumn{1}{c}{0.292}   & \multicolumn{1}{|c}{0.221}   & \multicolumn{1}{l|}{0.312} & 0.063 & 0.114 & 0.088 & 0.072 & 0.204 & 0.192 & 0.192 & 0.189\\ 
\multicolumn{1}{c|}{80}   & \multicolumn{1}{l}{0.111}   & \multicolumn{1}{c}{0.289}   & \multicolumn{1}{|c}{0.220}   & \multicolumn{1}{l|}{0.312} & 0.063 & 0.114 & 0.088 & 0.064 & 0.196 & 0.192 & 0.194 & 0.192\\ 
\multicolumn{1}{c|}{100}  & \multicolumn{1}{l}{\bf{0.109}}   & \multicolumn{1}{c}{\bf{0.286}}  &  \multicolumn{1}{|c}{\bf{0.219}}   & \multicolumn{1}{l|}{\bf{0.297}} & \bf{0.063} & \bf{0.114} & \bf{0.088} & \textbf{0.062} & \textbf{0.196} & \textbf{0.192} & \textbf{0.192} & \textbf{0.189}\\ 
\bottomrule
\end{tabular}
\captionof{table}{Performance of \oursma (SigLIP$_{\text{SO-400M}}$) versus the percentage of the retrieval dataset $\cal R$ for all tasks. Retrieval set $\cal R$ is in-domain.}
\label{tab:ablation-size}
\end{figure}

\paragraph{Size of Retrieval Set.} 
 To analyze the effect of size of retrieval set on the performance of \oursma, we randomly sub-sampled the \emph{in-domain retrieval set}. Table~\ref{tab:ablation-size} shows that, as dataset size decreases, the performance of \oursma can degrade for the captioning task. This is because the samples relevant (nearest neighbors) to the query are removed from the retrieval set. However, for classification and ITM, performance stays mostly unaltered  for sizes as low as $40\%$. This shows that these tasks are much less sensitive to the dataset size.

\begin{figure*}[t]\RawFloats
\centering
\scriptsize
\setlength{\tabcolsep}{2pt}
\begin{tabular}{c|cc|cc|ccc|ccccc}
\toprule
& \multicolumn{4}{c|}{Captioning} & \multicolumn{3}{c|}{Classification} & \multicolumn{5}{c}{ITM} \\
\multirow{3}{*}{Variant} & \multicolumn{2}{c|}{MS-COCO} & \multicolumn{2}{c|}{Flickr-30K} & \multirow{2}{*}{Flowers} & \multirow{2}{*}{Pets} & \multirow{2}{*}{UCF-101} & \multirow{2}{*}{SugarCrepe} & \multirow{2}{*}{Winoground} & \multirow{2}{*}{What'sUp} & \multirow{2}{*}{VL-checklist} & \multirow{2}{*}{Foil} \\                   
& Cider-N & Meteor & Cider-N & Meteor  & & & \\  
\midrule 
\multicolumn{1}{c|}{i2ir} & \multicolumn{1}{c}{0.135}          & \multicolumn{1}{c}{0.358}                   & \multicolumn{1}{c}{0.309}          & \multicolumn{1}{c|}{0.335} & \multicolumn{1}{c}{0.128}          & \multicolumn{1}{c}{0.154}                   & \multicolumn{1}{c|}{0.209}  & \multicolumn{1}{c}{0.142}          & \multicolumn{1}{c}{0.256}                   & \multicolumn{1}{c}{0.279}          & \multicolumn{1}{c}{0.245}  & 0.214                 \\ 

\multicolumn{1}{c|}{t2ir} & \multicolumn{1}{c}{0.152}          & \multicolumn{1}{c}{0.344}          & 0.264         & \multicolumn{1}{c|}{0.342}  & \multicolumn{1}{c}{0.116}          & \multicolumn{1}{c}{0.118}          & \multicolumn{1}{c|}{0.164}   & \multicolumn{1}{c}{0.116}          & \multicolumn{1}{c}{0.242}          & \multicolumn{1}{c}{0.266}  & \multicolumn{1}{c}{0.242}    &  0.208               \\ 
\multicolumn{1}{c|}{t2tr} & \multicolumn{1}{c}{0.162}          & \multicolumn{1}{c}{0.367}                    & \multicolumn{1}{c}{0.269}          & \multicolumn{1}{c|}{0.341}   & \multicolumn{1}{c}{0.106}          & \multicolumn{1}{c}{0.126}                    & \multicolumn{1}{c|}{0.169}    & \multicolumn{1}{c}{0.128}          & \multicolumn{1}{c}{0.238}                    & \multicolumn{1}{c}{0.257}          & \multicolumn{1}{c}{0.231}      &    0.216             \\ 
\multicolumn{1}{c|}{i2tr} & \multicolumn{1}{c}{\textbf{0.109}} & \multicolumn{1}{c}{\textbf{0.286}} &  \multicolumn{1}{c}{\textbf{0.219}} & \multicolumn{1}{c|}{\textbf{0.297}} & \multicolumn{1}{c}{\textbf{0.063}} & \multicolumn{1}{c}{\textbf{0.114}} &  \multicolumn{1}{c|}{\textbf{0.088}} & \multicolumn{1}{c}{\textbf{0.062}} & \multicolumn{1}{c}{\textbf{0.196}} &  \multicolumn{1}{c}{\textbf{0.192}} & \multicolumn{1}{c}{\textbf{0.192}} & \multicolumn{1}{c}{\textbf{0.189}}\\
\bottomrule
\end{tabular}
\captionof{table}{Retrieval Variants for all tasks.}
\label{tab:ablation-var}
\end{figure*}

\paragraph{Retrieval variants.} Experiments were performed to compare the performance of the different retrieval variants of Table~\ref{tab:variants}, on the captioning, image-text matching and classification task, using a neighborhood size $K = 15$ and contrastive scoring. Table \ref{tab:ablation-var} shows that, for the above tasks, image to text retrieval (i2tr) achieves the best performance under all metrics. 
This is consistent with previous observations that the CLIP similarity score (cosine similarity in CLIP embedding space) is more reliable when its two arguments are text embeddings, suggesting a better conceptual organization of the CLIP embedding for text~\citep{iscen2023improving}. Since, under i2tr, the retrieved proxy is a proxy caption, i2tr is expected to outperform i2ir, which returns an image proxy. On the other hand, the use of the predicted caption $f({\bf x})$ as a query is expected not to perform well, since any captioning errors will induce the retrieval of image-caption pairs from $\cal R$ that are not relevant to the captioning of $\bf x$. This is true for other two tasks as well. It explains why the t2ir and t2tr variants do not perform well.
Given these results, we use i2tr as the default retrieval implementation of \oursma. This variant is used in all other experiments of the paper.

\begin{figure*}[ht]\RawFloats
    \centering
    \setlength{\tabcolsep}{2pt}
    \begin{tabular}{cccc}
    \includegraphics[width=.24\linewidth]{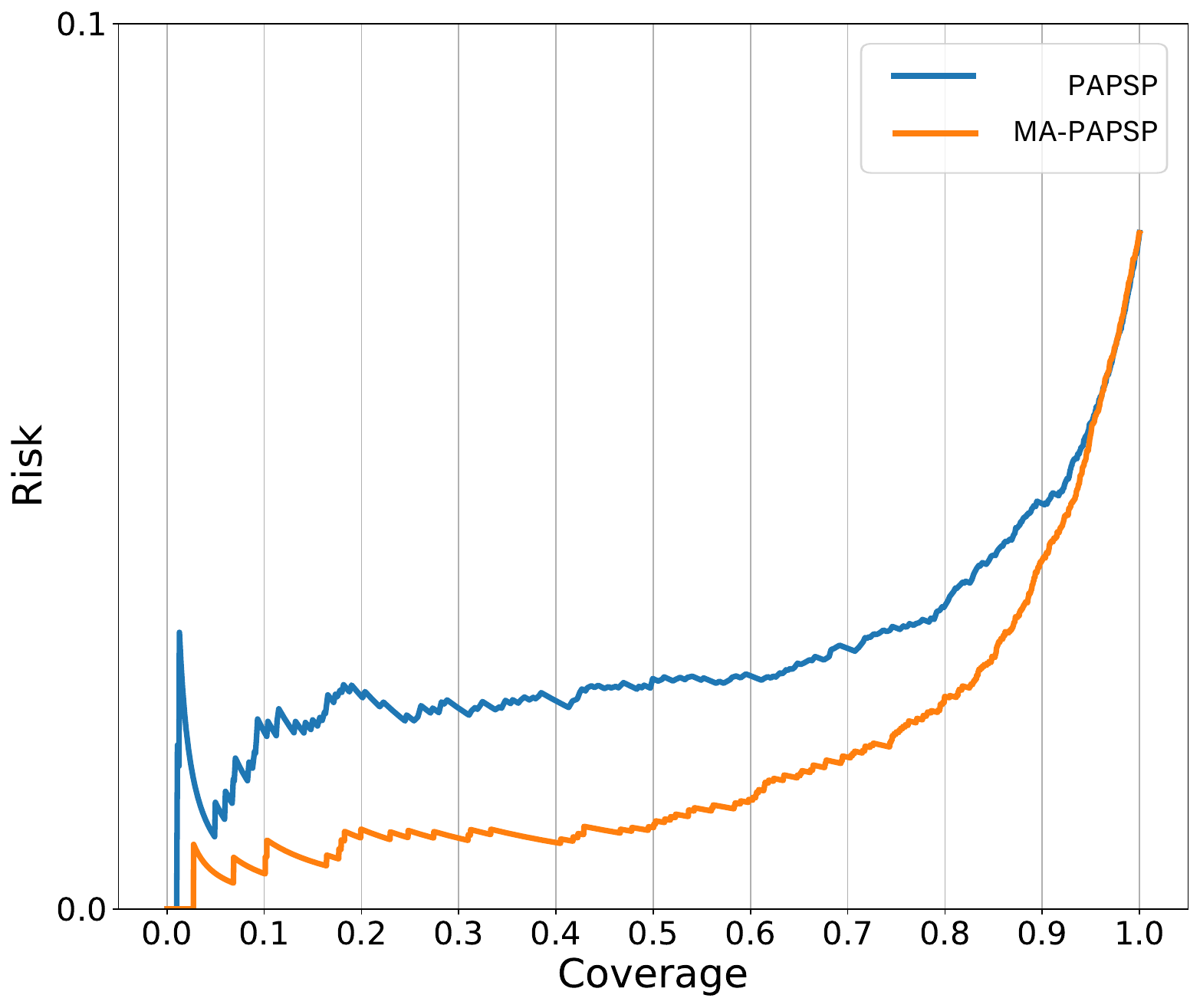} & \includegraphics[width=.24\linewidth]{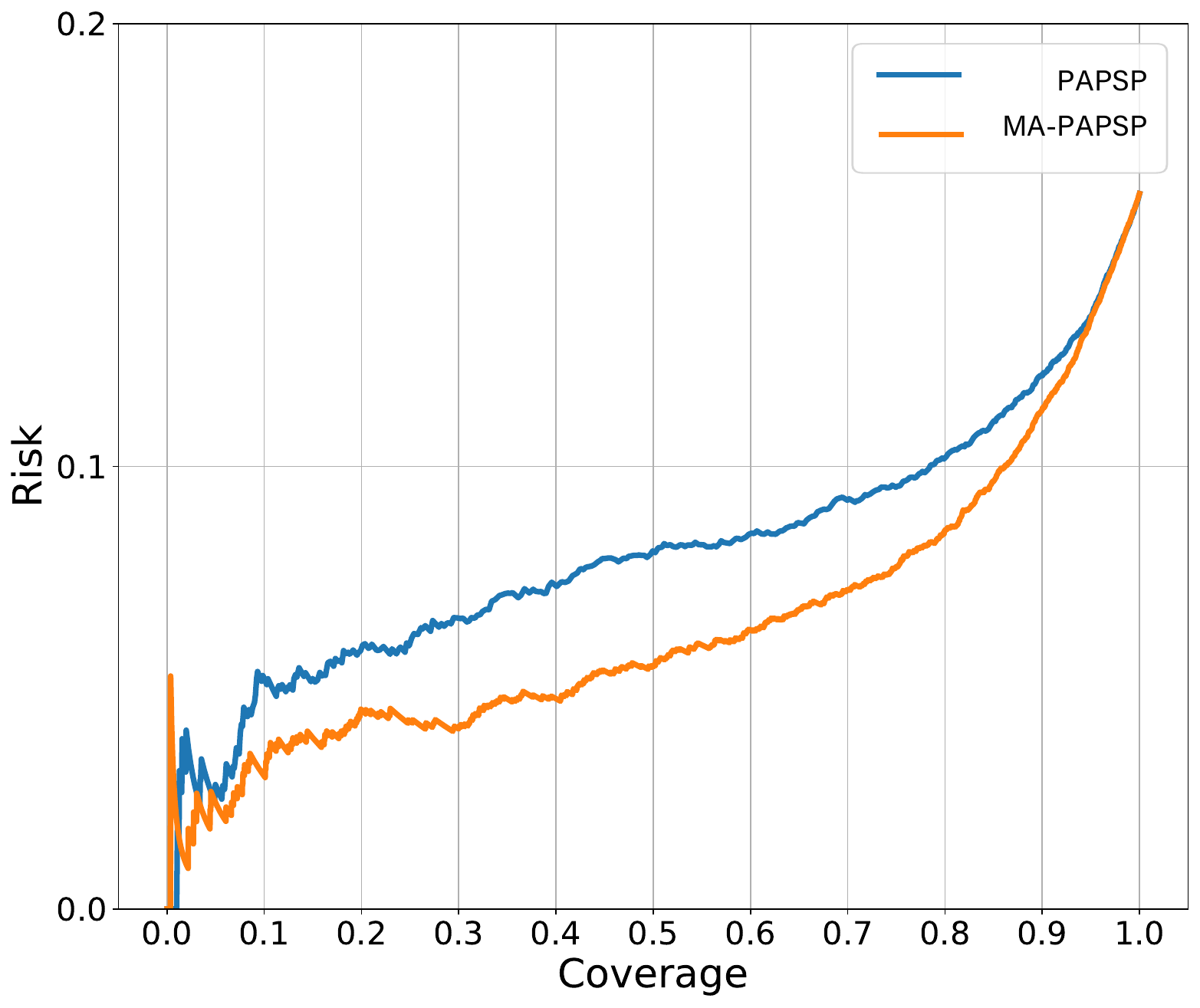} &
    \includegraphics[width=.24\linewidth]{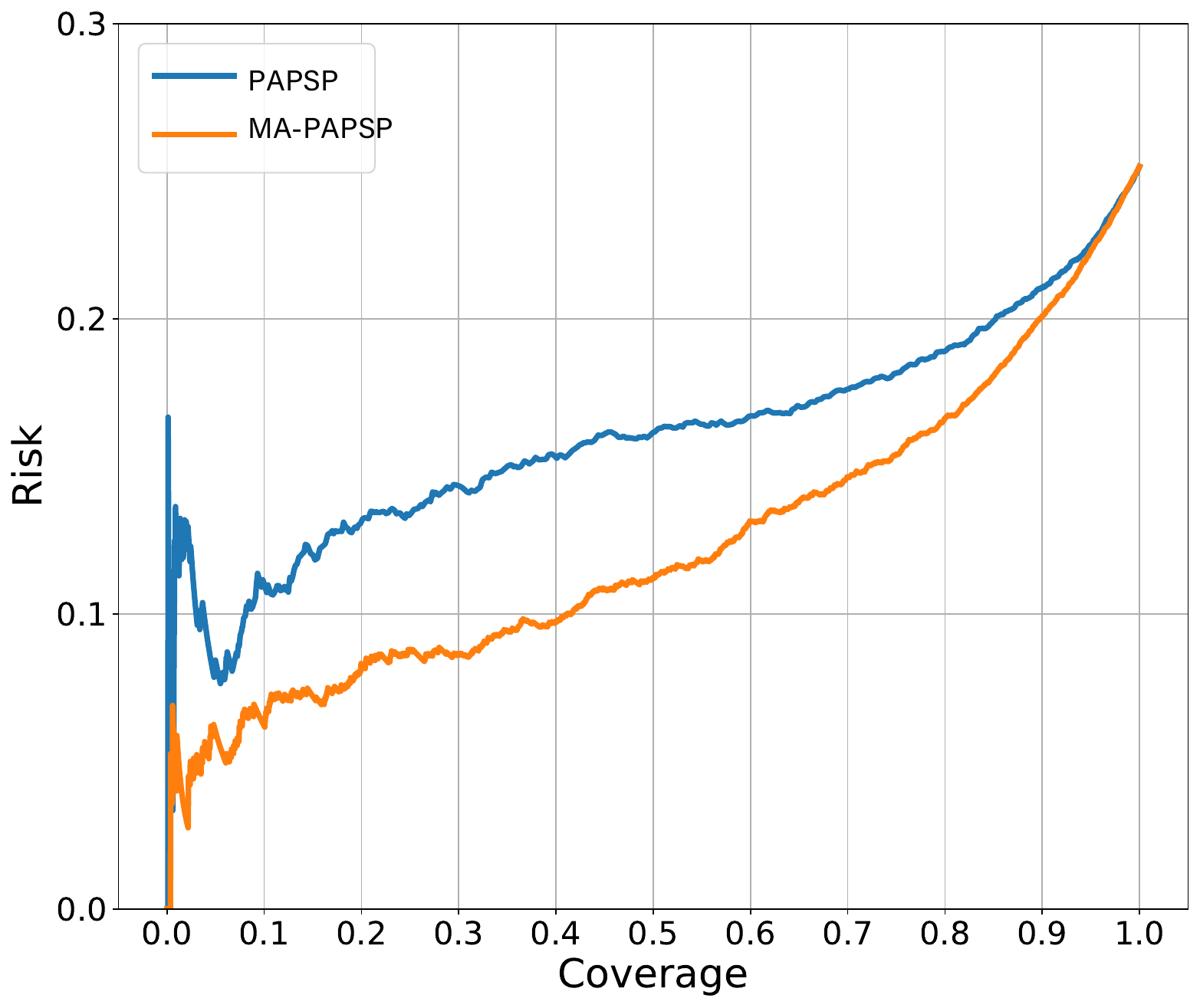} & \includegraphics[width=.24\linewidth]{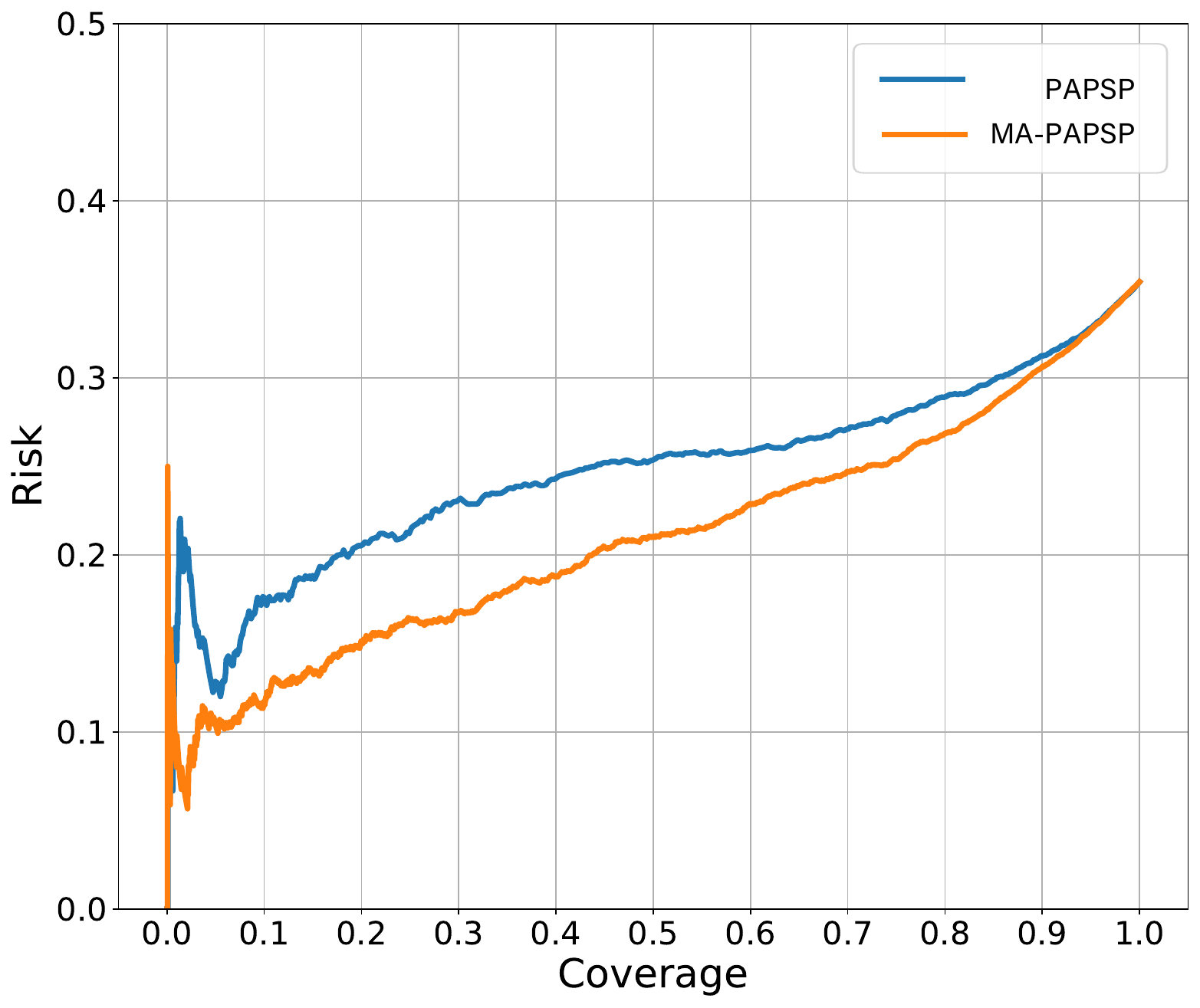}
    \end{tabular}
    \caption{Risk-Coverage curves of CLIP and \ours for selective image captioning under the Cider-N (N = 4) metric, for different values of the threshold $\beta$ used to define the risk. Left to right: $\beta = 0.2$, $\beta = 0.4$, $\beta = 0.6$, $\beta = 0.8$.}
    \label{fig:RiskCov}
\end{figure*}

\paragraph{Effect of relevance threshold.} The evaluation of selective prediction requires the specification of threshold $\beta$ in (\ref{eq:loss}) for the risk computation. We ablated the importance of this threshold by plotting risk-coverage curves for different values of $\beta$. Figure \ref{fig:RiskCov} shows an example for the Cider-N metric, for selective image captioning on the MS-COCO dataset, using \oursma and the OpenAI CLIP SP-VLM. While the four plots obtained with different values of $\beta$ are different, the relative behavior of \ours and \oursma curves is similar, with \oursma achieving substantially lower risk for most coverage values 
This shows that the exact value of $\beta$ is not critical. Note, however, that the same threshold should be used across experiments that involve comparison of AURC curves. We have followed this practice in all experiments.

\begin{figure}[hbt!]\RawFloats
\scriptsize
\centering
\setlength{\tabcolsep}{2pt}
\begin{tabular}{l|cc|cc|c}
\toprule
\multirow{2}{*}{Variant} & \multicolumn{2}{c|}{MS-COCO} & \multicolumn{2}{c|}{Flickr-30K} & \multirow{1}{*}{Inference} \\                 
& Cider-N & Meteor & Cider-N & Meteor & \multirow{1}{*}{Time}\\
\midrule
\multicolumn{6}{c}{\oursma without CC} \\
\midrule
\oursma  & 0.143 & 0.323 & 0.239 & 0.316 & -  \\  
\midrule
\multicolumn{6}{c}{\oursma w/ Rule-based CC} \\
\midrule
 Rule-Based                                                & 0.113   & 0.301  & 0.222   & 0.301 & 4.72 ms  \\  

 \midrule
 \multicolumn{6}{c}{\oursma w/ LLM-based CC - Small LLM} \\
 \midrule
 
 $\text{Qwen-2.5}_\text{3B}$                                                 & 0.117   & 0.306  & \textbf{0.213}   & 0.304 & 42.26 ms  \\
 
 $\text{Phi-3.5}_\text{3B}$                                               & 0.109 & 0.286  & 0.219 & 0.297 & 41.06 ms   \\

 \midrule
 \multicolumn{6}{c}{\oursma w/ LLM-based CC - Large LLM} \\
 \midrule
                           
 $\text{Qwen-2.5}_\text{7B}$                                                 & 0.111   & 0.283  & \textbf{0.217}    & 0.278 &  84.96 ms \\

  $\text{Llama-3}_\text{8B}$                                  & \textbf{0.108}   & \textbf{0.277}  & 0.219   & \textbf{0.276} &  92.65 ms  \\
     \bottomrule
\end{tabular}
\captionof{table}{Selective captioning AURC ($\downarrow$) for different methods of contrastive component (CC).}
\label{tab: aurc-design-llm}
\end{figure}

\paragraph{Contrastive component (CC).} For tasks like image-text matching and classification, negative samples are available by definition of the task. For image captioning, we compare two alternatives for generating contrastive examples: a \emph{rule-based} approach and an \emph{LM-based} approach. The rule-based method involves substituting objects, attributes, or verbs in the caption using a predefined set of rules. We used SpaCy to identify the part of speech—such as noun, adjective, or verb—and then replaced the identified word with another one from WordNet \citep{miller1995wordnet}. In contrast, the LM-based method takes a given context (provided in \ref{sec:qualres}) and prompt and outputs a modified prompt with swapped objects, attributes, and verbs. Unlike the rule-based approach, the LM-generated captions are generally more coherent and grammatically correct as shown in \ref{sec:qualres} a. Table \ref{tab: aurc-design-llm} shows that, in addition to faster inference of the rule-based approach than that of LM-based methods, it is able to perform at par with even the implementation of \oursma with a LLM. 
For example \oursma with Phi-3.5 has just an improvement of 4 points over \oursma with rule based contrastive scores for MS-COCO with Cider-N setting. This indicates more effective generation of contrastive samples. Further, while larger LLMs achieve better results, the differences are not very significant. For example, Llama-3 (8B parameter model) has an improvement of 1 point over Phi-3.5 (3B) for MS-COCO (Cider-N), but its inference time is almost double (92.65ms vs 43.06ms of Phi-3). Due to its better trade-off between performance and computation overhead, we use the rule-based approach in all our experiments.

\section{More related works}
\label{sec:morerelatedworks}

In this section, we talk about using more works that are related to \oursma. 

\paragraph{Foundation Models for Scoring.} Visual language representation models, like CLIP \citep{radford2021learning}, extract features from vision-language pairs and use contrastive learning \citep{ding2021support} to learn a space where the two modalities are aligned. This enables the design of alignment scores based on simple operations, like cosine similarity, that underlie measures like the CLIP score~\citep{hessel2021clipscore}. Thresholding the CLIP score is a low-complexity solution for selective prediction, which we denote as \ours and use as baseline. \oursma improves on this baseline.

\clearpage

\section{Qualitative results}
\label{sec:qualres}
\begin{figure*}[hbt!]\RawFloats
    \centering
    \includegraphics[width=\linewidth]{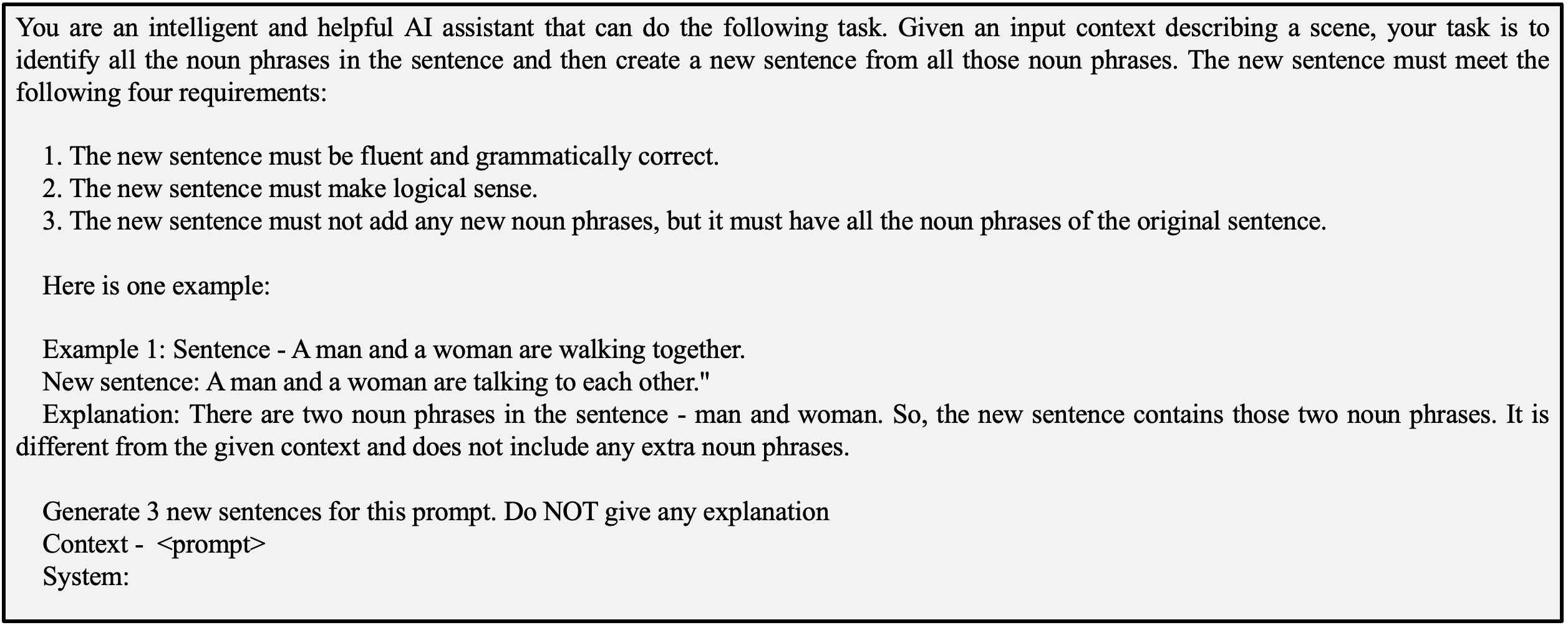}
    \caption{LLM Prompt used to generate negatives from a given prompt.}
    \label{fig: prompt}
\end{figure*}

\begin{figure*}[hbt!]
    \centering
    \setlength{\tabcolsep}{10pt}
    \begin{tabular}{cc}
    \includegraphics[width=0.45\linewidth]{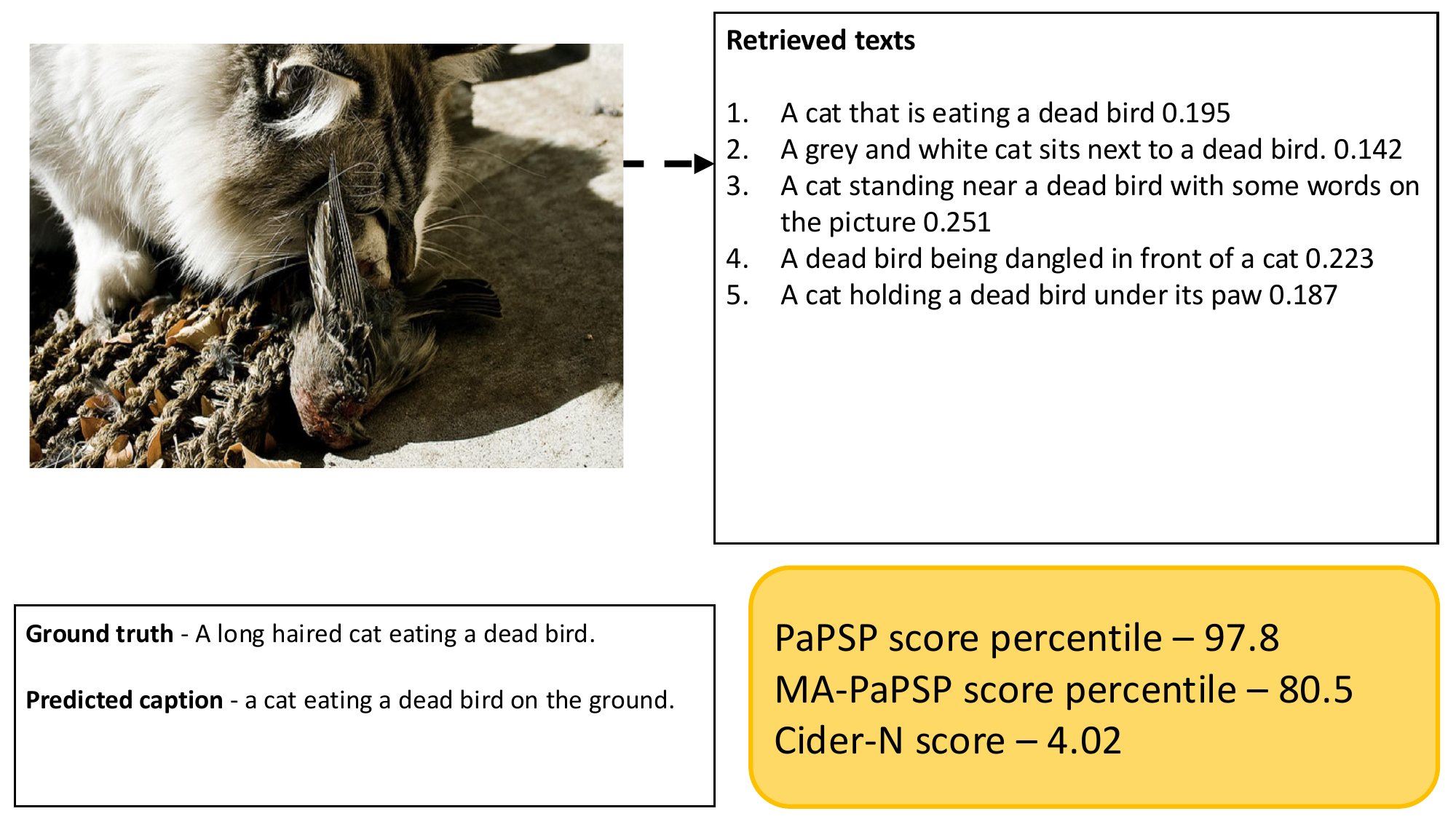} & \includegraphics[width=0.45\linewidth]{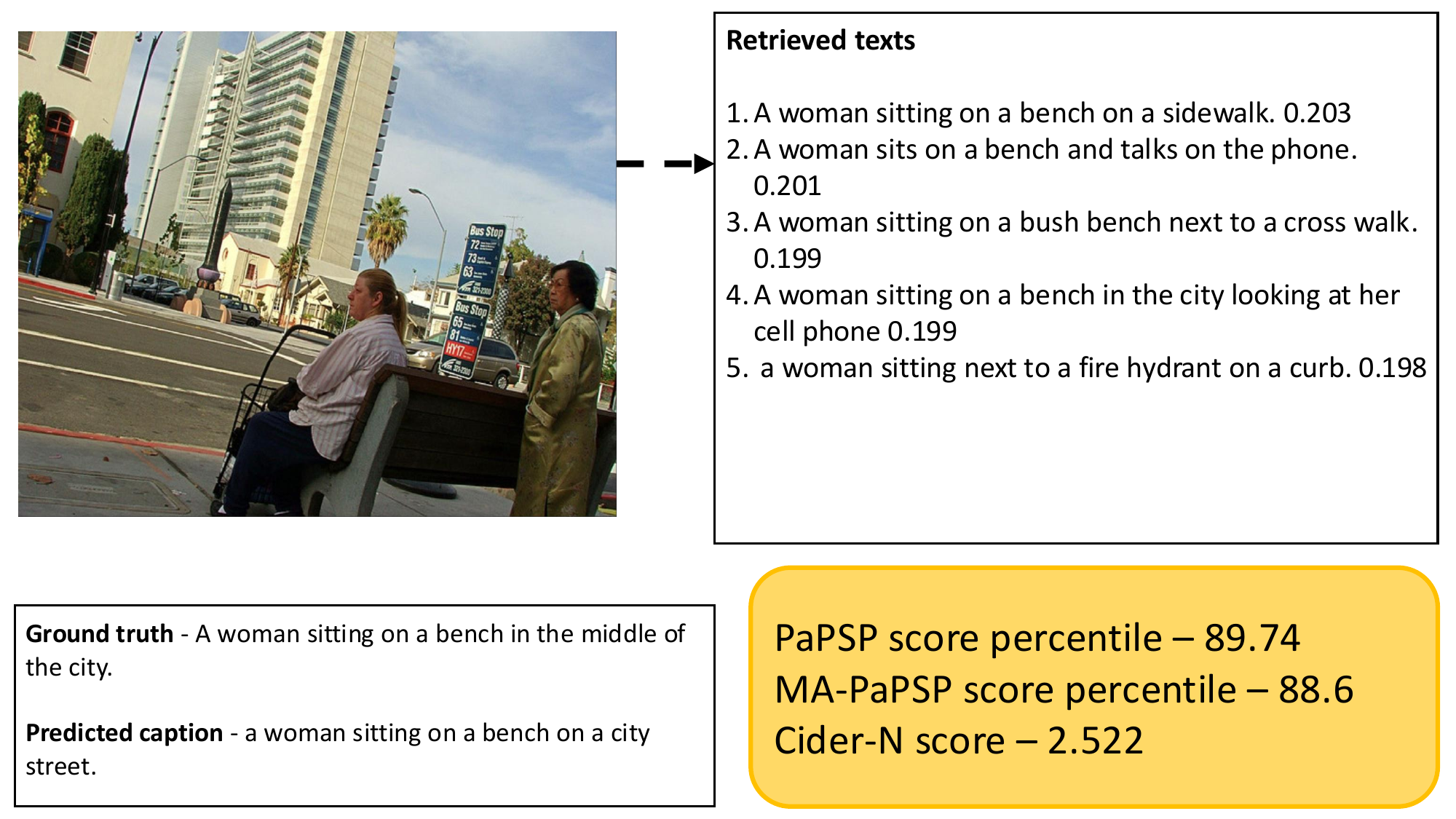} \\
    {} \\
    \includegraphics[width=0.45\linewidth]{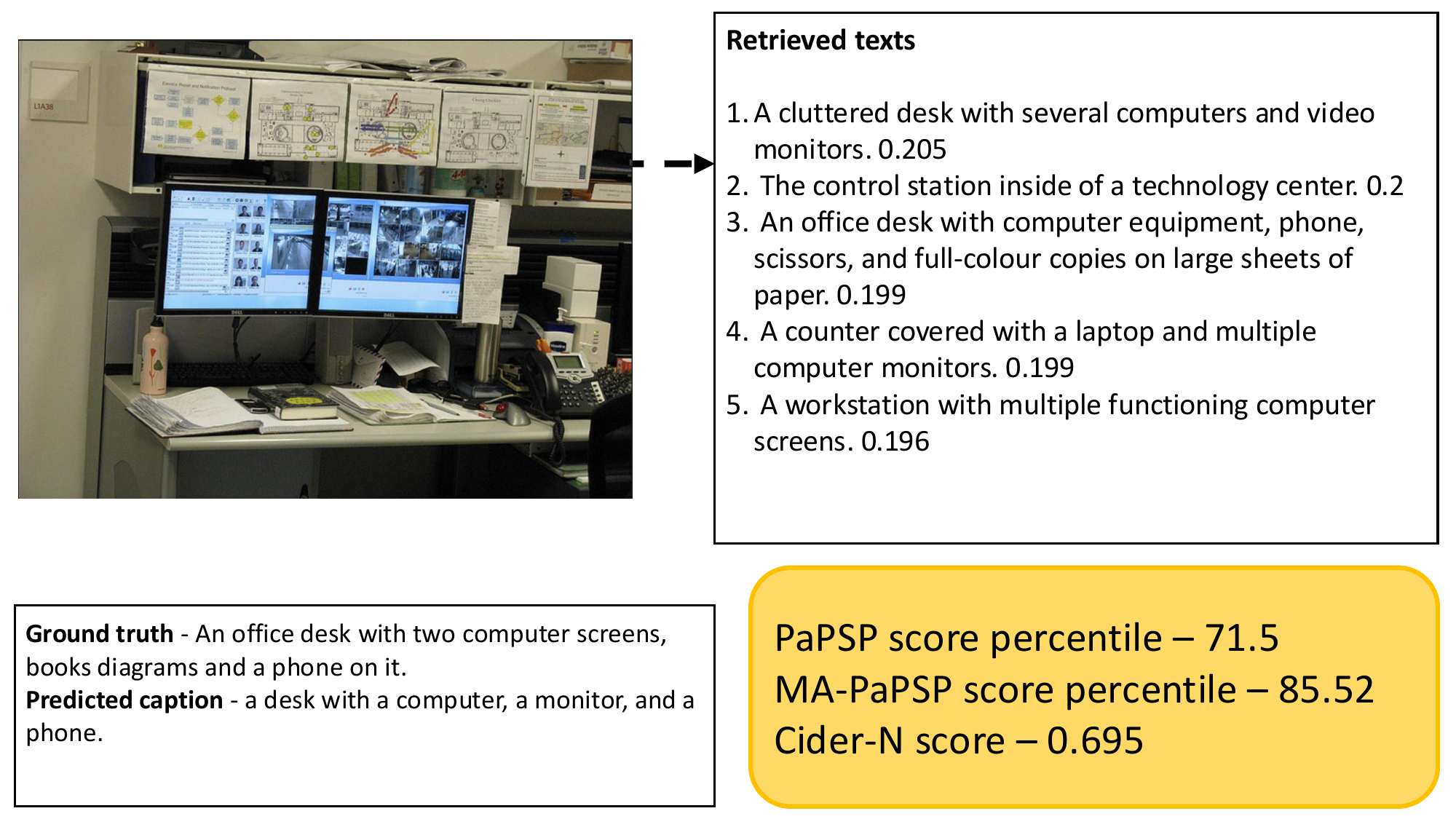} & \includegraphics[width=0.45\linewidth]{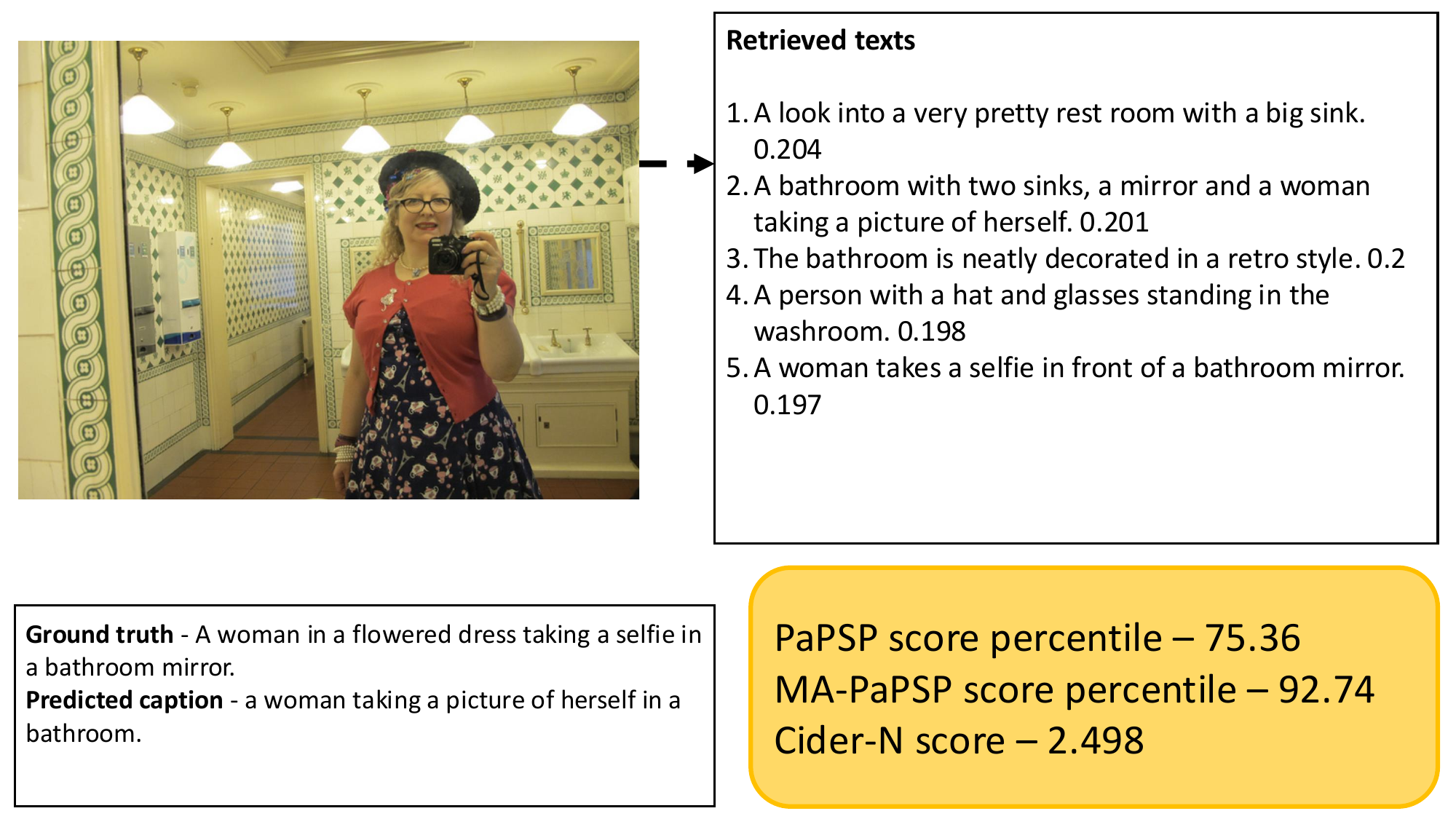}\\
    \end{tabular}
    \caption{This figure presents examples where both \ours and \oursma accept the inputs. Each example includes the following components: the image, its predicted caption, the ground-truth caption, the retrieved captions along with their normalized weights, \ours baseline scores, \oursma scores, and the ground-truth score (Cider-N).}
    \label{fig: Miscc1}
\end{figure*}

\begin{figure*}[hbt!]
    \centering
    \setlength{\tabcolsep}{10pt}
    \begin{tabular}{cc}
    \includegraphics[width=0.45\linewidth]{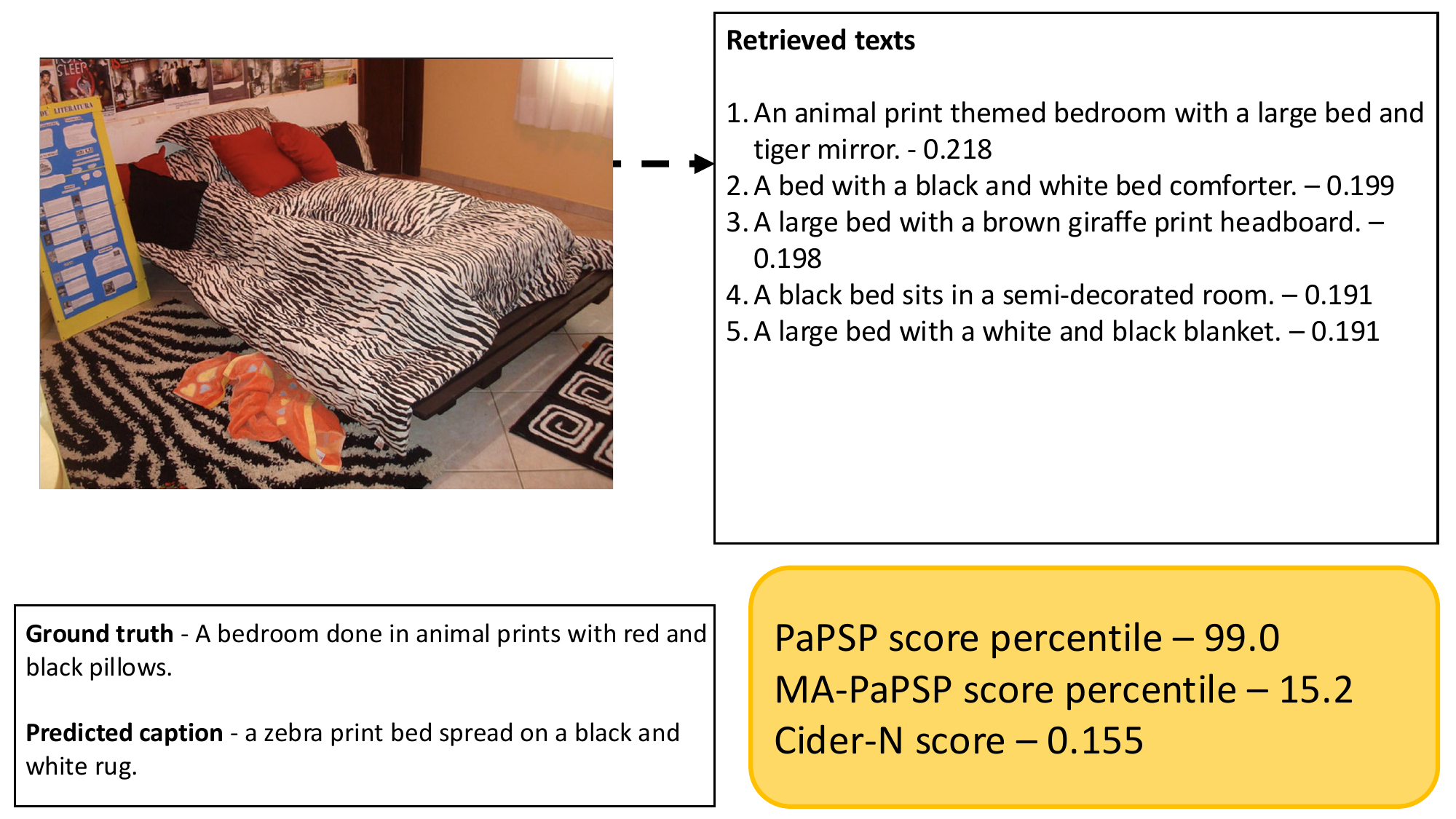} & \includegraphics[width=0.45\linewidth]{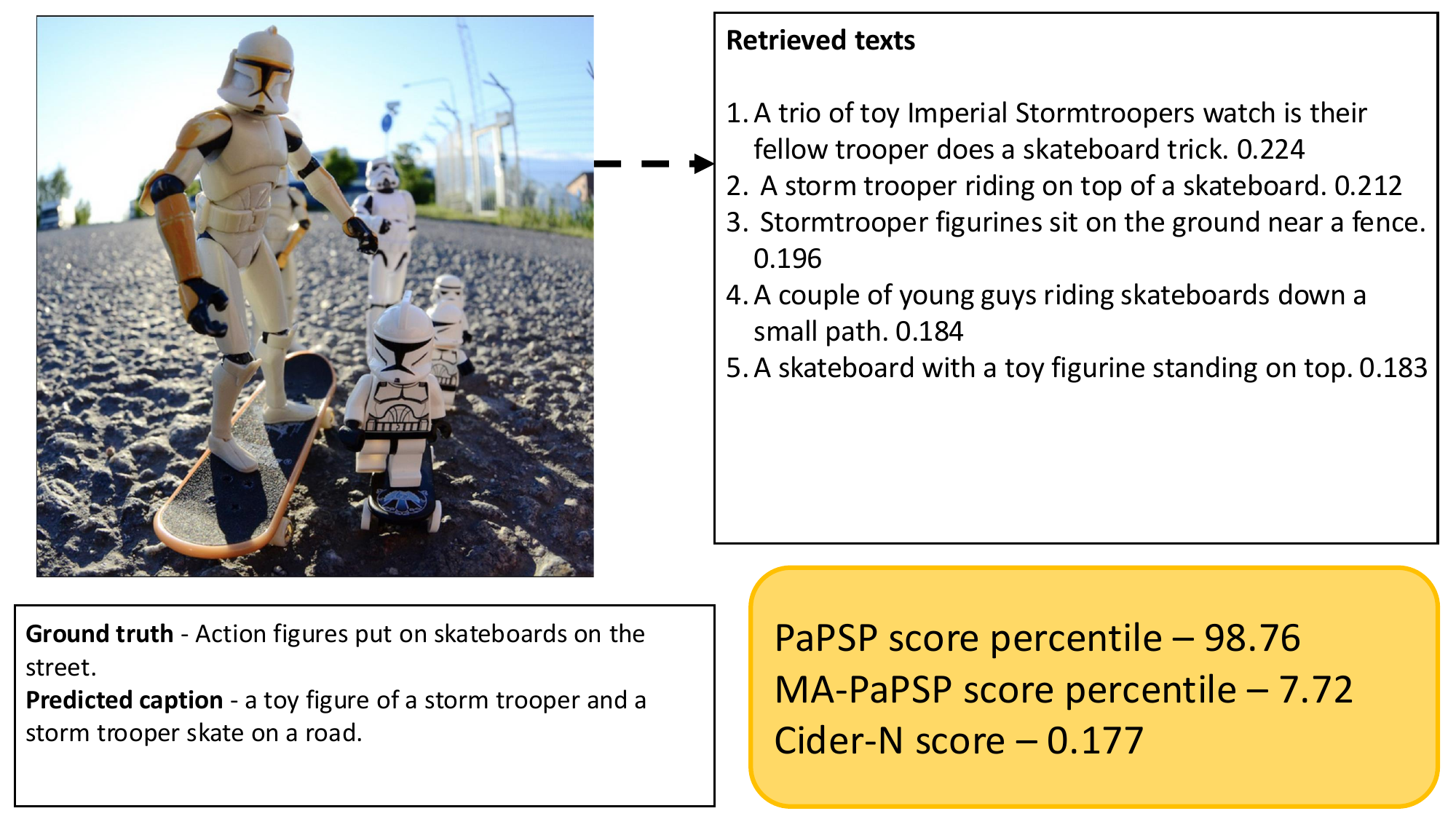}  \\
    {} \\
    \includegraphics[width=0.45\linewidth]{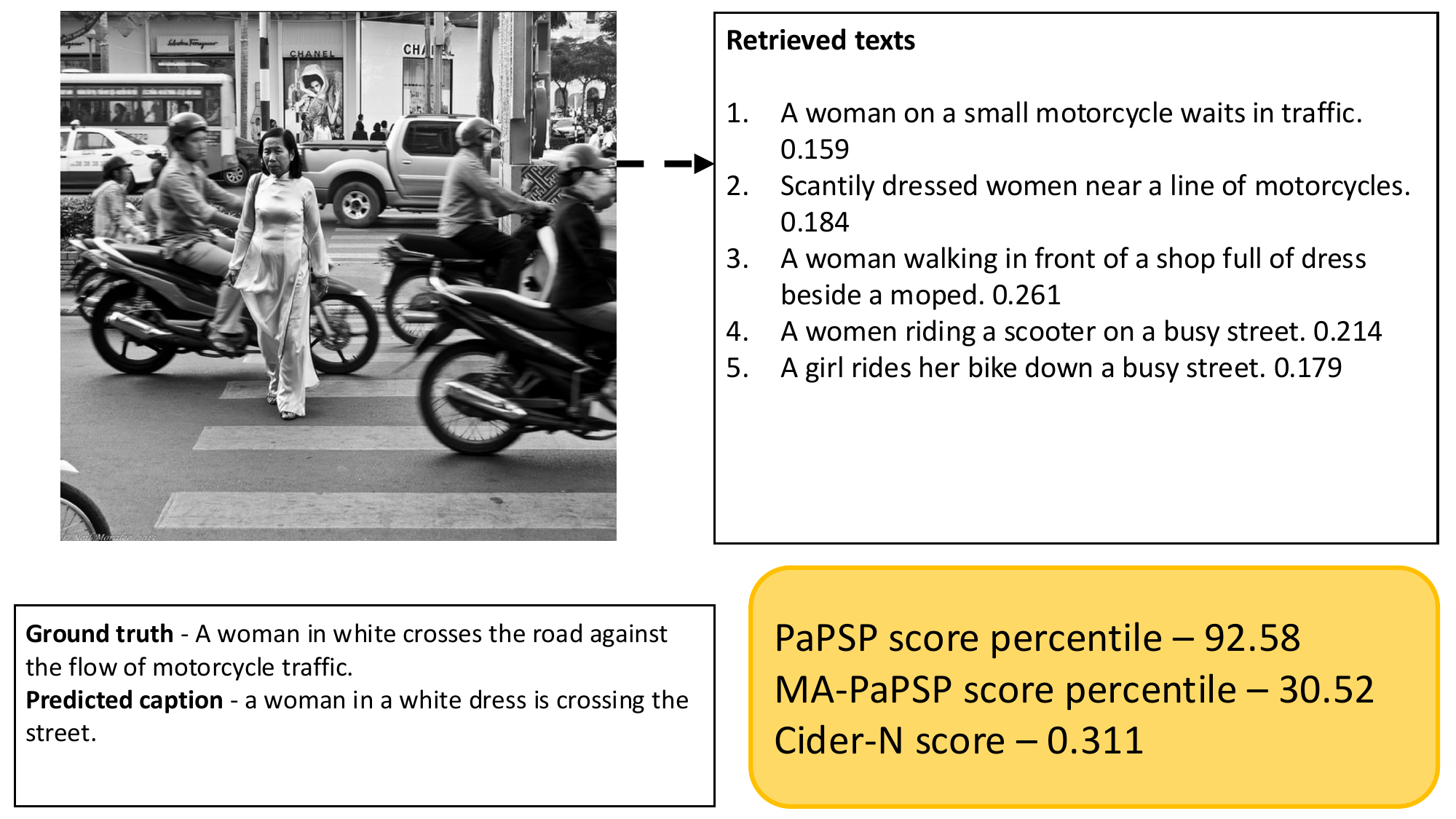} & \includegraphics[width=0.45\linewidth]{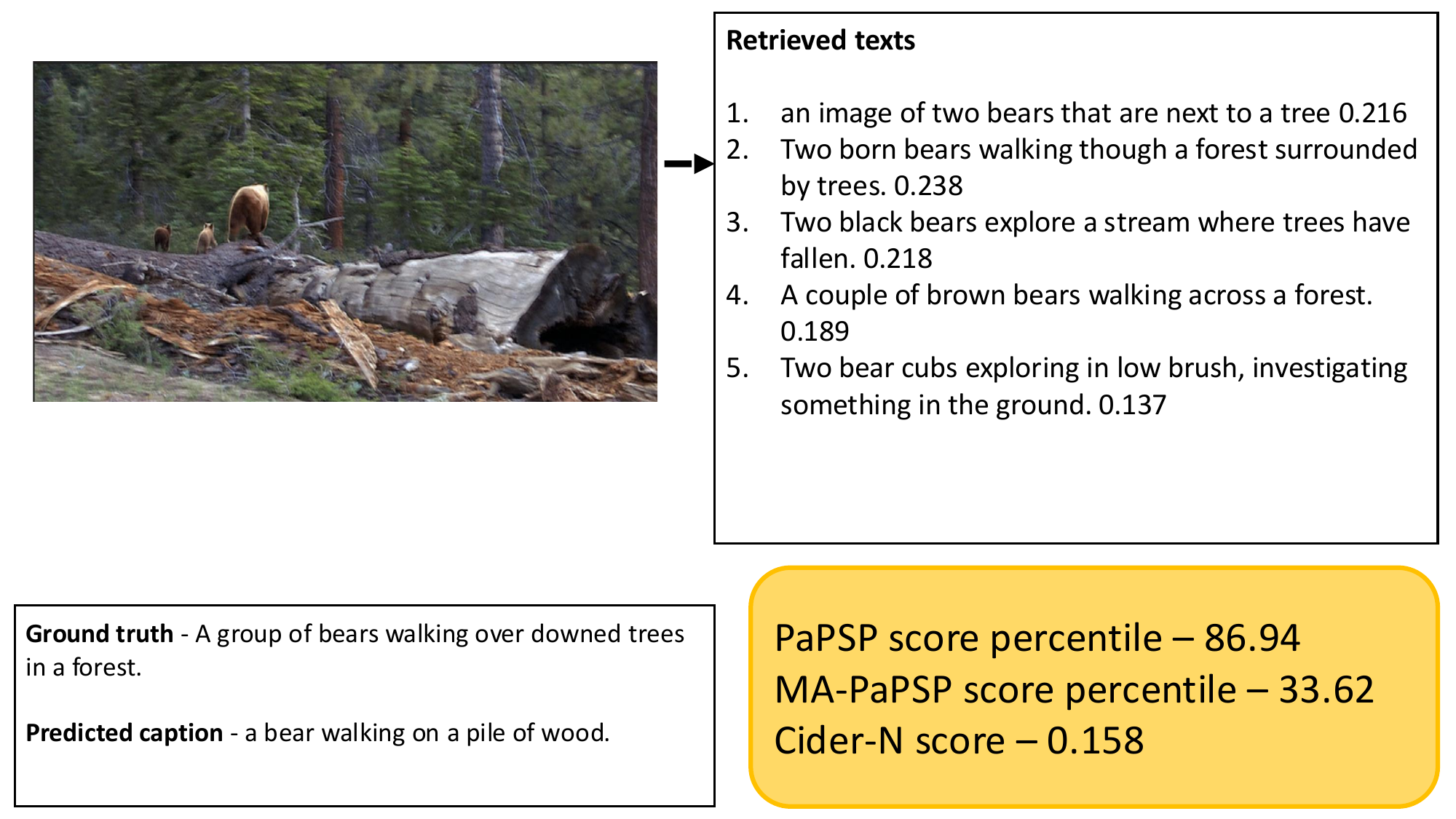} \\
    {} \\
    \end{tabular}
    \caption{This figure presents examples where \ours accepts and \oursma rejects the inputs. Each example includes the following components: the image, its predicted caption, the ground-truth caption, the retrieved captions along with their normalized weights, \ours baseline scores, \oursma scores, and the ground-truth score (Cider-N).}
    \label{fig: Miscc2}
\end{figure*}

\begin{figure*}[hbt!]
    \centering
    \setlength{\tabcolsep}{10pt}
    \begin{tabular}{cc}
    \includegraphics[width=0.45\linewidth]{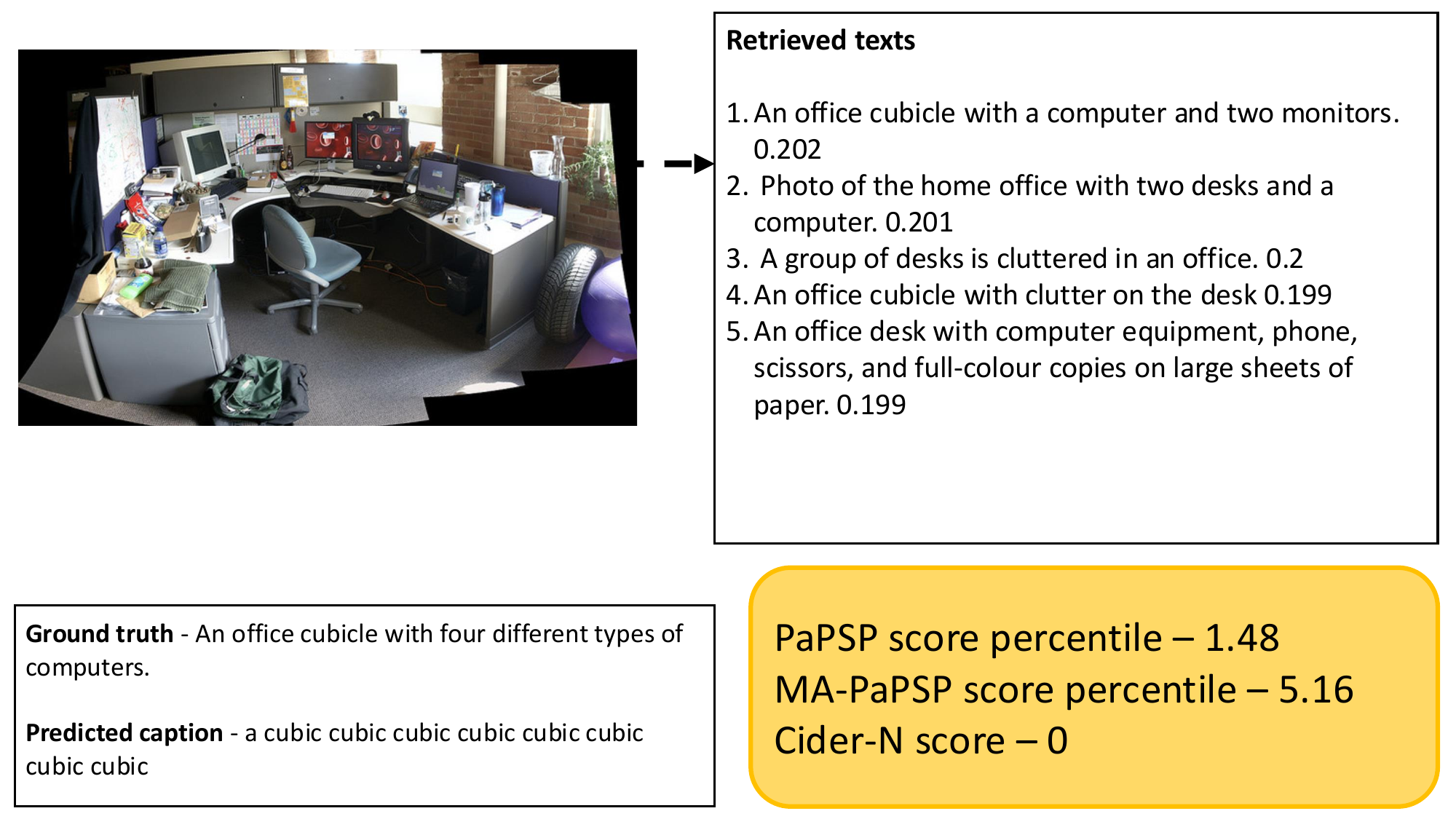} & \includegraphics[width=0.45\linewidth]{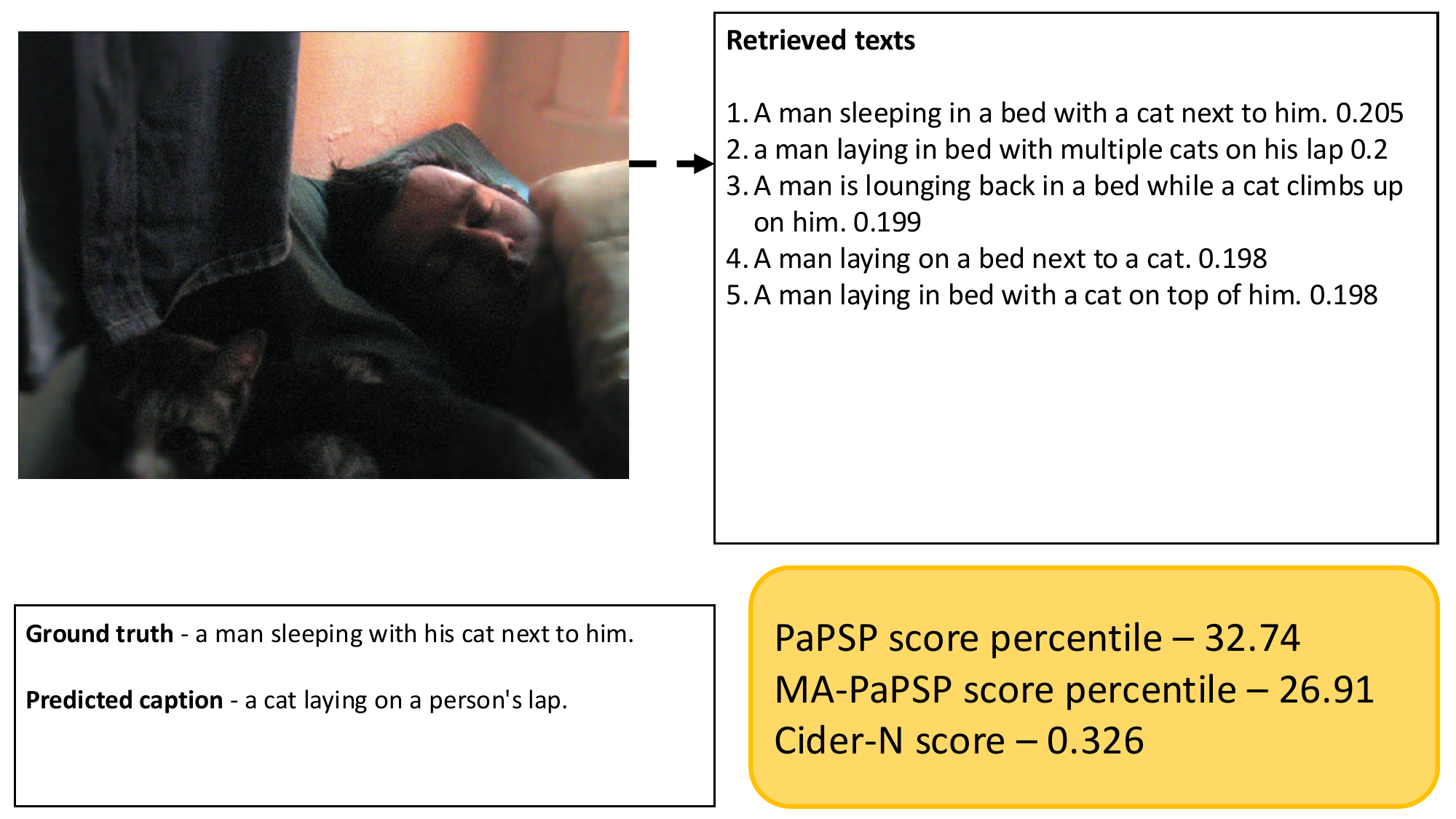}  \\
    {} \\
    \includegraphics[width=0.45\linewidth]{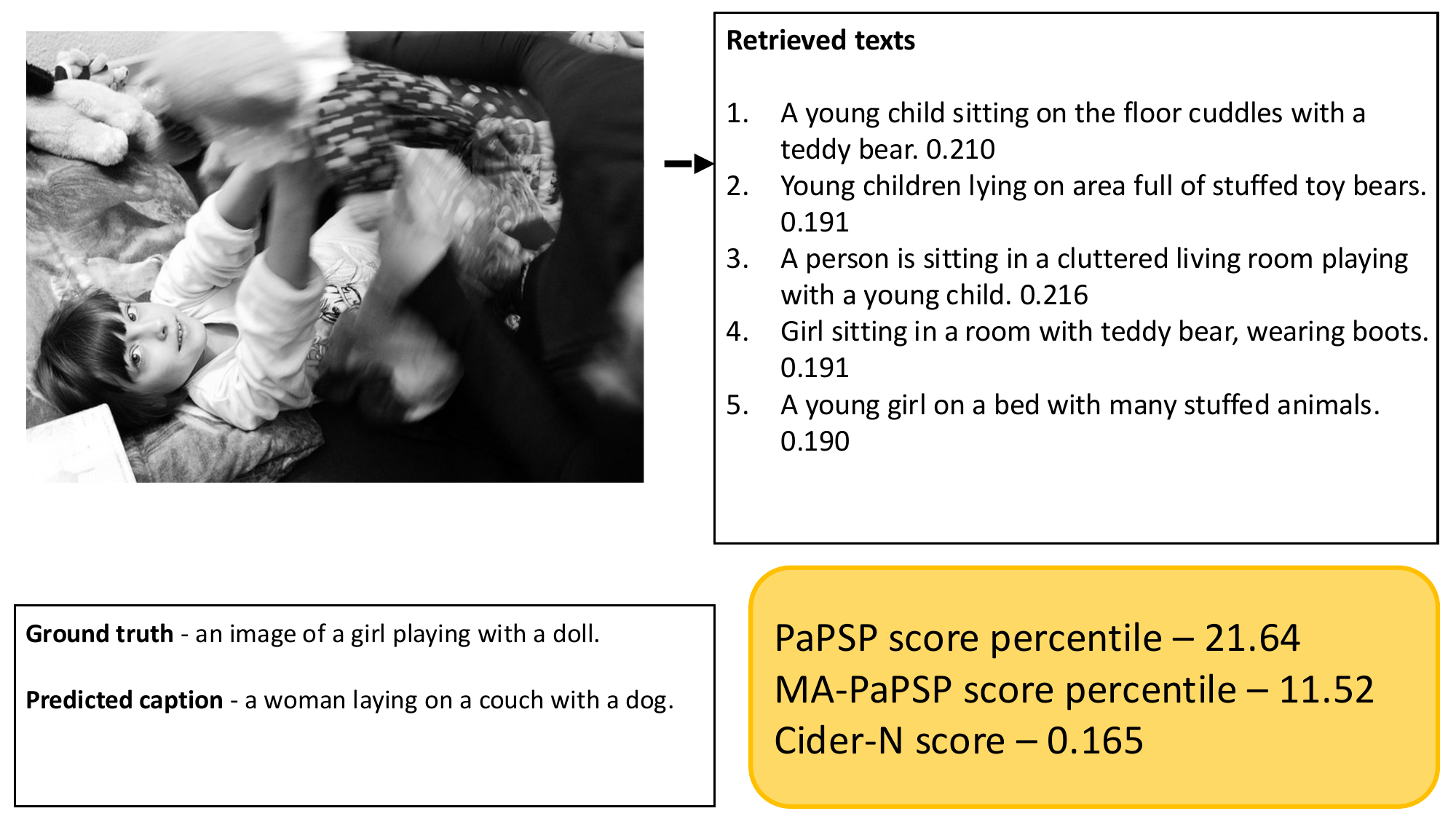} & \includegraphics[width=0.45\linewidth]{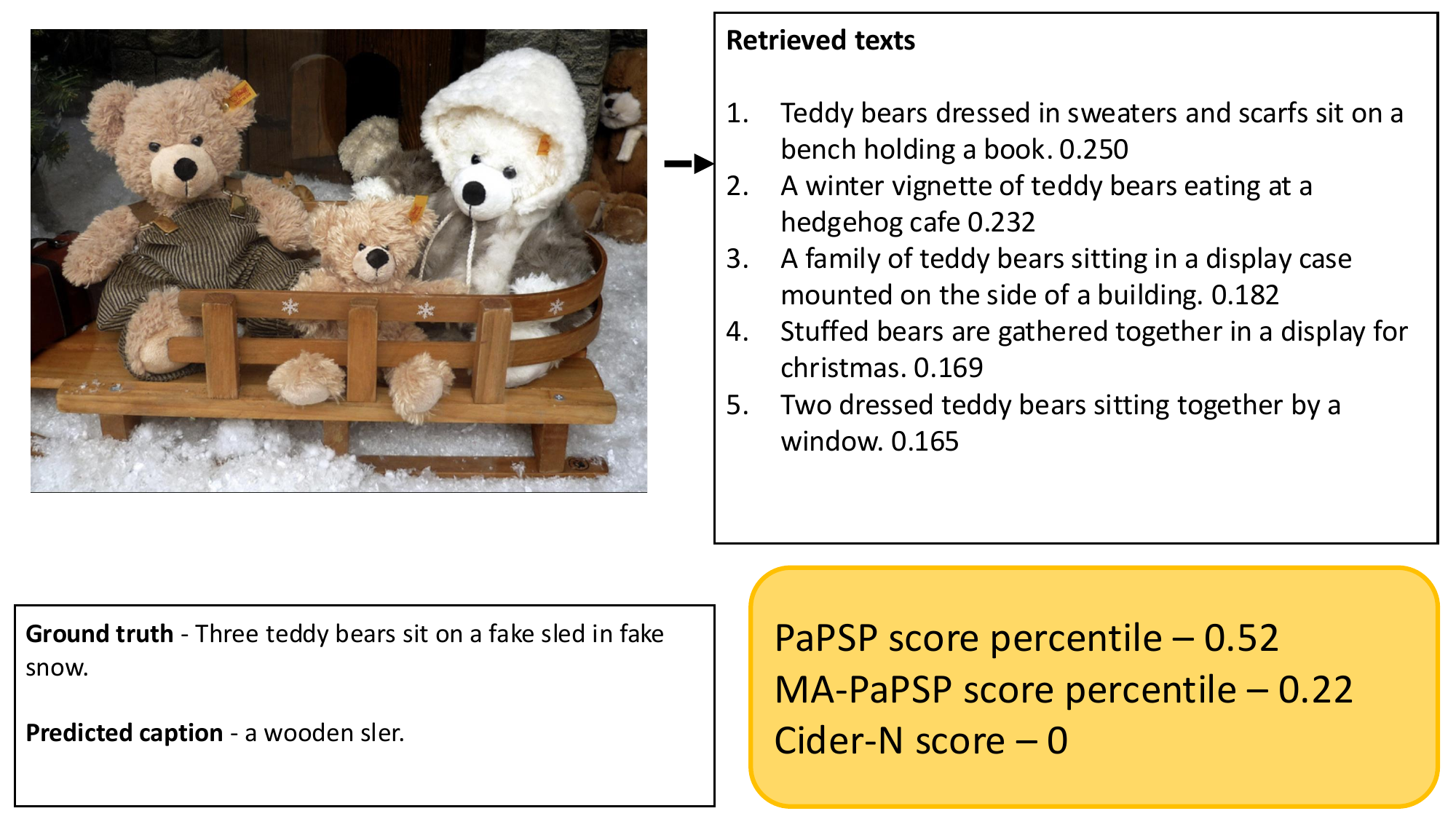} \\
    {} \\
    \end{tabular}
    \caption{This figure presents examples where both \ours and \oursma reject the inputs. Each example includes the following components: the image, its predicted caption, the ground-truth caption, the retrieved captions along with their normalized weights, \ours baseline scores, \oursma scores, and the ground-truth score (Cider-N).}
    \label{fig: Miscc3}
\end{figure*}

\begin{figure*}[hbt!]
    \centering
    \begin{tabular}{cc}
    \includegraphics[width=0.45\linewidth]{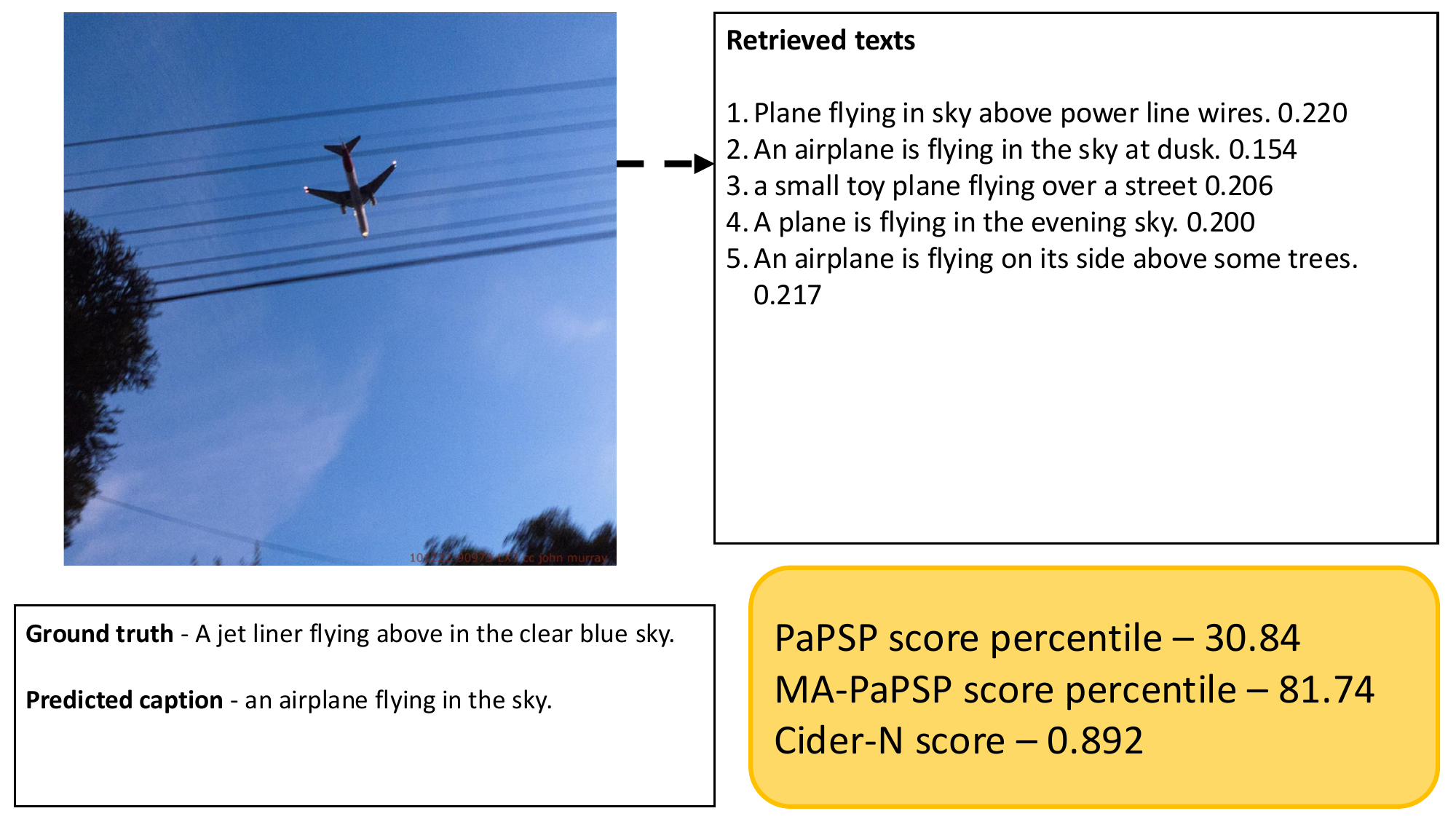} & \includegraphics[width=0.45\linewidth]{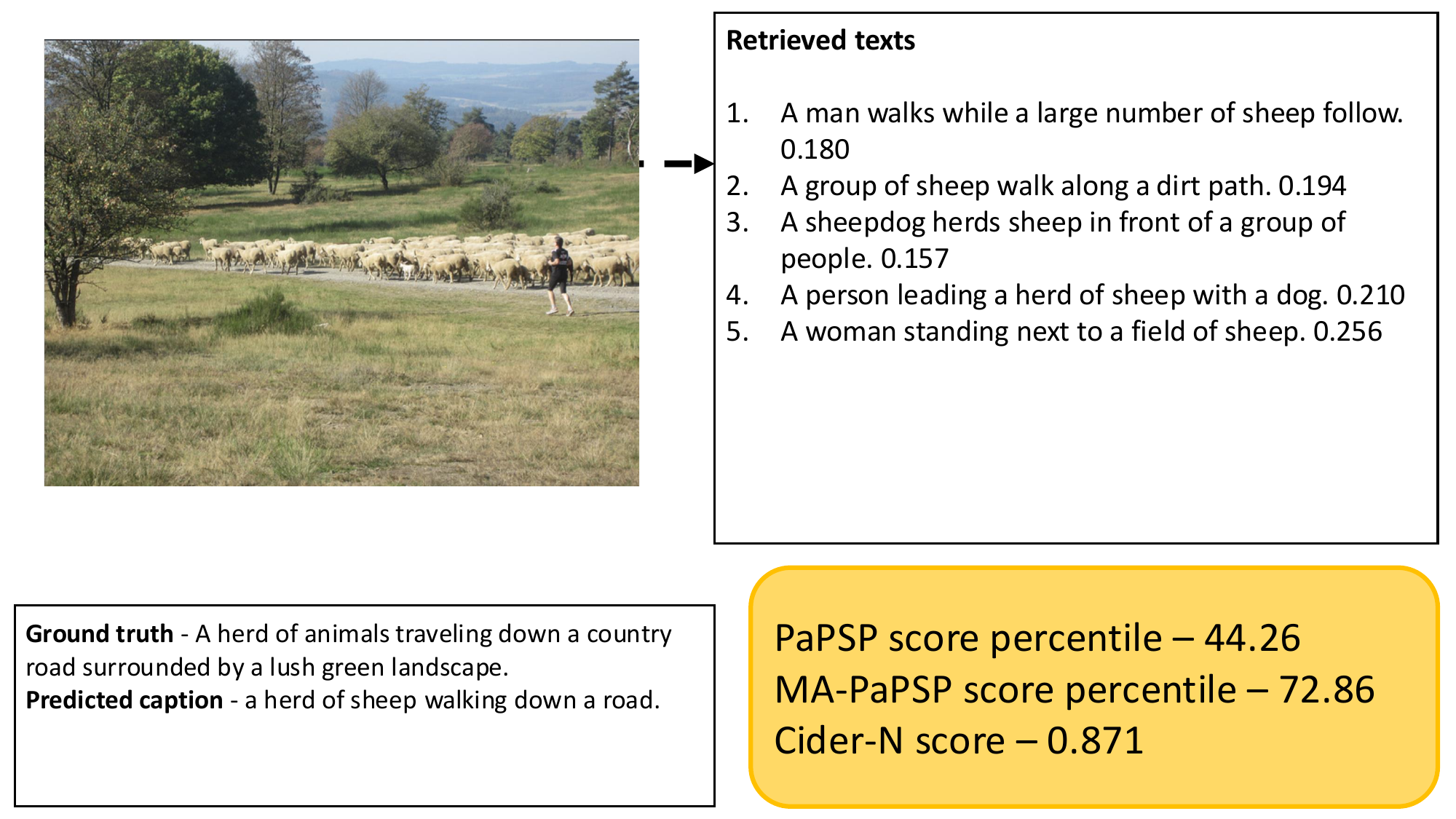}  \\
    {} \\
    \includegraphics[width=0.45\linewidth]{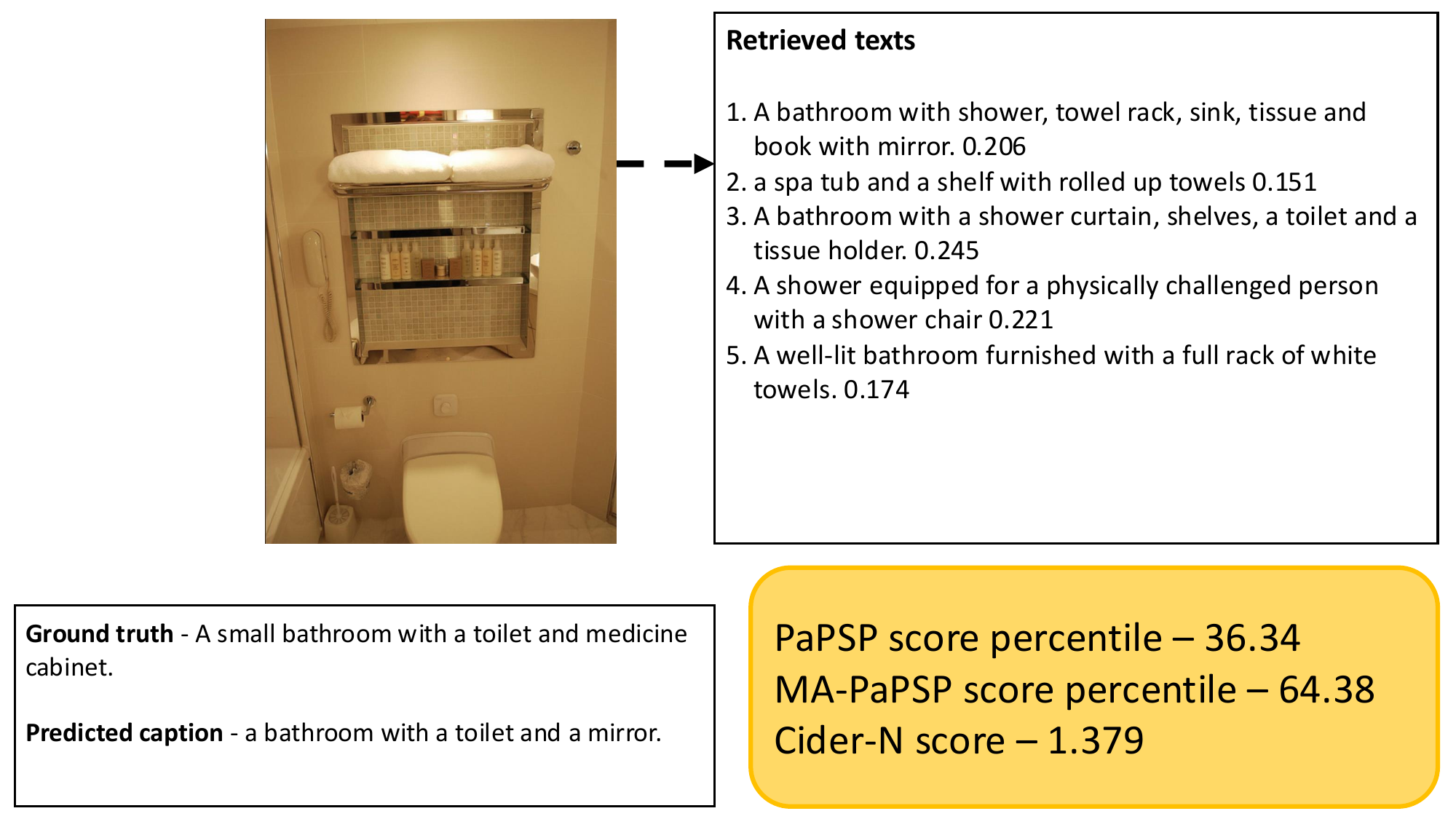} & \includegraphics[width=0.45\linewidth]{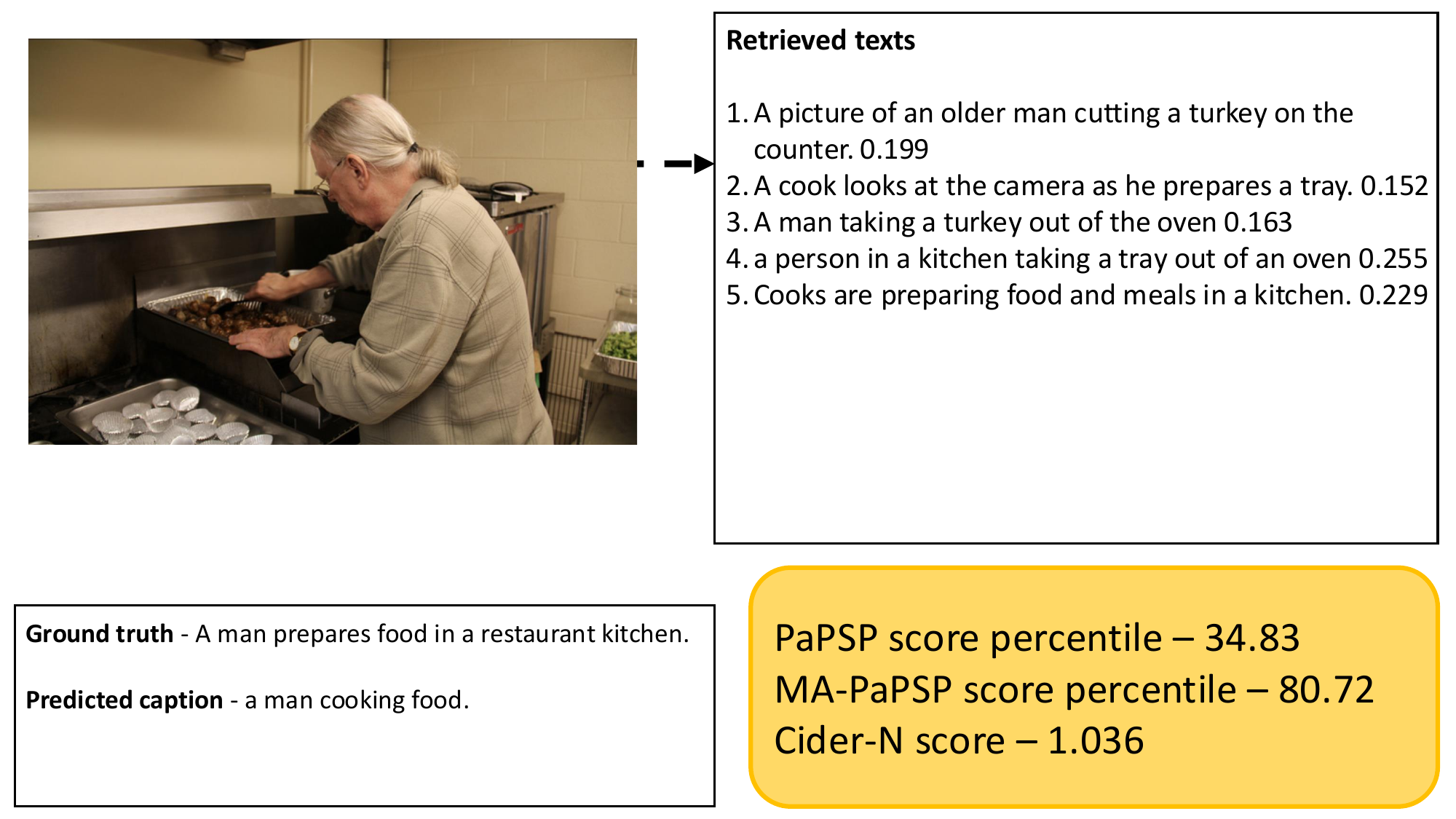} \\
    {} \\
    \end{tabular}
    \caption{This figure presents examples where \ours rejects and \oursma accepts the inputs. Each example includes the following components: the image, its predicted caption, the ground-truth caption, the retrieved captions along with their normalized weights, \ours baseline scores, \oursma scores, and the ground-truth score (Cider-N).}
    \label{fig: Miscc4}
\end{figure*}

We have presented the following two sets of qualitative figures - first, to show how the proxy component helps \oursma perform better than the baseline CLIP scores, we visualize some retrieved samples in Figures \ref{fig: Miscc1}, \ref{fig: Miscc2}, \ref{fig: Miscc3} and \ref{fig: Miscc4}. Next, figure \ref{fig: prompt} shows the prompt we were using to generate contrastive.

\begin{figure*}[hbt!]\RawFloats
    \centering
    \includegraphics[width=0.5\linewidth]{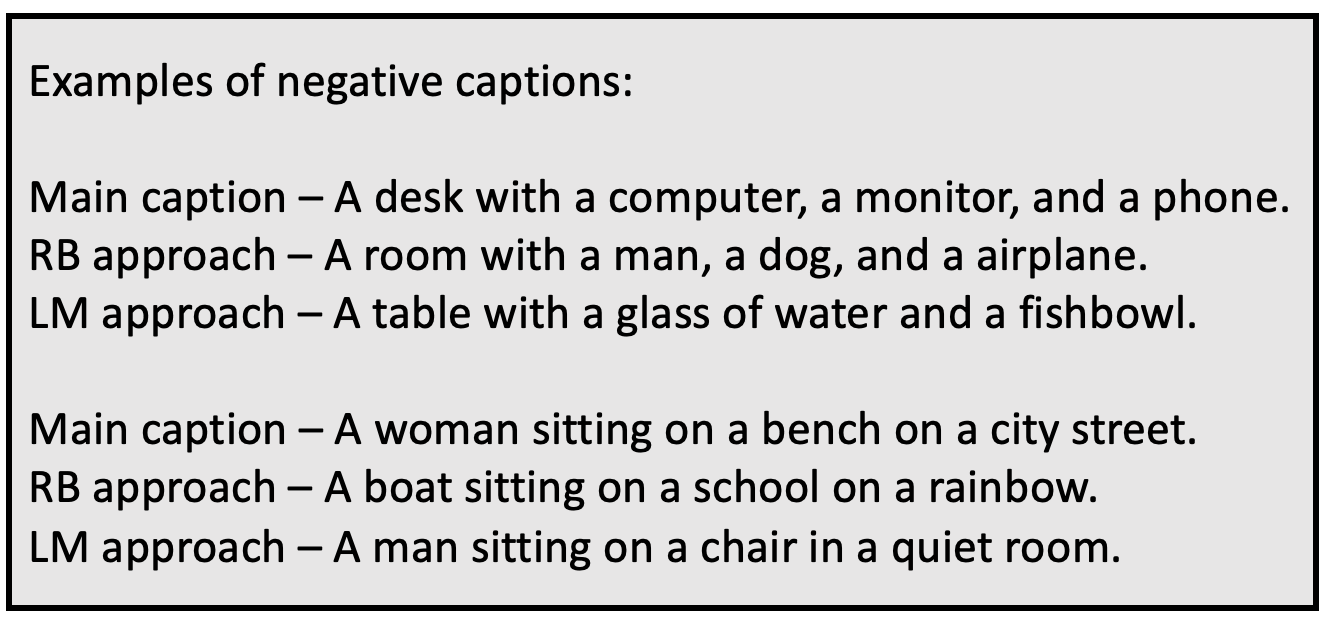}
    \caption{Example of negative captions generated by LM and RB.}
    \label{fig: prompt-1}
\end{figure*}

\clearpage

\clearpage

\end{document}